\documentclass[journal,10pt]{IEEEtran}
\usepackage{graphicx}
\usepackage{amssymb}
\usepackage{amsmath}
\usepackage{cite}
\usepackage{amsthm}
\usepackage{epstopdf}
\usepackage{color}
\usepackage{comment}
\usepackage{soul}
\usepackage{balance}
\usepackage{subcaption}
\usepackage{multirow}
\usepackage{multicol}
\usepackage{booktabs}
\usepackage{amsfonts}
\usepackage{epsfig}
\usepackage{url}
\usepackage{algorithm}
\usepackage{mathrsfs}
\usepackage{amsmath}
\usepackage{physics}
\usepackage{dirtytalk}
\usepackage{comment}
\usepackage[unicode]{hyperref}
\hypersetup{
    colorlinks=true,
    linkcolor=blue,
    citecolor=cyan,
    pdfborder={0 0 0},
}

\usepackage{enumitem}

\usepackage[dvipsnames]{xcolor}
\definecolor{duongfix}{rgb}{0,0,100}

\usepackage[english]{babel}
\usepackage[utf8]{inputenc}
\usepackage{textcomp}
\usepackage{algorithm}
\usepackage[noend]{algpseudocode}

\usepackage{amsmath,amsfonts,bm}

\def\1{\bm{1}}

\DeclareMathAlphabet{\mathsfit}{\encodingdefault}{\sfdefault}{m}{sl}
\SetMathAlphabet{\mathsfit}{bold}{\encodingdefault}{\sfdefault}{bx}{n}

\def\gB{{\mathcal{B}}}

\def\gD{{\mathcal{D}}}

\def\gS{{\mathcal{S}}}
\def\gT{{\mathcal{T}}}

\def\gX{{\mathcal{X}}}
\def\gY{{\mathcal{Y}}}

\newtheorem{definition}{Definition}
\newtheorem{issue}{Challenge}

\newtheorem{theorem}{Theorem}
\newtheorem{lemma}{Lemma}

\newtheorem{example}{Example}

\newtheorem{assumption}{Assumption}

\makeatletter
\def\ScaleIfNeeded{%
\ifdim\Gin@nat@width>\linewidth \linewidth \else \Gin@nat@width \fi
} \makeatother

\definecolor{fix}{rgb}{0,0,0}  

\begin{document}
\title{\huge Energy-Efficient and Real-Time Sensing for \\Federated Continual Learning via Sample-Driven Control}

\author{Minh~Ngoc~Luu, Minh-Duong~Nguyen, 
        Ebrahim~Bedeer, 
        Van Duc Nguyen,
        Dinh~Thai~Hoang,~\IEEEmembership{Senior Member,~IEEE},
        Diep~N.~Nguyen,~\IEEEmembership{Senior Member,~IEEE},
        and Quoc-Viet~Pham,~\IEEEmembership{Senior Member,~IEEE}
\thanks{Minh~Ngoc~Luu is with the School of Electrical and Electronic Engineering, Hanoi University of Science and Technology, Hanoi, Vietnam (e-mail: Minh.LN212161M@sis.hust.edu.vn).}
\thanks{Minh-Duong Nguyen is with the Department of Intelligent Computing and Data Science, VinUniversity, Hanoi, Vietnam (e-mail: mduongbkhn@gmail.com).}
\thanks{Ebrahim~Bedeer is with the Department of Electrical and Computer Engineering, University of Saskatchewan, Saskatoon, Canada S7N 5A9 (e-mail: e.bedeer@usask.ca).}
\thanks{Dinh~Thai~Hoang and Diep~N.~Nguyen are with the School of Electrical and Data Engineering, University of Technology Sydney, Sydney, NSW 2007, Australia (e-mail: \{hoang.dinh, diep.nguyen\}@uts.edu.au).}
\thanks{Van~Duc~Nguyen (corresponding author) is with the School of Electricals and Electronics, Hanoi University of Science and Technology, Hanoi, Vietnam (e-mail: duc.nguyenvan1@hust.edu.vn)}
\thanks{Quoc-Viet Pham is with the School of Computer Science and Statistics, Trinity College Dublin, Dublin 2, D02PN40, Ireland (e-mail: viet.pham@tcd.ie).}
\thanks{A part of this work was done when Minh-Duong Nguyen was with Pusan National University}
\thanks{The first two authors contributed equally to this work.}
}
\maketitle
\begin{abstract}
An intelligent Real-Time Sensing (RTS) system must continuously acquire, update, integrate, and apply knowledge to adapt to real-world dynamics. Managing distributed intelligence in this context requires Federated Continual Learning (FCL). However, effectively capturing the diverse characteristics of RTS data in FCL systems poses significant challenges, including
severely impacting computational and communication resources, escalating energy costs, and ultimately degrading overall system performance.
To overcome these challenges, we investigate how the data distribution shift from ideal to practical RTS scenarios affects Artificial Intelligence (AI) model performance by leveraging the \textit{generalization gap} concept. 
In this way, we can analyze how sampling time in RTS correlates with the decline in AI performance, computation cost, and communication efficiency.
Based on this observation, we develop a novel Sample-driven Control for Federated Continual Learning (SCFL) technique, specifically designed for mobile edge networks with RTS capabilities. 
In particular, SCFL is an optimization problem that harnesses the sampling process to concurrently minimize the generalization gap and improve overall accuracy while upholding the energy efficiency of the FCL framework.
To solve the highly complex and time-varying optimization problem, we introduce a new soft actor-critic algorithm with explicit and implicit constraints (A2C-EI). Our empirical experiments reveal that we can achieve higher efficiency compared to other DRL baselines. Notably,  SCFL can significantly reduce energy consumption up to $85\%$ while maintaining FL convergence and timely data transmission.
\end{abstract}
\section{Introduction}

Federated Learning (FL) \cite{2017-FL-FedAvg,2020-FL-Survey} is a distributed machine learning concept that is widely utilized in Internet-of-Things (IoT), underwater acoustic networks \cite{Shaheen2024, jmse12091656} and wireless systems \cite{2019-FL-FederatedPCA,le2024applications}. FL allows IoT devices to construct a collaborative learning model by locally training their collected data \cite{2025-FDG-FedOMG, 2023-FL-LKD, 10542323}. Instead of sharing the training data, the devices in FL can collaboratively perform a learning task by only uploading their local learning models to the aggregation server at the network edge \cite{liu2024vertical}. 
Although promising, FL encounters several critical challenges of adaptively learning new tasks, particularly in mobile edge networks, where there exist data streams from distributed devices at the same time.
\textcolor{black}{To resolve these issues, the research community has proposed} a new concept called Federated Continual Learning (FCL) \cite{2022-FCL,2021-FCL,2024-FCL-sur, 10444954,zhao2024towards} to optimize \textcolor{black}{data streams in} Real-Time Sensing (RTS) systems (e.g., underwater IoT devices \cite{s22228727}). 

\subsection{Challenges and Motivation}
\begin{figure}[t] 
\centering
    \begin{subfigure}[b]{0.3333\linewidth}
        \centering
        \includegraphics[width=\linewidth]{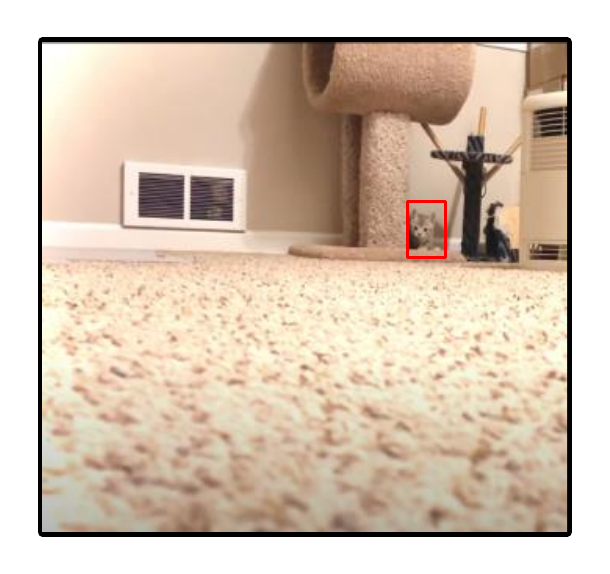}
        \caption{$t$}
        \label{fig:D_kitten01}
    \end{subfigure}%
    \begin{subfigure}[b]{0.3333\linewidth}
        \centering
        \includegraphics[width=\linewidth]{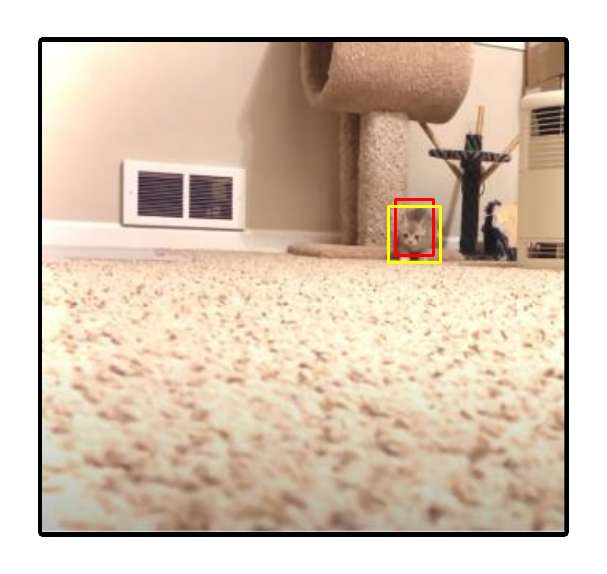}
        \caption{$t + \tau_{u}$}
        \label{fig:D_kitten02}
    \end{subfigure}%
    \begin{subfigure}[b]{0.3333\linewidth}
        \centering
        \includegraphics[width=\linewidth]{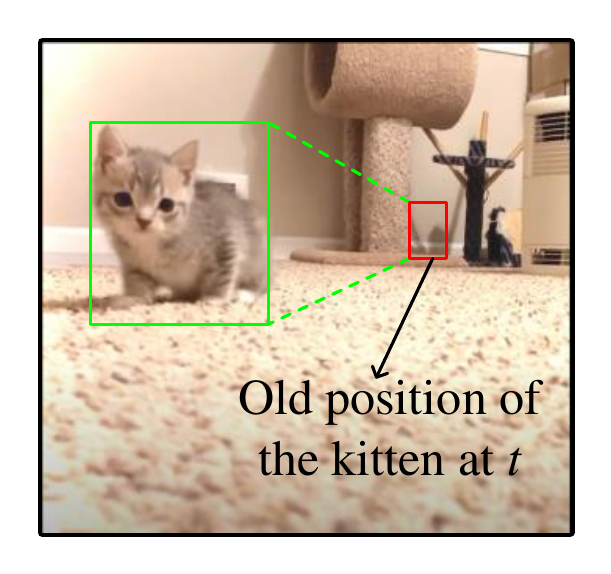}
        \caption{$t + k\times\tau_{u}$}
        \label{fig:D_kitten03}
    \end{subfigure}%
    \caption{Illustration of how sampling delay impacts overfitting in AI. Images from ViralHog's video: \url{www.youtube.com/watch?v=hZWXS_ns0Mo&t=31s}}
\label{fig:D_kitten}
\end{figure}

FCL can proceed data streams in RTS systems to adapt the changes in dynamic mobile edge networks. However, similar to the vanilla FL, it faces significant challenges related to computation, communication, and energy efficiency due to the distributed nature of the system and the resource limitation of distributed clients. 
In order to address these challenges, recent studies have focused on various techniques, including model compression \cite{2022-FL-HCFL}, quantization \cite{10018182, Chen_2024_CVPR, 10091800, 10368103}, sparsification \cite{Jiang_Borcea_2023, 10256151, 10360319}, and joint computational and communicational resource management \cite{2020-FL-JointLearning-CommunicationsFramework, 2019-FL-OptimizationModel, 2020-FL-EnergyEfficientFL, 2021-FL-ConvergenceAnalysis-ResourceAllocation}.
{
\color{black}Despite the aforementioned advancements, the role of data sampling in FCL systems remains underexplored, particularly concerning its impact on energy-constrained edge devices. While some studies have incorporated data sampling strategies, such as using the age of data as a criterion to improve efficiency, these methods may overlook the value of older data in mitigating catastrophic forgetting \cite{10444954}.
Recent work has begun to address these forgetting challenges. For instance, \cite{dong2023no} proposed the Local-Global Anti-forgetting (LGA) model, which combines category-balanced loss functions and proxy-based semantic distillation to mitigate both local and global catastrophic forgetting. Similarly, \cite{dong2023federated} introduced a forgetting-balanced learning framework for federated semantic segmentation that utilizes class-balanced pseudo labeling and a task transition monitor under privacy constraints. However, issues like data correlation, particularly critical in RTS, remain largely unaddressed in the context of FCL \cite{dong2023no, dong2023federated}.
}
For the sake of illustration, we consider the following example:


\begin{example}\label{ex:D_umbling_Kute_Kitten}
    We have three images of a kitten running towards the security camera observed at various times following specific intervals that begin from an initial point in time $t$ in Fig.~\ref{fig:D_kitten}. The security camera is trained for the object detection task. To fully comprehend the influence of real-time sampling on the data collected from security cameras, we consider the two descriptions for the differences of the images in three instances.
    \begin{itemize}
        \item Fig.~\ref{fig:D_kitten01} at time $t$ depicts the kitten at a distance from the camera, blurred and difficult to identify. After a very short time interval $\tau_{u}$, the cat in Fig.~\ref{fig:D_kitten02} has almost the same detail and size as it is in Fig.~\ref{fig:D_kitten01}.
        \item In Fig.~\ref{fig:D_kitten03} at time $t + k\times\tau_{u}$, the kitten is noticeably different from itself in Figs.~\ref{fig:D_kitten01} and \ref{fig:D_kitten02}. Thus, the information about the kitten in these figures is significantly different.
    \end{itemize}
\end{example}

It is observed from Example~\ref{ex:D_umbling_Kute_Kitten} that utilizing a short sampling time in the RTS setting of the FCL system results in small variations between consecutively sampled images. As a consequence, when considering a batch of sampled data in a specific timeframe, the characteristics learned by the Artificial Intelligence (AI) model tend to be skewed toward their identity when they are situated at a distance from the camera. This bias in the AI model struggling to identify the kitten when it is close to the camera, due to the insufficiency of data with related instances. 
\textcolor{black}{Moreover, without a mechanism to skip redundant data during collection, the dataset may include many identical or highly similar samples. This redundancy increases memory usage and processing time. More importantly, training on such repetitive data can harm the model’s ability to generalize, increasing the risk of overfitting \cite{jiang2023importance,neyshabur2017exploring}.}
In summary, inefficient sampling processes can impose additional computational burdens and exacerbate communication overhead, posing challenges for deploying AI services on resource-constrained devices at the network edge. In addition, this inefficiency often results in slower convergence, thereby increasing the number of communication rounds required to achieve satisfactory performance.
\begin{issue}\label{chal_1}
    The AI model in FCL faces challenges in processing redundant data which affects the unbalanced characteristics, leading to \textbf{prolonged processing time} and a \textbf{risk of overfitting}.
\end{issue}

\begin{issue}\label{chal_2}
    One straightforward approach to address Challenge~\ref{chal_1} is to 
    skip
    \textcolor{black}{more} samples from sensing devices, such as cameras and sensors (i.e., increasing sampling interval); however, it results in increased \textbf{sampling energy consumption} by the FCL system and longer waiting times to gather sufficient data for the training process \cite{jin-energy-awareness-2022, Luo-cost-effective-2021}.
\end{issue}
To address these two challenges, it is important to strike a balance that optimizes the sampling interval while efficiently utilizing energy and time resources. Consequently, to enhance the training performance of distributed AI models, it is essential to simultaneously achieve two objectives, including 1) \emph{Sample data with diverse information} (i.e., sufficiently large sampling interval between two samples) to increase AI model's performance and 2) \emph{Consider sampling energy consumption} of distributed clients to enhance energy efficiency and enable energy-constrained edge evices to perform AI model's training and deployment.
 
{\color{black}
\subsection{Related Works}\label{sec:relate_works}
Recently, a few works have begun to explore FL on dynamic client data, which are called FCL. Our research focuses on enhancing FCL by developing strategies to select training samples based on their relative importance.

The LGA method \cite{dong2023no} proposes a technique to balance class-wise gradient contributions, aiming to alleviate catastrophic forgetting resulting from class imbalance across incremental tasks. Similarly, Re-Fed \cite{Li_2024_CVPR} introduces a mechanism to compute an importance score, which guides the selective retention of samples in the replay buffer to improve memory efficiency.
An alternative approach to sample selection during continual learning involves the use of learnable coreset selection losses. BCSR \cite{NEURIPS2023_a0251e49} formulates a bi-level optimization framework to select coresets that encapsulate the most salient features of previously encountered data. DECO \cite{10656539} proposes a label-guided, gradient-based similarity scoring system to refine coreset selection, ensuring equitable representation across all classes. OCS \cite{yoon2022online} selects a representative and diverse subset of samples from each minibatch that maintains high affinity with previously seen data. CSReL \cite{tong2025coreset} employs a reducible loss metric to quantify the impact of including specific samples on model performance.
Other methods focus on gradient approximation. GCR \cite{9880055} maintains a coreset that approximates the gradient of all previously observed data with respect to the current model parameters. GradMA \cite{Luo_2023_CVPR}, GPM \cite{saha2021gradient}, and FS-DGPM \cite{deng2021flattening} preserve knowledge of prior tasks by storing gradient components that capture the most informative aspects of past data.

While these approaches demonstrate strong performance, they do not consider the trade-off between reducing redundant data vua data selection (Challenge~\ref{chal_1}) and the resource limitations of IoT devices when sampling data (Challenge~\ref{chal_2}). Our work addresses this gap by incorporating device constraints into the sample selection process.
}

\subsection{Contributions}

In this work, we aim to develop an innovative framework that not only addresses Challenge~\ref{chal_1} and Challenge~\ref{chal_2} but also facilitates effective learning across non-Independent and Identically Distributed (non-IID) datasets for the FCL system within RTS IoT networks. 

Firstly, to understand the degradation of AI models in different learning scenarios, we utilize a concept called the \emph{generalization gap}. Specifically, \emph{generalization gap} can approximate the difference between an AI model’s performance on a training dataset and its performance on the unseen global dataset by considering the statistical distance between distributions of two datasets (i.e., distribution shift)\cite{2021-DG-IRL-DDT, 2020-DG-EntropyReg}. By establishing a relationship between RTS sampling intervals and distributional shifts, we can formulate the generalization gap that captures the degradation in AI performance \cite{2020-DL-DataOverfit} when the FCL system is installed in RTS IoT Networks.

Secondly, based on the proposed \emph{RTS-driven generalization gap} function, we introduce a novel convergence analysis for FCL. This analysis employs sample-driven control to manage data distribution shifts at distributed clients, using the \emph{generalization gap} of mutual information. Our approach ensures that the overall system achieves optimal energy efficiency for the FCL convergence. Based on this convergence analysis, we develop a new joint communication and computation model for FCL known as Sample-driven Control for Federated Continual Learning (SCFL) which can mitigate data-related issues encountered in RTS IoT networks. To be more precise, our optimization problem aims to minimize the total energy consumption required to achieve FCL convergence. Unlike existing approaches focused on energy efficiency for FL networks \cite{2020-FL-EnergyEfficientFL, 2019-FL-OptimizationModel}, our method accounts for FCL convergence on non-ideal datasets by incorporating sampling intervals into the convergence function. \textit{To the best of our knowledge, this work is the first attempt to represent a sample-driven FCL model to account for the dynamics of the RTS IoT data collection process along with the energy limitations, and resource allocation constraints.}

Thirdly, to ensure the flexibility and stability of the SCFL model within the context of FCL techniques, we develop a Deep Reinforcement Learning (DRL) based algorithm that can find the best policy for IoT devices. To solve this highly complex problem effectively, we introduce an innovative approach named Soft Actor-Critic with Explicit Constraints and Implicit Constraints (A2C-EI). Apart from the conventional Soft Actor-Critic concept, A2C-EI disentangles the optimization problem constraints into two distinct components: Explicit Constraints (ECS), which are directly tied to the action outputs, and Implicit Constraints (ICS), which encompass factors unrelated to the action outputs. Hence, through the active configuration of action outputs based on ECS, we can markedly diminish the computational complexity of optimization problems. Our major contributions can be summarized as follows.
\begin{itemize}
    \item We derive a theoretical analysis of the convergence rate of the FCL system under the RTS settings. Our proposed convergence theorem can handle non-IID issues induced by the sampling process of RTS IoT devices for sample-driven FCL control.
    \item We propose SCFL, the first-ever joint communication and computation model to cover and account for non-IID challenges precipitated by the RTS IoT networks data collection process in FCL concept.
    \item We develop a highly-effective DRL approach for SCFL defined as A2C-EI to optimize and deal with the complex and time-varying SCFL problem. 
    \item We conduct a comprehensive performance evaluation to showcase the effectiveness of the proposed sampling-driven control solution and offer valuable insights under various simulation settings. 
\end{itemize}

\subsection{Paper Organization}
The rest of this paper is structured as follows. In Section \ref{sec:System_model}, we present the system model and FCL system model. Section \ref{sec:ResourceAllocation} provides a comprehensive overview of our proposed SCFL approach, along with the problem formulation. In Section \ref{sec:DRL_SCFL}, we delve into the investigation of a DRL approach for SCFL. Experimental settings and results are discussed in Section \ref{sec:Results}. Finally, in Section \ref{sec:Conclusion}, we conclude the paper.

\section{System Model}\label{sec:System_model}
\begin{figure}[!t]
\centering
\includegraphics[width = 1.0\linewidth]{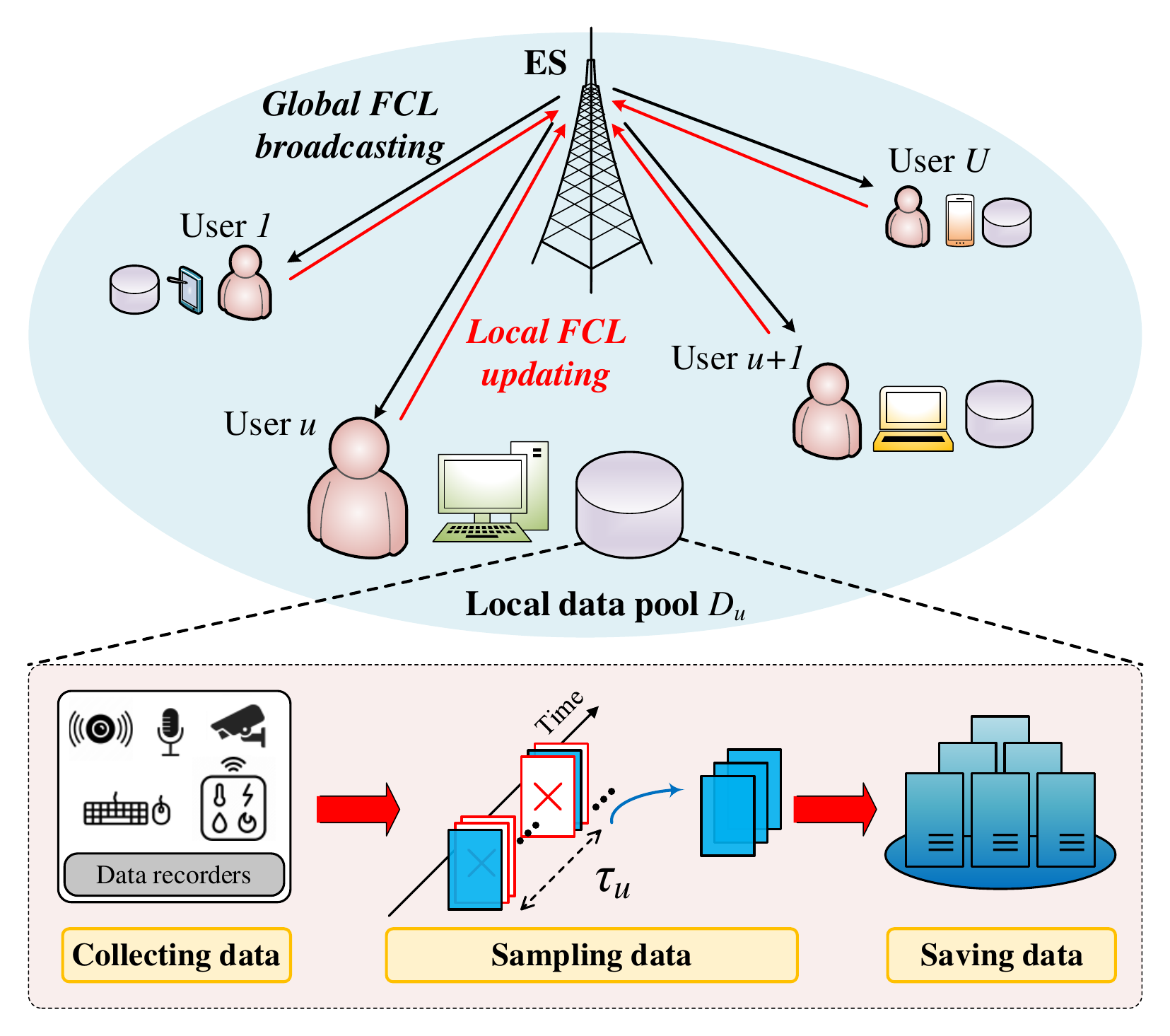}
\caption{The architecture of our FCL framework with online data sensing.}
\label{fig:system-model}
\end{figure}

We consider a network that consists of one edge server (ES) that is equipped with computing capabilities and serves $U$ users, denoted as $\mathcal{U}=\{1,2,\ldots,U\}$ (as shown in Fig.~\ref{fig:system-model}). Each user $u$ has a local dataset $\mathcal{D}_u$ with $D_u$ data samples. For each dataset $\mathcal{D}_u = \{x_{u,i}, y_{u,i}\}^{D_u}_{i=1}$, $x_{u,i} \in \mathbb{R}^d$ is an input vector of user $u$ with dimensionality $d$, and $y_{u,i}\in\mathbb{R}$ is its corresponding output (e.g., the data ground truth labeled by experts). In the considered network, IoT devices do sample the training data from the locally available datasets. IoT devices can also sample the data from a static environment, where the data distributions remain fixed during the learning process. The ES and all IoT devices of users cooperatively perform an FCL algorithm over wireless networks for data analysis and inference. Hereinafter, the FCL model that is trained by each user's dataset is called the \textit{local FCL model}, while the FCL model that is generated by aggregation algorithms at the ES (e.g., FedAvg \cite{2017-FL-FedAvg}, FedProx \cite{2020-FL-FedProx}, SCAFFOLD \cite{2020-FL-Scaffold}) is called the \textit{global FCL model}.

\subsection{Computation and Transmission Model}
The FCL cycle between the ES and the users in its service area consists of four steps: 1) Data sampling, which is performed at the user, with arbitrary sampling interval; 2) Local computation, which is performed at the user, with numerous local iterations; 3) Wireless communication (the users send their local FCL models after finishing and receiving parameters from ES); and 4) ES operation, which comprises aggregation and broadcasting.
\subsubsection{Local Sampling}
User $u$ is equipped with real-time sensors to capture data for online training. Each sensor possesses optimal sampling capabilities $\tau_u$ (e.g., the shortest possible time that a sensor can capture two consecutive data samples). We denote $k_u$ as the number of sampling times skipped between two actual consecutive data points recorded by the sensor. The sensor consumes power $\mathcal{P}_u$ (Jules per sample) to capture each sampling data point. Thus, we can estimate the total sampling energy consumed by user $u$ as:
\begin{equation}
   E_{u}^{\mathrm{S}}=k_u \mathcal{P}_u.
\label{eq:ES}
\end{equation}
\subsubsection{Local Computation}
The computation time required for data processing at user $u$ is defined as
\begin{equation}
\label{eq:local-computing-time}
    t^{\mathrm{comp}}_u = I_u\frac{C_u D_u}{f_u},\quad \forall u \in \mathcal{U},
\end{equation}
where $I_u$ is the number of local iterations executed by each user $u$, $C_u$ [cycles/sample] and $f_u$ [Hz] is the number of CPU cycles needed to compute one sample data and the CPU frequency of user $u$, respectively. We denote $\kappa$ as the effective switching capacitance determined by the chip architecture. Consequently, user $u$ requires the following energy for local training \cite{2020-FL-EnergyEfficientFL}: 
\begin{equation}
   E_{u}^{\mathrm{C}}=\kappa I_u C_u D_u f_u^2.
\end{equation}
\subsubsection{Wireless Communication}
After local computation, all the users employ frequency domain multiple access to submit their local FCL model to the ES. According to \cite{2020-FL-EnergyEfficientFL}, the user $u$'s achievable rate is
\begin{equation}
\label{eq:transmission-rate}
    r_u=b_u \log _2\left(1+\frac{g_u p_u}{n_0 b_u}\right),\quad \forall u \in \mathcal{U},~\sum^U_{u = 1}b_u \leq B,
\end{equation}
where $b_u$ is the user's allocated bandwidth, $g_u$ is the channel gain of the wireless channel between user $u$ and the ES, $p_u$ is the average transmit power of user $u$, $n_0$ is the Gaussian noise power spectral density, and $B$ is the total bandwidth of FCL system.
{\color{black}Since all users employ a similar local model architecture, the size of the local model updates $M_0$ transmitted by each user over the wireless network remains constant.}
Inspired by \cite{2020-FL-EnergyEfficientFL}, our objective is to ensure that the transmission time $t_u^{\mathrm{trans}}$ for each user $u$ does not exceed the maximum timeout $T_{\mathrm{max}}$. Specifically, to upload model with size $M_0$ during transmission time $t_u^{\mathrm{trans}}$, we hold the condition that $r_u t_u^{\mathrm{trans}} \geq M_0$, where $r_u$ is the data rate of user $u$. Consequently, the transmit energy required by user $u$ to transfer data over a duration of $t_u^{\mathrm{trans}}$ is 
\begin{equation}
    E^T_{u} = p_ut^{\mathrm{trans}}_u.
\end{equation}
\subsubsection{ES Operation}
The ES receives the local parameters from all the users in the global iteration and aggregates them for the global FCL model at this stage. The ES transmits the global FCL model to all the users on the downlink. Because of the ES's strong transmit power and the large bandwidth available for data broadcasting, the downstream time is disregarded compared to the uplink data transmission time. During this phase, the ES cannot access the user's $\mathcal{D}_u$ dataset. This signifies that the FCL algorithm's privacy protection criterion protects the user's privacy.

According to the aforementioned FCL model system, each user's energy consumption comprises both local computing energy $E^C_u$, sampling energy $E^S_u$ and wireless transmission energy $E^T_u$. The FCL system implements a number of global communication rounds defined as $I_\mathrm{glob}$, so the total energy consumption by all the users is:
\begin{equation}
    E = I_\mathrm{glob} \sum\nolimits_{u = 1}^{U} \left( E_u^C+E_u^T+E_u^S\right).
\end{equation}
The entire time required to complete the execution of the FCL algorithm is referred to as the completion time. Based on~\eqref{eq:local-computing-time} and the transmission time $t_u^{\mathrm{trans}}$, the completion time of each user $u$ is defined as follows:
\begin{equation}\label{Ttotal}
    T_{u,\mathrm{total}} = I_\mathrm{glob}(t^{\mathrm{comp}}_u+t_u^{\mathrm{trans}}) = I_\mathrm{glob}\left(I_u\frac{C_u D_u}{f_u}+t_u^{\mathrm{trans}}\right).
\end{equation}
\textcolor{black}{Let $T_{\mathrm{max}}$ denote the maximum completion time of a user for the entire FCL algorithm. The total completion time $T_{u,\mathrm{total}}$ of each user $u$ is determined by implementing FDMA \cite{2020-FL-EnergyEfficientFL} within the restricted time for all users and is defined as}
\begin{equation}
    T_{u,\mathrm{total}} \leq T_{\mathrm{max}},\quad \forall u \in \mathcal{U}.
\end{equation}
\subsection{Federated Continual Learning Model}
\label{sec:FL-model}
\subsubsection{Generic Model}
We denote $\omega^{(n)}$ as the parameters of the global FL model at the $n$-th global iteration, and $\vartheta^{(n)}_u$ signifies the disparity between the global model and the local model of each IoT device. $\omega^{(n)} + \vartheta^{(n)}_u$ denotes the local parameters of user $u$ at the $n$-th global iteration. In this context, the loss function for each user $u$ in the $n$-th global iteration with dataset $\mathcal{D}_u$ as $\ell(x_{u,i}, y_{u,i};\omega^{(n)})$ can be defined as follows \cite{2020-FL-EnergyEfficientFL}:
\begin{align}
    \mathcal{L}_u\left(\omega^{(n)}\right) = \frac{1}{D_u} \sum^{D_u}_{i=1}\ell\left(x_{u,i}, y_{u,i};\omega^{(n)}\right),
\label{eq:FL-problem}
\end{align}
where the empirical loss can be mapped as follows: 
$$\ell\left(\cdot\right) = 
\begin{cases}
\frac{1}{2}\left(x_{u,i}\times \omega^{(n)} - y_{u,i}\right)^2, \text{ for linear regression,}\\
-\log\left(1+ \exp{-y_{u,i}\left( x_{u,i}^\top\times\omega^{(n)}+\mathrm{bias}\right)}\right),\\ \quad\text{for logistic regression}.
\end{cases}$$ 

In the FL problem, all the users $u\in \mathcal{U}$ aims to train the depreciation of the global loss function:
\begin{align}\label{eq:glob_loss}
    \min_{\omega} \mathcal{L}\left(\omega^{(n)}\right) \triangleq \frac{1}{U} \sum\nolimits^{U}_{i=1}\mathcal{L}_u\left(\omega^{(n)}\right).
\end{align}
In practice, each user solves the local optimization problem:
\begin{align}
\label{eq:Yang11}
\min_{\vartheta^{(n)}_u \in \mathbb{R}^d} \mathcal{G}_u(\omega^{(n)},\vartheta^{(n)}_u) &\triangleq \mathcal{L}_u(\omega^{(n)}+\vartheta^{(n)}_u)\\
&-(\nabla \mathcal{L}_u(\omega^{(n)})-\xi\nabla \mathcal{L}(\omega^{(n)}))^\top\vartheta^{(n)}_u.\notag
\end{align}
When $\vartheta^{(n)}_u=0$, we can obtain
\begin{align}
\label{Eq:EE10}
\mathcal{G}_u(w^{(n)},0)=\mathcal{L}_u(w^{(n)}),
\end{align}
where $\vartheta^{(n)}_u$ is the learned step on user $u$ of parameters $\omega^{(n)}$. To establish the convergence condition for the gradient method, we adopt the concept of local accuracy from \cite{2020-FL-EnergyEfficientFL}. In this context, we define $\vartheta^{(n),(i)}_u$ of the problem~\eqref{eq:Yang11} to have accuracy $\varpi$ if
\begin{align} 
    & \mathcal{G}_u\left(\omega^{(n)},  \vartheta_u^{(n),(i+1)}\right)-\mathcal{G}_u\left(\omega^{(n)},  \vartheta_u^{(n) *}\right) \notag \\
    & \leq \varpi \left(\mathcal{G}_u\left( \omega^{(n)}, 0\right)-\mathcal{G}_u\left( \omega^{(n)},  \vartheta_u^{(n) *}\right)\right), \label{eq:Yang15}\\
    & \omega^{(n+1)}=\omega^{(n)}+\frac{1}{U}\sum^{U}_{u=1}\vartheta^{(n)}_u, \label{Eq:10} \\   
    & I_u \geq \frac{2}{(2-L\delta)\delta\mu}\log_2(1/\varpi) \label{eq:yang16},
\end{align}
where $\varpi$ is the desired local accuracy of the AI model on distributed users. The equations~\eqref{eq:Yang15}, \eqref{Eq:10}, \eqref{eq:yang16} are adopted from \cite{2020-FL-EnergyEfficientFL} as mentioned above. We define a global accuracy $\varrho$ for the global FL model. The solution $\omega^{(n)}$ with accuracy $\varrho$ is a point that
\begin{align}
\mathcal{L}\left(\omega^{(n)}\right)-\mathcal{L}\left(\omega^*\right) \leq \varrho\left(\mathcal{L}\left(\omega^{(0)}\right)-\mathcal{L}\left(\omega^*\right)\right),
\label{eq:global-acc}
\end{align}
where $\omega^*$ is the actual optimal solution of problem \eqref{eq:FL-problem}. As evident from \eqref{eq:Yang15} and \eqref{eq:global-acc}, both local and global accuracy are assessed based on the extent to which an AI model deviates from the initial model and approaches the optimal point. For example, if the global accuracy value $\varrho = 0$, the AI model at round $n$ achieves the optimal position, and otherwise. 
\subsubsection{Continual Learning in Federated Learning with RTS system}
Consider an FCL system with $U$ users, the dataset $\gD_u$ of each user $u$ is sampled from a data pool $\Bar{\gD}_u = \{\Bar{\gD}^{t}_{u}\vert~t=1,\ldots,T\}$, where $T$ is a total of sensing time and each sample $\Bar{\gD}^{i}_{u} = (x_{u,i}, y_{u,i})$ represents a data instance. It is noteworthy that the sampling time between two consecutive samples for user $u$ is $\tau_u$. In our work, given sampling rule $s\in\gS$, we can choose the dataset $\gD_u$ by skipping samples from the ideal data pool $\Bar{\gD}_u$ according to the following formula:
\begin{align}
    \gD_u = \{\bar{\gD}^{t}_{u}\vert t = i\times k + 1; i = 0,\ldots,\lfloor T/k_u \rceil-1\},
\end{align}
where $\lfloor\cdot\rceil$ refers to the rounded value. As the data being sampled continually, the current FL system faces the following challenges in terms of convergence performance
\begin{definition}[Long-tailed learning]\label{def1:long_tail_dis}
    The lower skip values $k_u$ result in the long-tailed distribution of old data. When new data is sampled in the system, it is hard for the data to contribute to the entire FL system, resulting in the ignorance of the learning on new data samples.
\end{definition}
\begin{definition}[Data shortage]
\label{def2:data_shortage}
    The higher skip values $k_u$ result in data shortages as the total number of available data $D_u=\lfloor T/k_u \rceil$ is limited. Thus, reducing the FL generalization.
\end{definition}
It is crucial to emphasize that selecting $k_u$ values that are too high or too low can substantially degrade the performance of the FCL system. Thus, identifying an optimal $k_u$ selection strategy is essential for achieving robust performance in RTS systems.
To understand the drawbacks of current FL works in FCL concept more clearly, we take \eqref{eq:FL-problem} and \eqref{eq:global-acc} from \cite{2020-FL-EnergyEfficientFL} into consideration. The joint loss function $\mathcal{L}(w^{(n)})$ is considered on the ideal dataset (i.e., the dataset is not affected by external factors such as sampling delay that cause the non-IID). When using the FCL system with a non-ideal dataset, we are \emph{unable to gauge how non-ideal factors affect FCL convergence accurately}. To address this problem and enable optimization for the FCL system with the practical test dataset, we take into account the concept of the \textit{generalization gap}, a metric that measures the decrease in AI performance when the data suffers from the distribution shift \cite{2023-FL-DistributionShift}. 

\section{SCFL: Generalization Gap based Convergence Analysis and Resource Allocation}\label{sec:ResourceAllocation}
In this section, we first investigate how the sampling process affects the FCL training performance (i.e., how the sampling affects the overfitting). {\color{black}Accordingly, we propose a sample-driven optimization framework designed to minimize network energy consumption while mitigating the effects of long-tailed learning and data shortages in FCL.}
\subsection{Generalization Gap of Decentralized Model}
Given the FCL model as demonstrated in Section~\ref{sec:FL-model}, we consider the distribution of dataset $Z_u$ on user $u$ as $\mathbb{P}_u = \mathbb{P}(Z_u)$. 
To measure the FCL performance when validated on the test dataset compared to that when training on the training dataset, we leverage the terminology of \emph{generalization gap}. Specifically, the generalization gap measures the difference between the training loss $\mathcal{L}(Z^{\text{train}}, w)$ (i.e., the loss received from the training process on the training data set) and the test loss $\mathcal{L}(Z^{\text{test}}, w)$ (i.e., the loss evaluated on the test set concerning the trained model on the training dataset).

\begin{definition}
The local generalization gap is determined as:
    \begin{align}
    \mathfrak{R}^{(n)}_u 
    &= \mathbb{E}_{\{x_i, y_i\} \sim \mathcal{D}_u} \left[ \mathbb{E}_{(x,y)\sim \mathcal{D}^{\mathrm{train}}_{u}} \left[\ell(x,y;w^{(n)}_u) \right] - \right. \notag \\    & \left. ~~~~~~~~~~~~~~~~~~
    \mathbb{E}_{(x,y)\sim \mathcal{D}^{\mathrm{test}}}\left[\ell(x,y;w^{(n)}_u) \right] \right],
\end{align}
where $\mathbb{E}(\cdot)$ represents the expectation over the specific distribution. $w^{(n)}_u$ is the model parameters of user $u$ and defined by $w^{(n)}_u = w^{(n)} + \vartheta^{(n)}_u$. $\mathcal{D}^{\mathrm{train}}_u$ represents the empirical train data set of user $u$ in the RTS system. In practice, $\mathcal{D}^{\mathrm{train}}$ can be empirically considered as the combination of local datasets, and is demonstrated as $\mathcal{D}^{\mathrm{train}} = \{ \mathcal{D}^{\mathrm{train}}_u\}^{U}_{u=1}$.
\end{definition}
And thus, we have the following global generalization gap.
\begin{lemma}\label{lemma:global-generalization-gap}
    The global generalization gap is determined as:
    \begin{align}
        \mathfrak{R}_{\mathrm{glob}}=\frac{1}{U}\sum_{u=1}^U \mathfrak{R}_{u}.
    \end{align}
\end{lemma}
\begin{proof}
    The proof is demonstrated in Appendix~\ref{appendix:global-generalization-gap}.
\end{proof}

\subsection{Generalization Gap and Local Sampling}
In FCL, to consider the sampling problem induced by Definition~\ref{def1:long_tail_dis} and Definition~\ref{def2:data_shortage}, our premise is to connect the generalization gap to data sampling via the \emph{data exploration bias} \cite{2020-DL-DataOverfit}. Thus, we first adopt the following assumptions. 
\begin{assumption}[Section~II \cite{2020-DL-DataOverfit}] \label{ass:sampling-rule}
    Given the global data pool $\mathcal{D}$, the data from each user $u$ can be considered to be sampled before the FCL sampling rule $s \in \mathcal{S}$.
    \begin{align}
        \mathbb{P}(X \vert x \in \mathcal{D}_u, u = \{1,2,\dots, U\} ) = \mathbb{P}(\mathcal{D}\vert s).
    \end{align}
\end{assumption}
When considering the sampling rule by the sampling delay $\tau_u$ of the data sensing on each user $u$, we have the following:
\begin{assumption}[Section~V.G~\cite{2020-DL-DataOverfit}]
    The FCL sampling rule on user $u$ from the limitless stream data pool is time-homogeneous and stationary with stationary distribution $\pi$ and satisfies a uniform mixing condition
    \begin{align}
        \max_{x_u} \mathrm{D_{KL}}(\mathbb{P}(x_{u,s} \vert x_{u,s-1})\Vert \pi) \leq c_0 e^{-c_1\tau_u}.
    \end{align}
    Here, $\tau_u$ represents the time delay between two sampled consecutive data points on the IoT device of user $u$, $\mathrm{D_{KL}}(\cdot\Vert \cdot)$ represents the KL divergence between two distributions.
\label{ass:mdp-sampling}
 \end{assumption}
Assumption~\ref{ass:sampling-rule} reveals a sampling mechanism in an IoT network. Specifically, in the IoT network, each device's set of sampling rules $s \in \mathcal{S}$ has different characteristics (e.g., different data distribution $\pi$, sampling rate, and data characteristics). The specific characteristics that influence the sampling rule for user $u$ result in a divergence between the dataset sampled by user $u$ and the user with the ideal sampling rule. 
%

Meanwhile, Assumption~\ref{ass:mdp-sampling} represents the relationship between the IoT device's sampling rule can be considered as a Markov Decision Process (MDP). 
Thus, the current sampling distribution depends on the last sampled data. 

In FCL, the limitless stream data pool can be conceptualized as an ideal dataset $\Tilde{\mathrm{Z}}$. Consequently, $\mathrm{D_{KL}}(\mathbb{P}(x_{u,s} \vert x_{u,s-1})\Vert \pi)$ can be expressed as the mutual information $I(p(z|\Tilde{\mathrm{Z}}),p(z|\mathrm{Z}))$ between the data distribution sampled from the actual dataset $p\left(z|\mathrm{Z}\right)$ and that of $p\left(z|\Tilde{\mathrm{Z}}\right)$.
Using $I(p(z|\Tilde{\mathrm{Z}}),p(z|\mathrm{Z}))$, we can formulate the problem in Defintion~\ref{def1:long_tail_dis} and Definition~\ref{def2:data_shortage} into the generalization problem of FCL convergence through mutual information. By leveraging Assumption~\ref{ass:mdp-sampling} we have definition of mutual formation depend on $k$ skipped samples as
\begin{align}
    I\left(p\left(z|\Tilde{\mathrm{Z}}\right),p\left(z|\mathrm{Z}\right)\right)
    = c_0 e^{-c_1\times k\times\tau_u},
\label{eq:information-usage}
\end{align}
where $\tau_u$ is the last sampled time at user $u$, $c_0$ is defined as the subtraction entropy of mutual information and data entropy, $c_1$ is the time-variant coefficient, which shows the transition rate of data. There exists mutual information $I(\cdot)$ between any two data samples within the data sampling period. Consequently, when sampling data over a brief period, the sampled data volume is relatively small. Utilizing regression in the machine learning process, the $c_0$, and $c_1$ values of the mutual information and time-variant for this dataset can be easily calculated. Based on regression, we provide two parameters $c_0$, $c_1$ during the simulation to determine the dataset for the SFCL algorithm. 

{\color{black}As observed from Eq.~\eqref{eq:information-usage}, as the sampling wait time between two samples is larger, the mutual information between two samples reduces. Therefore, \emph{the data becomes more IID with others}. This phenomenon enhances the generalization of the user's learning, leading to a significant improvement in FCL learning. However, if the sampling delay exceeds the threshold defined by the data shortage limitation in Definition~\ref{def2:data_shortage}, the amount of data collected from each IoT device decreases, \emph{leading to information loss}.
Consequently, it is essential to balance the sampling computation efficiency in the RTS system with the preservation of meaningful information in the collected dataset.
}

\subsection{Convergence Analysis Under-sampling Control Process}
\label{sec:convergence-analysis}
To analyze the convergence rate of SCFL, we first make the following assumptions on the loss function:
\begin{assumption}[$L$-Lipschiz Smoothness]\label{eq:YangA1}
    Each local objective function is Lipschitz smooth, that is, $\Vert \nabla \mathcal{L}_u(x) - \nabla \mathcal{L}_u(y)\Vert \leq L\Vert x-y \Vert, ~\forall x,y \in \mathbb{R}^d, \forall u \in \{1,2,\dots, U\}$.
\end{assumption}
\begin{assumption}[$\mu$-Strongly Convex]\label{eq:YangA2}
    Each local objective function is strongly convex, that is, $\Vert \nabla \mathcal{L}_u(x) - \nabla \mathcal{L}_u(y)\Vert \geq \mu\Vert x-y \Vert, ~\forall x,y \in \mathbb{R}^d, \forall u \in \{1,2,\dots,U\}$. 
\end{assumption}
\textcolor{black}{
Assumptions~\ref{eq:YangA1} and \ref{eq:YangA2} are commonly employed in the standard analysis of SGD \cite{2020-FL-MiniBatchSGD}. These assumptions are readily satisfied by widely used FL \cite{2021-FL-FedNova, 2020-FL-FedProx, 2020-FL-JointLearning-CommunicationsFramework, 2020-FL-EnergyEfficientFL}.} Motivated from \cite{2020-FL-PerFedAvg}, we introduce two bounds on unbiased characteristics of Gradient and Hessian.
\begin{lemma}[Unbiased Gradient and Bounded Variance] \label{lemma:unbiased-gradient}
    Given local objective functions, the stochastic gradient at each client with an arbitrary sampling rule is an unbiased estimator of the global gradient $\nabla\tilde{\mathcal{L}}(\cdot)$ and has bounded variance:
    \begin{align}
        &\nabla\tilde{\mathcal{L}}\left(\omega^{(n)}\right) - \nabla \mathcal{L}\left(\omega^{(n)}\right)\leq \\
        &\|\vartheta^{(n)}_u\|2^{H(\mathrm{Z})-H(\tilde{Z})}\sqrt{2\left[H\left(p\left(z|\mathrm{Z}\right)\right) - I\left(p\left(z|\Tilde{\mathrm{Z}}\right),p\left(z|\mathrm{Z}\right)\right)\right]}.\notag
    \end{align}
\end{lemma}
\begin{proof}
    The proof is demonstrated in Appendix~\ref{appendix:unbiased-gradient}.
\end{proof}

\begin{lemma}[Unbiased Hessian and Bounded Variance] \label{lemma:unbiased-hessian}
    Given local objective functions, the stochastic Hessian at each client with an arbitrary FCL sampling rule is an unbiased estimator of the global gradient $\nabla^2\tilde{\mathcal{L}}(\cdot)$ and has bounded variance: 
    \begin{align}
        &\nabla^2\tilde{\mathcal{L}}\left(\omega^{(n)}\right) - \nabla^2 \mathcal{L}\left(\omega^{(n)}\right)\\
        \leq &L2^{H(\mathrm{Z})-H(\tilde{Z})}\sqrt{2\left[H\left(p\left(z|\mathrm{Z}\right)\right) - I\left(p\left(z|\Tilde{\mathrm{Z}}\right),p\left(z|\mathrm{Z}\right)\right)\right]}.\notag
    \end{align}
\end{lemma}
\begin{proof}
    The proof is demonstrated in Appendix~\ref{appendix:unbiased-hessian}.
\end{proof}

Subsequently, we derive the following theorem.
\begin{theorem}[Theorem on inductive generalization gap]
    Due to the influence of the generalization gap, the value of the validation loss changes after global iterations. The number of global iterations required to achieve the global accuracy $\varrho$ with sampling rule $\mathcal{T}$ can be represented as 
\begin{equation}
\label{eq:I_glob}
   I_\mathrm{glob}(\tau_u,\varpi) \geq \ln{\left(\frac{1}{\varrho}\right)}\frac{2 UL^2 \xi}{\left[\xi\left(L+2\right)\Psi+\frac{\xi L}{U}-\varpi\mu\right]},
\end{equation}
where $\Psi$ is the generalization gap statement and is demonstrated in Equation~\eqref{eq:psi} and is demonstrated as
\begin{align}
    \Psi = 2^{H(\mathrm{Z})-H(\tilde{Z})}\sqrt{2\left[H\left(p\left(z|\mathrm{Z}\right)\right) - I\left(p\left(z|\Tilde{\mathrm{Z}}\right),p\left(z|\mathrm{Z}\right)\right)\right]}, \notag
\end{align}
and $I\left(p\left(z|\Tilde{\mathrm{Z}}\right),p\left(z|\mathrm{Z}\right)\right)$ is a function of $k_u$ as in Eq.~\eqref{eq:information-usage}.
\end{theorem}
\begin{proof}
     The proof is demonstrated in Appendix~\ref{appendix:inductive-generalization-gap}: 
\end{proof}
The proposed theorem offers an alternative method for assessing the overall communication rounds of the FCL system when distributed devices $u \in  \{1,2,\dots, U\}$, are influenced by non-ideal data. This non-ideal data arises due to variations in sampling times represented as $k_u\times \tau_u$. 
By controlling $k_u$, we can reduce the generalization gap of the FCL system, thus, indirectly deal with the long-tailed learning (Def.~\ref{def1:long_tail_dis}) and limited data availability (Def.~\ref{def2:data_shortage}).
Subsequently, we can establish a unified problem formulation for the FCL system specifically designed for deployment in RTS IoT networks.
\subsection{Problem Formulation}
\label{sec:problem-formulation}
In this work, we aim to investigate an optimization problem of minimizing the overall energy usage of all the users, taking into account the fluctuations in their settings within specified resource and time constraints. We denote $\mathbf{t}=\left[t^\mathrm{trans}_1, \ldots, t^\mathrm{trans}_U\right]^\top$, $\mathbf{b}=\left[b_1, \ldots, b_U\right]^\top$, $\mathbf{f}=\left[f_1, \ldots, f_U\right]^\top$, $\mathbf{p}=\left[p_1, \ldots, p_U\right]^\top$, $\mathbf{k}=\left[k_1, \ldots, k_U\right]^\top$, $A_u=\frac{2}{(2-L\delta)\delta\mu} C_u D_u$, where $\top$ is the transpose operator. The subscriptions $f_u^{\max}$ and $p_u^{\max}$ are respectively the maximum local computation capacity and maximum value of the average transmit power of user $u$, respectively. We can express the optimization problem of energy consumption as follows:

\begin{subequations}
	\label{subeqn-opt-pro-general:opt-pro-main}
	\begin{alignat} {3}
		& \min_{\mathbf{t}, \mathbf{b}, \mathbf{f}, \mathbf{p}, \varpi, \textbf{k}} 
		&	&  I_\mathrm{glob}(k_u, \varpi) \sum_{u = 1}^{U} \left[ E_u^C(\mathbf{f})+ E_u^S(k_u) + E_u^T(\mathbf{t}, \mathbf{b},\mathbf{p}) \right],
		\label{opt-pro-general:opt-pro}\\
		& ~~~~\text{s.t.}
		&	& I_\mathrm{glob}(k_u, \varpi)\left(\frac{A_u \log_2 (1/\varpi)}{f_u} + t^\mathrm{trans}_u\right) \notag \\
            &   &   &  \quad\quad\quad\quad\quad\quad\quad\quad\quad \leq T_{\mathrm{max}},~\forall u \in \mathcal{U},
            \label{subeqn-opt-pro-general:time-transmission-constraint}\\
		&   &   & t^\mathrm{trans}_u b_u \log_2 (1 + \frac{g_u p_u}{n_0 b_u}) \geq D_{0}, \forall u \in \mathcal{U},
		\label{subeqn-opt-pro-general:min-bandwidth-constraint}\\ 
		&   &   & 0 < f_u \leq f^{\mathrm{max}}_u, \forall u \in \mathcal{U},
        \label{subeqn-opt-pro-general:computation-rate} \\	
		&   &   & 0 < p_u \leq p^{\mathrm{max}}_u, \forall u \in \mathcal{U},
		\label{subeqn-opt-pro-general:power} \\
        &   &   & \sum\nolimits^U_{u = 1}b_u \leq B, \forall u \in \mathcal{U},
		\label{subeqn-opt-pro-general:bandwidth} \\	
		&   &   & 0 < \varpi < 1,
		\label{subeqn-opt-pro-general:learning-rate}\\
		&   &   & t^\mathrm{trans}_u \geq 0, b_u \geq 0, \forall u \in \mathcal{U},
		\label{subeqn-opt-pro-general:bandwidth-transmissiontime} \\
		&   &   & k_\mathrm{l\_thres} \leq k_u \leq k_\mathrm{u\_thres}, \forall u \in \mathcal{U},
		\label{subeqn-opt-pro-general:sample-control}
    \end{alignat}
\end{subequations}

where the completion time constraint in~\eqref{subeqn-opt-pro-general:time-transmission-constraint} is obtained by substituting $I_u$ from~\eqref{Eq:10} and $I_\mathrm{glob}(k_u, \varpi)$ from~\eqref{eq:I_glob}. The data transmission constraint in~\eqref{subeqn-opt-pro-general:min-bandwidth-constraint} is derived to the achievable rate in~\eqref{eq:transmission-rate}. \eqref{subeqn-opt-pro-general:computation-rate},~\eqref{subeqn-opt-pro-general:power} and~\eqref{subeqn-opt-pro-general:bandwidth} denote the constraints of CPU frequency, transmit power, and bandwidth allocation, respectively. Constraint~\eqref{subeqn-opt-pro-general:learning-rate} is the feasible range for local accuracy,~\eqref{subeqn-opt-pro-general:bandwidth-transmissiontime} indicates the feasible values of transmission time and bandwidth, and~\eqref{subeqn-opt-pro-general:sample-control} is the constraint of the number of skipped sample with lower boundaries $k_\mathrm{l\_thres}$ and upper boundaries $k_\mathrm{u\_thres}$.
\subsection{Relaxing the Problem Formulation}
To be more specific, due to the fixed multiplication between the number of communication rounds $I_\mathrm{glob}(k, \varpi)$ and total energy of each round $\sum_{u = 1}^{U} \left[ E_u^C(\mathbf{f})+E_u^T(\mathbf{t}, \mathbf{b},\mathbf{p}) \right]$, the proposed optimization problem is only feasible when the total energy function remains variant among communication rounds. This also means that the FCL network is assumed to be homogeneous (e.g., the users' positions remain unchanged, and the channel model is static). To alleviate this issue, we reformulate the problem \eqref{subeqn-opt-pro-general:opt-pro-main} as follows: 
	\begin{alignat} {3}	
		& \min_{\mathbf{t}, \mathbf{b}, \mathbf{f}, \mathbf{p}, \varpi, k} 
		&	 &  \sum^{I_\mathrm{glob}(\tau, \varpi)}_{i=1} \sum_{u = 1}^{U} \left[ E_{u,i}^C(\mathbf{f})+E_{u,i}^T(\mathbf{t}, \mathbf{b},\mathbf{p}) \right],
        \label{opt-pro-general2:opt-pro} \\
		& ~~~~\text{s.t.}
		&	&  \eqref{subeqn-opt-pro-general:time-transmission-constraint}, 
                   \eqref{subeqn-opt-pro-general:min-bandwidth-constraint},
                   \eqref{subeqn-opt-pro-general:computation-rate},
                   \eqref{subeqn-opt-pro-general:power},
                   \eqref{subeqn-opt-pro-general:bandwidth},
                   \eqref{subeqn-opt-pro-general:learning-rate},
                   \eqref{subeqn-opt-pro-general:bandwidth-transmissiontime},
                   \eqref{subeqn-opt-pro-general:sample-control},\notag
\end{alignat}
where the main optimization problem is to find the minimum total energy over $I_\mathrm{glob}(\tau, \varpi)$ rounds. 
However, by solving the problem solely under constraint \eqref{subeqn-opt-pro-general:time-transmission-constraint} is NP-hard, and it is hard to be solved efficiently. 
To efficiently solve the problem, we first relax the \eqref{subeqn-opt-pro-general:time-transmission-constraint} into the more simple and straightforward constraint. Specifically, instead of finding the total transmission time on every user as mentioned in \eqref{subeqn-opt-pro-general:time-transmission-constraint}, we instead find the solution for $t^\mathrm{trans}_u$ that satisfies the maximum time out on every round $T_{\mathrm{max-round}}$: 
\begin{lemma}[]
    The maximum total timeout $T_\textrm{max}$ is always greater than that of the transmission time of the system with the optimized number of FCL communication rounds $I_\mathrm{glob}(k, \varpi)$.
    \begin{align}
        T_{\max} 
        &\geq I_\mathrm{glob}(k_u, \varpi)\times\left(\frac{A_u \log_2 (1/\varpi)}{f_u} + t^\mathrm{trans}_u\right)\\\notag
        &\geq 
        \underset{\mathbf{\tau}, \mathbf{\varpi}}{\min}~I_\mathrm{glob}(k_u, \varpi)\times \left(\frac{A_u \log_2 (1/\varpi)}{f_u} + t^\mathrm{trans}_u\right).
    \end{align}
\label{lemma:worst-case-time}    
\end{lemma}
\vspace{-0.5cm}
Leveraging Lemma~\ref{lemma:worst-case-time}, we set the maximum timeout for transmission at each epoch to be $T_{\mathrm{max-round}}=T_\textrm{max}/\widetilde{I}_\mathrm{glob}(k_u, \varpi)$, where $\widetilde{I}_\mathrm{glob}(k_u, \varpi)$ represents the total number of communication rounds observed in the previous episode when the problem was not optimized. This strategy allows for the progressive optimization of $t^\mathrm{trans}_u$ and $I_\mathrm{glob}(k_u, \varpi)$. Consequently, we can find improved solutions for  $t^\mathrm{trans}_u$ by gradually reducing $T_{\mathrm{max-round}}$. To achieve this, we simplify the overall energy minimization problem as: 
\begin{subequations}
	\label{subeqn-opt-pro-general3:opt-pro-main}
	\begin{alignat} {3}
		& \min_{\mathbf{t}, \mathbf{b}, \mathbf{f}, \mathbf{p}}
		&	 &  \sum_{u = 1}^{U} \left[ E_{u}^C(\mathbf{f})+E_{u}^T(\mathbf{t}, \mathbf{b},\mathbf{p}) \right],
        \label{opt-pro-general3:opt-pro} \\
		& ~~\text{s.t.}
		&	&      \eqref{subeqn-opt-pro-general:min-bandwidth-constraint},
                   \eqref{subeqn-opt-pro-general:computation-rate},
                   \eqref{subeqn-opt-pro-general:power},
                   \eqref{subeqn-opt-pro-general:bandwidth},
                   \eqref{subeqn-opt-pro-general:learning-rate},
                   \eqref{subeqn-opt-pro-general:bandwidth-transmissiontime},
                   \eqref{subeqn-opt-pro-general:sample-control},\notag \\
        &   &	& \frac{A_u \log_2 (1/\varpi)}{f_u} + t^{\mathrm{trans}} \leq T_{\mathrm{max-round}}, \forall u \in \mathcal{U}, \label{subeqn-opt-pro-general3:time-transmission-constraint} \\
        &   &   & I_\mathrm{glob}=\arg\min_{\mathbf{\tau}, \mathbf{\varpi}} \ln{\left(\frac{1}{\varrho}\right)}\frac{2 UL^2 \xi}{\left[\xi\left(L+2\right)\Psi+\frac{\xi L}{U}-\varpi\mu\right]}.
		\label{subeqn-opt-pro-general:total-iterations}       
\end{alignat}
\end{subequations}
In~\eqref{subeqn-opt-pro-general3:time-transmission-constraint}, we convert the maximum total time for the FCL process $T_{\mathrm{max}}$ to $T_{\mathrm{max-round}}$. By doing so, we transform the problem into two interdependent problems.

By manipulating the number of samples to be skipped $k_u$, the learning becomes inconsistent as the global iterations are not predefined at the beginning of the learning state, thus making the optimization problem more complicated. This complexity renders traditional optimization methods inadequate for solving resource allocation in SCFL systems within IoT networks. DRL algorithms have demonstrated their ability to solve previously intractable policy-making problems characterized by high-dimensional states and action spaces. Building upon these observations, we are inspired to leverage DRL algorithms to tackle the resource allocation problem in the SCFL system.

\section{A Deep Reinforcement Learning Approach for SCFL}\label{sec:DRL_SCFL}
In this section, we provide a concise overview of the structures and processes involved in our algorithm based on A2C, namely A2C-EI. To effectively employ DRL for solving the minimization problem, it is crucial to adjust the optimization problem to align with the operational principles of DRL.

\subsection{Closed-form Expression for Optimization Problem}
In order to address the optimization problems~\eqref{opt-pro-general:opt-pro} using DRL, it is essential to determine the closed-form representations of these problems. To achieve this, we begin by categorizing the constraints in the optimization problem into two distinct groups: ECS and ICS. Consequently, we integrate the ECS design into the A2C architecture, which can be named A2C-EI, via the reconfiguration of the network output as follows.

\textit{1) Explicit constraints:}
The term ``explicit constraints" refers to constraints that are straightforward to configure. These constraints primarily involve setting limits on system variables such as bandwidth, computation capacity, and transmit power. Leveraging the outputs of a deep model can significantly enhance these components without incurring substantial computational overhead. Specifically, we can describe the constraints in an alternative manner by utilizing the fundamental activation function. This approach allows for significant improvements in undertaking ECS to reduce computational complexity.
\begin{enumerate}[label=(\roman*)]
    \item In order to confine the variables within a specific range (e.g. $0 < p_u \leq p^{\text{max}}_u$), we employ the sigmoid function as the initial step. Subsequently, the data is normalized using the appropriate normalization function to ensure it falls within the desired range.
    \item To enforce a limit on the value and ensure that the variables consistently sum up to a value below a predefined threshold (e.g. $0 < \sum^U_{u = 1}b_u \leq B$), we utilize the softmax function. This allows us to appropriately constrain the variables while preserving their relative proportions. Following this, we can employ a normalization process, similar to the one described in (i), to further refine the variables and ensure they adhere to the desired range.
\end{enumerate}
Implementing this strategy can simplify the loss function and enhance the performance of the DRL algorithm. Instead of directly embedding constraints into the reward function, the constraints are rigidly enforced, ensuring that variables automatically adhere to the required boundaries. This simplification of the loss function avoids the complexities introduced by constraints and mitigates the stronger non-convex nature of the loss function.

Furthermore, this approach improves the adaptability of the model, enabling it to rapidly learn and adjust to new constraints. By reducing computational complexity and enhancing system adaptability, we can optimize performance without compromising result quality. Moreover, this technique significantly boosts the DRL algorithm's performance by facilitating swift learning and adaptation to new constraints. Through the reduction of computational complexity and the increase in system adaptability, we can optimize the system's performance without impacting result quality.

\textit{2) Implicit constraints:} Unlike explicit constraints, ``implicit constraints" are more challenging to configure and cannot be resolved solely by normalizing the model's output. To address this issue, we employ the penalty approach \cite{2021-RL-Noma-DDPG} as a means to tackle these constraints effectively.

Utilizing the actor architecture output configuration through ECS, our algorithm is straightforward, easy to implement, and seamlessly compatible with various DRL approaches. Notably, our approach consistently outperforms conventional DRL methods that do not employ the Implicit-Explicit Constraints (EI) configuration, which will be demonstrated in detail in the following section.
\subsection{Designing Observation for SCFL}\label{subsec:Ob_SCFL}
\textit{1) State Space:} We denote $g^t_u$ is the channel gain between ES and the user $u$ at the time step $t$. Then the  state space of the FCL system, denoted by $\mathbf{S}$:
\begin{equation}
    \mathbf{s}^t = \left\{g^t_1,g^t_2,\ldots, g^t_U \right\}.
\end{equation}
As a consequence, the state space $\mathbf{S}$ can be defined as $\mathbf{S}=\{\mathbf{s}^1,\ldots, \mathbf{s}^{T_{\mathrm{step}}}\}$, where $T_{\mathrm{step}}$ is the total number of time step used for the RL training phase.

\textit{2) Action Space:} Let $\mathbf{A}$ denote the action space of the system. Given a certain state $\mathbf{s}$, a control action is perfomed to determine  $\mathbf{t}=\left[t^{\mathrm{trans}}_1, \ldots, t^{\mathrm{trans}}_U\right]^\top$ as the set of transmission time of user, $\mathbf{b}=\left[b_1, \ldots, b_U\right]^\top$ as the set of user's bandwidth, $\mathbf{f}=\left[f_1, \ldots, f_U\right]^\top$ is the set of user's computation capacity and $\mathbf{p}=\left[p_1, \ldots, p_U\right]^\top$ as the set of average transmission power of user $u$. Thus, the action space can be defined as 
\begin{align}
    \mathbf{A}&=\{\mathbf{a}_1,\ldots,\mathbf{a}_U\}\\ \notag
    &=\{t^{\mathrm{trans}}_1, b_1, f_1, p_1,\ldots,t^{\mathrm{trans}}_U, b_U, f_U, p_U\}.
\end{align}

\textit{3) Reward Space:}
Considering the FCL system model, our objective is to optimize the overall energy consumption while preserving the AI performance through local computation. This includes the execution of AI models during task execution, task training, and transmission of local parameters to ES. Intending to minimize total energy consumption over time, we adopt the total energy consumed as the reward metric.

In order to achieve this objective, we aim to establish an upper bound on the completion time, ensuring timely completion. Additionally, we impose constraints on the transmission time to prevent excessive divergence in total energy consumption beyond a predetermined threshold. These constraints are utilized as penalties in our optimization framework. Therefore, the reward can be defined as follows:
\begin{align}
    \mathbf{R} = \{\mathbf{r}^1 (\mathbf{s}^1,\mathbf{a}^1), \ldots, \mathbf{r}^T (\mathbf{s}^T,\mathbf{a}^T)\},
\end{align}
where the immediate reward $\mathbf{r}^t$ as
\begin{align}
    \mathbf{r}^t (\mathbf{s}^t,\mathbf{a}^t)= - \sum_{u = 1}^{U} \left( E_u^C+E_u^S+E_u^T\right) + \lambda_1\mathcal{P}_1 + \lambda_2\mathcal{P}_2,
\label{eq:temporal-reward}
\end{align}
where coefficients $\lambda_1, \lambda_2$ are utilized as controls for the regularization. The penalties, denoted as $\mathcal{P}_1$ and $\mathcal{P}_2$, are defined as constraints on the total energy consumption, accounting for the variations in the user's settings over time. The core concept behind these constraints is that if a specific value surpasses a predetermined threshold, the AI model aims to minimize the deviation from that threshold. Therefore, the penalty for the completion time of each global round of computing and transmission tasks for users is defined as follows:
\begin{align}
    \mathcal{P}_1 = \max\left\langle\left[\left(\frac{A_u \log_2 (1/\varpi)}{\mathcal{L}_u} + t^{\mathrm{trans}}\right) - T_{\mathrm{max-round}}\right], 0 \right\rangle ,
\end{align}
and the penalty function for constraints on the transmission time:
\begin{align}
    \mathcal{P}_2 = \max\left\langle\left[s - t^{\mathrm{trans}}_u b_u \log_2 \Big(1 + \frac{g_u p_u}{N_0 b_u}\Big)\right], 0 \right\rangle ,
\end{align}
where the $\max\langle\cdot,\cdot\rangle$ function emulates the resilience property by imposing a penalty only when the round completion time exceeds the upper threshold.

Equation~\eqref{eq:temporal-reward} only considers the ICS while neglecting the ECS. Despite the simplified reward function, A2C-EI's learning process can guarantee better convergence compared to traditional A2C. To substantiate our theoretical analysis of the A2C-EI approach, we have established definitions for the boundaries of the action components within the A2C-EI observation's reward section as follows:
\begin{definition}\label{def:ECS_x}
    Given $P_{\mathrm{ECS},i}$ as the penalty for the action's values $x_i$ according to ECS and come over the boundary:
    \begin{align}
    P_{\mathrm{ECS},i} &= \abs{\max\left\langle x_i - x_i^{\mathrm{max}}, 0 \right\rangle} \notag \\
                       &+ \abs{\max\left\langle x_i^{\mathrm{min}} -x_i, 0 \right\rangle} \geq 0,
    \end{align}
    where $i$ represents the index of different ECS constraints.
\label{def:ECS-P}
\end{definition}
From Definition~\ref{def:ECS-P}, we have the following lemma.
\begin{lemma}
    Given the EI design, the action values $x_i$ always have $x_i^{\mathrm{min}} < x_i \leq x^{\mathrm{max}},~\forall i$, so that penalty on ECS $P_{\mathrm{ECS-EI},i} = 0,~ \forall x$. 
\label{lemma:ECS-P}
\end{lemma}
\textit{Proof}. The proof is demonstrated in Appendix~\ref{appendix:ECS_x}.
\begin{theorem}
Any DRL approach with the integrated EI design (DRL-EI) consistently yields higher optimal rewards compared to the DRL approach without the EI design. In other words,
    \begin{align}
        \mathbf{r}^*_\mathrm{DRL-EI}(\mathbf{s}^t, \mathbf{a}^t) \geq \mathbf{r}^*_\mathrm{DRL}(\mathbf{s}^t, \mathbf{a}^t),~\forall \mathbf{s} \in \mathbf{S},~\forall \mathbf{a} \in \mathbf{A},
    \end{align}
where the $\mathbf{r}^*_\mathrm{DRL-EI}, \mathbf{r}^*_\mathrm{DRL}$ represents the reward of the DRL approach with and without integrated EI design, respectively.
\label{theorem:ECS_reward}
\end{theorem}
\textit{Proof}. The proof is demonstrated in Appendix~\ref{appendix:ECS_reward}. 

Theorem~\ref{theorem:ECS_reward} demonstrates that the EI design enhances the learning performance of any DRL approach. Consequently, with the application of A2C-EI, we can consistently attain superior performance compared to the conventional A2C method.
Given the immediate reward defined as in Equation~\eqref{eq:temporal-reward}, we have the accumulative long-term reward for the system, which can be expressed as follows:
\begin{align}
    \mathcal{R}(\pi) = \sum^{I_\mathrm{glob}}_{t} \gamma^{I_\mathrm{glob}-t}\mathbf{r}^t (s^t,\pi(s^t)),
\label{eq:accumulative-reward}
\end{align}
where $\mathcal{R}(\pi)$ is the accumulative long-term reward of the agent under policy $\pi(\cdot)$ and $\mathbf{r}^t(s^t, \pi(s^t))$ at time step $t$. The subscript $\gamma$ denotes the discount factor for the reward that reflects how much the reward depends on the past performance (i.e., when $\gamma$ is near $0$, the policy evaluation ignores the historical performance, and vice versa). As we can see from Eq.~\eqref{eq:accumulative-reward}, the accumulative reward function is similar to the main optimization in Eq.~\eqref{opt-pro-general:opt-pro}. Therefore, applying DRL to our relaxed optimization problem in Eq.~\eqref{opt-pro-general3:opt-pro} also solves the original optimization problem. Thus, our DRL approach can solve the optimization problem under the FCL system with heterogeneous settings, while retaining feasibility as in \cite{2020-FL-EnergyEfficientFL}. Furthermore, the state space $\mathbf{S}$ contains only one communicating class, i.e., from a given state the process can go to any other states after $t$ steps. In other words, the MDP with states $\mathbf{S}$ is irreducible. Thus, for every $\pi$, the average throughput $\mathcal{R}(\pi)$ is well defined and does not depend on the initial state. 

\subsection{Proposed A2C-EI Algorithm} \label{sec:A2C-SCFL}
\begin{figure}[t]
\centering
\includegraphics[width =1\linewidth]{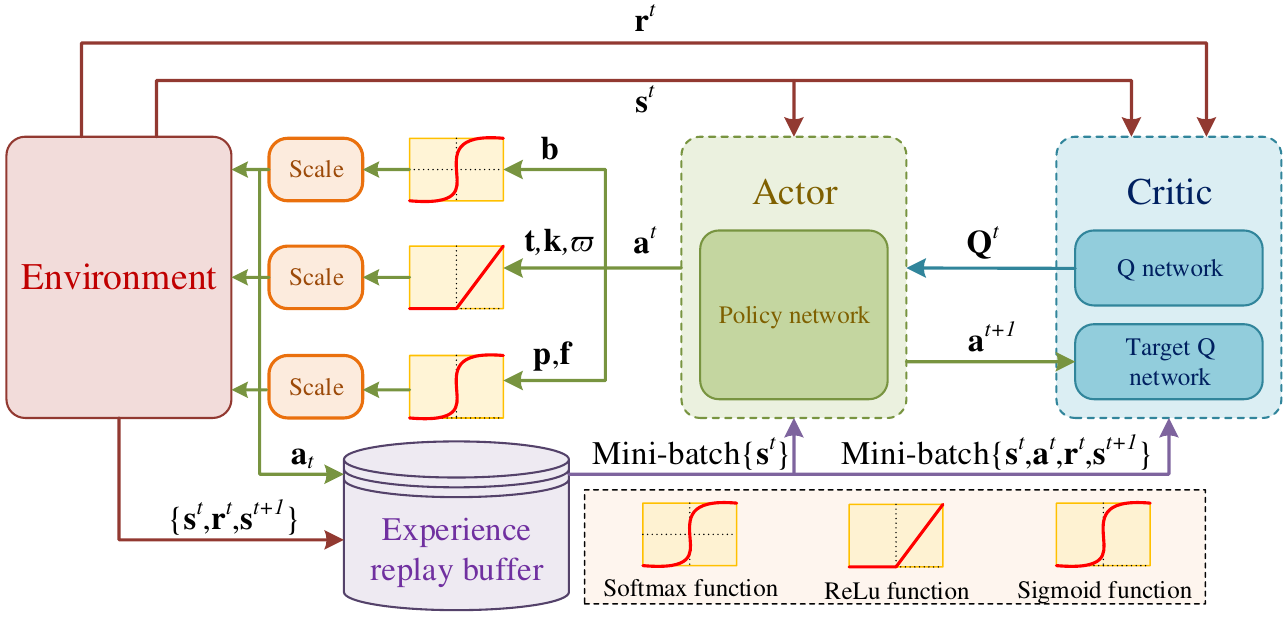}
\caption{\textbf{A2C-EI} Architecture. The main process builds on the \textbf{A2C} algorithm, which has two neural networks: \textit{Actor} (policy function) and \textit{Critic} (value function). The actor-critic receives $\mathbf{s}^t$ and $\mathbf{a}^t \sim \pi_{\theta}(\cdot|\mathbf{s}^t)$ from \textit{Environment}. Then \textit{Actor} calculates the next action $\mathbf{a}^{t+1}$ using \textit{Policy network} and sends it to the \textit{Critic}. \textit{Critic} computes the Q-function and sends it back to \textit{Actor}. The action from \textit{Actor} is then scaled using our \textbf{EI} functions: \textbf{E}CS (\textit{Sigmoid function} and \textit{Softmax function}) for exploration-exploitation and \textbf{I}CS (\textit{ReLU function}) for reduced computational complexity. The scaled action is performed in the environment and stored in the \textit{Experience replay buffer} $\mho$. \textit{Actor} sample minibatches $\{\mathbf{s}^t\}$ and \textit{Critic} sample minibatches $\{\mathbf{s}^t,\mathbf{a}^t,\mathbf{r}^t,\mathbf{s}^{t+1}\}$ from $\mho$ to iteratively update their networks until convergence.}
\label{fig:A2C}
\end{figure}
\begin{algorithm}[t]
    \caption{{Proposed A2C-EI Algorithm}}
    \label{alg:A2C-EI}
\begin{algorithmic}
    \State {\bfseries Require:} Initialize policy parameters $\theta$, Q-network parameter $\phi$, target Q-network weights $\phi_{\mathrm{targ}}$, experience replay buffer $\mho$, the total number of time steps for the RL training phase $T_{\mathrm{step}}$.
\State{\bfseries Repeat}
    \While{$t \leq T_{\mathrm{step}}$}
    \State Observe state $\mathbf{s}^t$ and select action $\mathbf{a}^t \sim \pi_{\theta}(\cdot|\mathbf{s}^t)$.
    \State Apply \textbf{\textit{ECS}} and \textbf{\textit{ICS}} then scale $\mathbf{a}^t$.
    \State Execute $\mathbf{a}^t$ in the environment.
    \State Observe next state $\mathbf{s}^{t+1}$, and reward $\mathbf{r}^{t}$.
    \State Store $\left(\mathbf{s}^t,\mathbf{a}^t,\mathbf{r}^t,\mathbf{s}^{t+1},d \right)$ in experience replay buffer $\mho$.
    \State \textbf{If} $\mathbf{s}^{t+1}$ is terminal, reset the environment state.
    \For{$i$ in range (however many updates)} 
    \State Randomly sample a batch of transitions
    \State $\mathcal{B}=\{\left(\mathbf{s}^t,\mathbf{a}^t,\mathbf{r}^t,\mathbf{s}^{t+1},d \right)\} \in \mho$
    \State Compute targets for the Q-functions:
    \begin{align*}
         y&\left(\mathbf{r}^t,\mathbf{s}^{t+1}, d\right) 
         =\mathbf{r}^t + \gamma (1-d)\left(\min Q_{\phi_{\mathrm{targ}}}\left(\mathbf{s}^{t+1}, \tilde{\mathbf{a}}^{t+1}\right)\right.\\
         &\left.\quad- \alpha\log \pi_{\theta}(\tilde{\mathbf{a}}^{t+1}|\mathbf{s}^{t+1})\right), \quad\tilde{\mathbf{a}}^{t+1} \in \pi^{t+1}\left(\cdot \mid \mathbf{s}^t\right), 
    \end{align*}
    where $\tilde{\mathbf{a}}^{t+1}$ is highlighted that the next actions have to be sampled fresh from the policy, whereas by contrast, $\mathbf{s}^{t+1}$ and $\mathbf{r}^{t}$ come from the experience relay buffer $\mho$.
    \State Update Q-functions by one step of gradient descent
    \begin{align*}
        \nabla_{\phi}\frac{1}{|\mathcal{B}|}\sum
        \left(Q_{\phi_{\mathrm{targ}}}\left(\mathbf{s}^{t}, \mathbf{a}^{t}\right)-y\left(\mathbf{r}^t,\mathbf{s}^{t+1}, d\right) \right)^2.
    \end{align*}
    \State Update policy by one step of gradient ascent
    \begin{align*}
        \nabla_{\theta}\frac{1}{|\mathcal{B}|}\sum_{\mathbf{s}^{t}\in \mathcal{B}}
        \left(\min Q_{\phi}\left(\mathbf{s}^{t}, \tilde{\mathbf{a}}^{t}_{\theta}(\mathbf{s}^{t})\right) - \alpha\log \pi_{\theta}(\tilde{\mathbf{a}}^{t}_{\theta}(\mathbf{s}^{t})| \mathbf{s}^{t}) \right)^2.
    \end{align*}
    \State Update target networks with
    \begin{align*}
        \phi_{\mathrm{targ}} \leftarrow \rho\phi_{\mathrm{targ}} + (1-\rho)\phi,
    \end{align*}
    where $\rho\sim \beta(5,1)$, where $\beta(\cdot)$ denotes the Beta distribution.
    \EndFor
    \EndWhile
    \State{\bfseries until} convergence
\end{algorithmic}\end{algorithm}

We proposed A2C-EI, which is demonstrated in Fig.~\ref{fig:A2C} and Algorithm~\ref{alg:A2C-EI}, as a novel DRL solution for optimizing the aforementioned problem formulation. We first start with the background of A2C-EI by the A2C approach.

A2C is an advanced DRL algorithm that operates within the maximum entropy reinforcement learning framework. It follows an off-policy actor-critic approach, where the actor strives to maximize both the expected reward and the entropy of its policy. This unique objective encourages the agent to excel at the task while maintaining a high level of randomness in its actions. The combination of off-policy updates and a stable stochastic actor-critic formulation in A2C leads to improved performance and outperforms both on-policy and off-policy methods in a variety of continuous control benchmark tasks.

By utilizing off-policy updates, A2C can learn from data collected by different policies, enhancing its sample efficiency. It leverages a stable stochastic actor-critic formulation, which provides a more robust and reliable learning framework. These features contribute to A2C's state-of-the-art performance across various continuous control tasks. We define a parameterized $Q_{\phi}(\mathbf{s}^t, \mathbf{a}^t)$ as soft Q-function with the Q-network parameter is $\phi$ and a manageable policy $\pi_{\theta}(\mathbf{s}^t| \mathbf{a}^t)$ with network parameter $\theta$. The expressive neural networks can be utilized to model the soft Q-function, while the policy can be represented as a Gaussian distribution with its mean and covariance parameterized by neural networks. In addition, the soft state value function of A2C is defined as
\begin{align}
    V(\mathbf{s}^t) = \mathbf{E}_{\mathbf{a}^t \thicksim \pi} \left[Q_{\phi}(\mathbf{s}^t, \mathbf{a}^t) - \alpha\pi_{\theta}(\mathbf{s}^t| \mathbf{a}^t) \right],
\end{align}
where $\alpha$ is the temperature coefficient. 

The parameters of the soft Q-function can be optimized to minimize the soft Bellman remainder:
\begin{align}
    J_Q(\phi)&=\mathbb{E}_{\left(\mathbf{s}^t, \mathbf{a}^t\right) \sim \mho}\left[\frac{1}{2}\left(Q_\phi\left(\mathbf{s}^t, \mathbf{a}^t\right)-\left(r\left(\mathbf{s}^t, \mathbf{a}^t\right)\right.\right.\right.\notag\\
    &\left.\left.\left.+\gamma \mathbb{E}_{\mathbf{s}_{t+1} \sim p}\left[V_{\bar{\phi}}\left(\mathbf{s}_{t+1}\right)\right]\right)\right)^2\right],
\end{align}
where $\bar{\phi}$ is the target Q-network parameters. Then the soft Bellman can be optimized with stochastic gradients as follows
\begin{align}
    \hat{\nabla}_\phi J_Q(\phi)=&\nabla_\phi Q_\phi\left(\mathbf{a}^t, \mathbf{s}^t\right)\left(Q_\phi\left(\mathbf{s}^t, \mathbf{a}^t\right)-r\left(\mathbf{s}^t, \mathbf{a}^t\right)\right.\notag\\
    &\left.-\gamma V\left(\mathbf{s}^{t+1}\right)\right).
\end{align}
The target updates with parameters $\bar{\theta}$ are calculated as an exponentially moving average of the soft Q-function weights and have been demonstrated to stabilize training \cite{SAC-2018}. Furthermore, the policy is adjusted towards the exponential of the new soft Q-function during the policy improvement stage. This particular modification can be assured to result in a better policy in terms of soft value. Because tractable policies are preferred in reality, we shall limit the policy to some set of policies $\prod$, which can correspond, for example, to a parameterized family of distributions such as Gaussian distributions. The policy is improved into the intended collection of policies to account for the limitation imposed by policy $\pi \in \prod$. While we could use any projection in principle, the information projection specified in terms of the KL divergence becomes more convenient. Therefore, during the policy enhancement stage, we can update the policy for each state as follows:
\begin{align}
    \pi_{\textrm{new}}=\underset{\pi^{t+1} \in \Pi}{\arg\min}~\mathrm{D}_{\mathrm{KL}}\left(\pi^{t+1}\left(\cdot \mid \mathbf{s}^t\right) \Big\Vert \frac{\exp \left(Q^{\pi{\mathrm{old}}}\left(\mathbf{s}^t, \cdot\right)\right)}{Z^{\pi_{\mathrm{old}}}\left(\mathbf{s}^t\right)}\right).
\end{align}
\begin{figure*}[t]
\centering
\subfloat[Different sampling delays \label{fig:eval-sampling-delay}]{\includegraphics[width=0.246\linewidth]{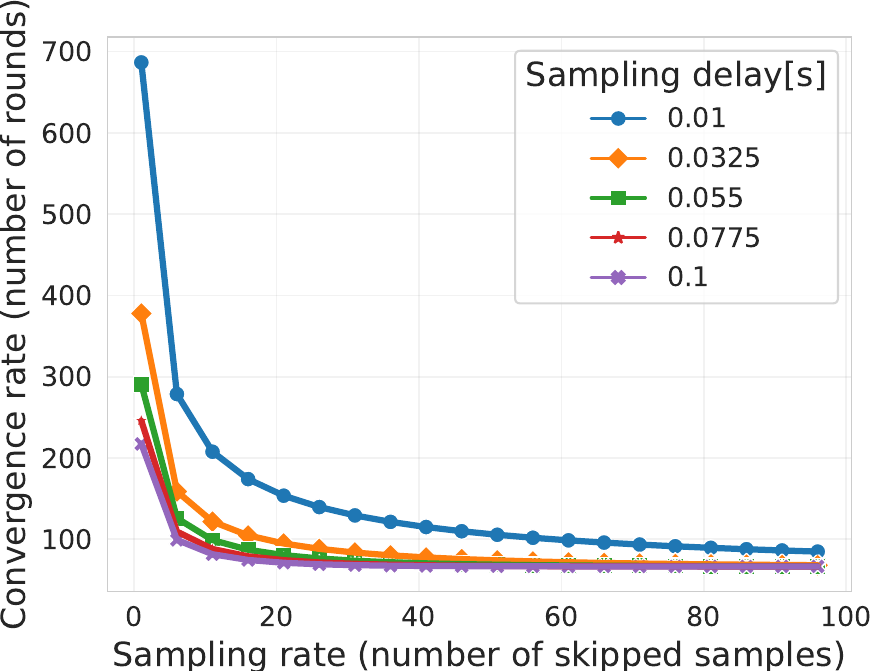}}
    \hfill
	\subfloat[Different $L$-smooth\label{fig:eval-L-smooth}]{\includegraphics[width=0.246\linewidth]{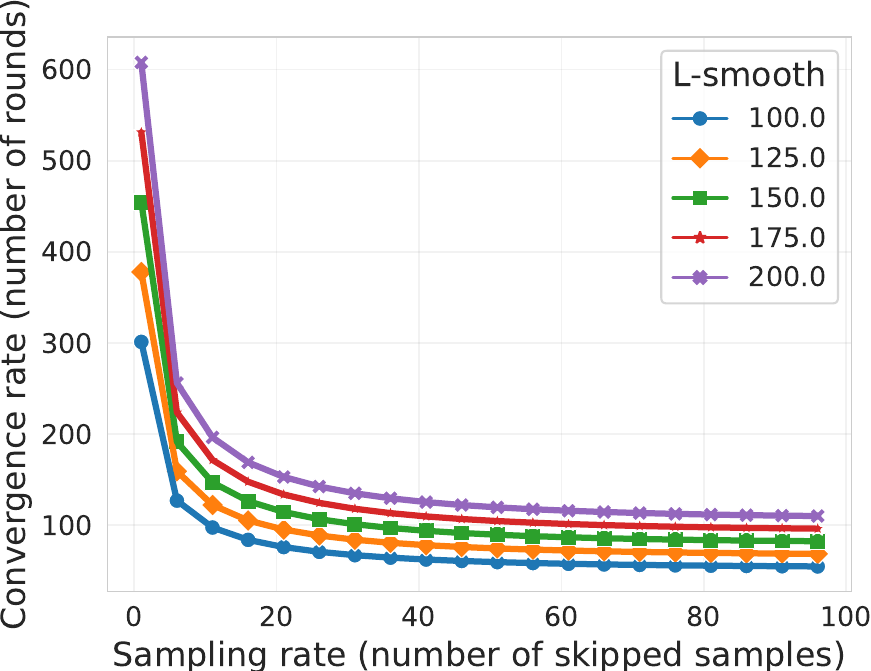}}
    \hfill
	\subfloat[Different global accuracies \label{fig:eval-global-accuracy}]{\includegraphics[width=0.246\linewidth]{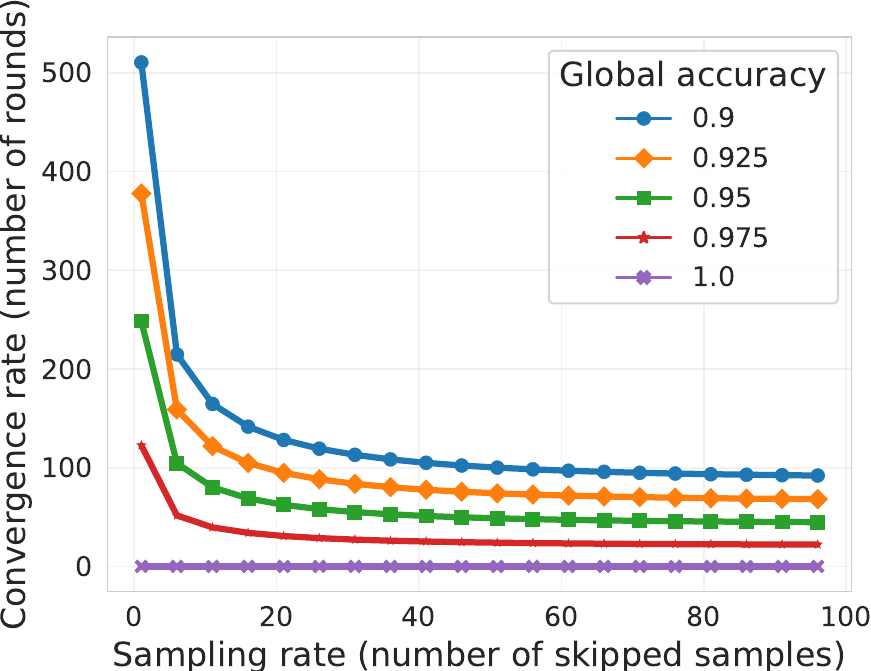}}
    \hfill
	\subfloat[Different number of users\label{fig:eval-number-of-user}]{\includegraphics[width=0.246\linewidth]{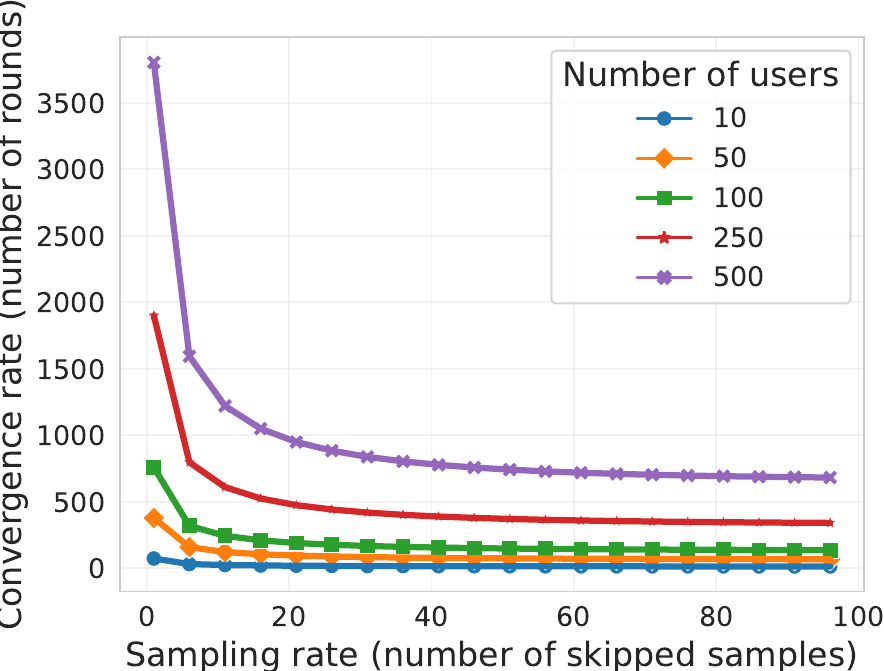}}
    \caption{Illustration of the lower boundary on the number of global iterations.}
    \label{fig:lower-boundary-iterations}
\end{figure*}

As a result, the policy parameters can be learned by directly minimizing the anticipated KL divergence, as shown below
\begin{align}\label{eq:KL_policy}
    J_\pi(\theta)=\mathbb{E}_{\mathbf{s}^t \sim \mho}\left[\mathrm{D}_{\mathrm{KL}}\left(\pi_\theta\left(\cdot \mid \mathbf{s}^t\right) \Big\Vert \frac{\exp \left(Q_\phi\left(\mathbf{s}^t, \cdot\right)\right)}{Z_\phi\left(\mathbf{s}^t\right)}\right)\right],
\end{align}
where $\mho$ is the experience replay buffer. To minimize $J_{\pi}$ and lower the variance estimator of the target density Q-function, the policy  is parameterized by using a neural network transformation 
\begin{align}
    \mathrm{a}^t = f_{\theta}\left(\aleph^t;\mathrm{s}^t\right),
\end{align}
where $\aleph^t$ is an input noise vector, which is sampled from a fixed spherical Gaussian distribution. Therefore, the policy in \eqref{eq:KL_policy} could be rewrite as
\begin{align}
    J_\pi(\theta)=&\mathbb{E}_{\mathbf{s}^t \sim \mho, \aleph^t \mathcal{N}}\left[\log \pi_\theta\left(f_\phi\left(\aleph^t ; \mathbf{s}^t\right) \mid \mathbf{s}^t\right)\right.\notag\\
    &\left.-Q_\phi\left(\mathbf{s}^t, f_\theta\left(\aleph^t ; \mathbf{s}^t\right)\right)\right],
\end{align}
with $\pi_{\theta}$ is implicitly defined in terms of $f_{\theta}$. By integrating EI design into A2C, our proposed approach A2C-EI can leverage the stability in learning while being robust against the high-complexity problem. To obtain effective learning performance for the SCFL system, A2C-EI is a flexible mix of the regular A2C algorithm with explicit restrictions from SCFL. The pseudo algorithm for A2C-EI is presented in Algorithm~\ref{alg:A2C-EI}.

{\color{black}
\subsection{Discussion on Complexity Analysis and Implementation}

The proposed A2C-EI framework in our work utilizes a lightweight neural network architecture, consisting of only a single hidden layer per stream. Coupled with advancements in network compression \cite{Chen_2024_CVPR}, approximate computing \cite{8792369}, and hardware accelerators, this simplified architecture facilitates the practical deployment of our solution on standard IoT hardware.

Nonetheless, for ultra low-power transmitters where implementing DRL locally remains infeasible, we propose an alternative architecture involving a central controller situated at the gateway. This controller, equipped with sufficient computational capacity, executes the A2C-EI algorithm. It is important to clarify that the high number of algorithmic iterations refers to the training process of the A2C-EI algorithm, not to real-time interactions between the transmitter and gateway. Consequently, the associated computational and communication overheads are manageable in practice.

In this framework, at each state, the transmitter selects an action based on the current policy and observes the corresponding environmental feedback. The resulting experience tuple, denoted as $(s^t;a^t;r^t;s^{t+1})$, is stored in a local memory pool. Periodically, for example, once per day, the transmitter uploads the accumulated experiences to the gateway. The gateway then leverages this data to train the deep neural network and estimate the Q-values for various state-action pairs, thereby refining the optimal policy. Once updated, this policy is transmitted back to the device, where it is stored for use in real-time decision-making. Due to the asynchronous and batch-based nature, the proposed system remains compatible with low-latency application requirements.
}

\section{Simulation Evaluation}\label{sec:Results}
In this section, we study the performance of the proposed SCFL for IoT networks with RTS by using computer simulation results. All statistical results are averaged over $10$ independent runs.
\subsection{IoT Network Settings}
The ES is located at the center of a circular area with a radius of $1$ km, serving 100 users uniformly distributed. The path loss model is given by $128.1 + 37.6 \log_{10} d$, where $d$ is the distance between the ES and the user (in km) \cite{ni2024path}. The standard deviation of shadow fading is 8 dB \cite{van2021latency}, and the Gaussian noise power spectral density is $n_0 = -174$ dBm/Hz \cite{bazzi2023secure}. The transmit power is in the range of $(0, 20]$ dBm, and the system bandwidth is $B = 20$ MHz. We define the max number of sampling data skipped as $100$. The number of CPU cycles for computing one sample data $C_u$ is uniformly distributed in $[1:3] \times 10^4$ cycles/sample, and computation capacity on each user set equally $f_u = 2$ GHz. The effective switched capacitance in local computation is $\kappa = 10^{-28}$ \cite{2016-MEC-DynamicOffloading}. To evaluate the FCL transmission, we use the ResNet-9 \cite{He_2016_CVPR} with a data size of $28$ MB.
 
\subsection{AI Model Settings}
To properly simulate the AI task behavior, we consider the key features that capture all characteristics of training data and training AI model, i.e., $L$-smooth and $\mu$-strongly convex. To this end, we deploy a classification task on the CIFAR-10 dataset \cite{2010-DL-CIFAR}. We sample data and feed through the AI model to consider the Hessian of the loss function. We follow the theory \cite{2020-DL-LAMB}, in which the minimizer tends to be the sharpest at the initial phase of the AI training. Therefore, by considering the Hessian of the loss function of the untrained AI model with the specific dataset, the minimizer can achieve the highest second derivative value, which is approximately close to the $L$-smooth and $\mu$ strongly convex value. Our theoretical implementation for $L$-smooth and $\mu$ strongly convex estimation can be found via experimental evaluation code\footnote[1]{\url{https://github.com/Skyd-FL/SCFL/blob/main/results/theoretical_evals/Lsmooth_Estimation.ipynb}.}. According to our evaluation, we found that the values of $L$ and $\mu$ on MNIST and CIFAR-10 dataset are higher than $90$ and $150$, respectively. Therefore, we choose $L$ and $\mu$ for our experimental evaluation with a set of values with an absolute value equal to $\{100, 125, 150, 175, 200\}$ ($L$ is the set with positive values while $\mu$ is the sets with negative one).
\begin{figure*}[!ht]
    \raggedright 
    \subfloat[\label{fig:subplot5}]{\includegraphics[width=0.24\linewidth]{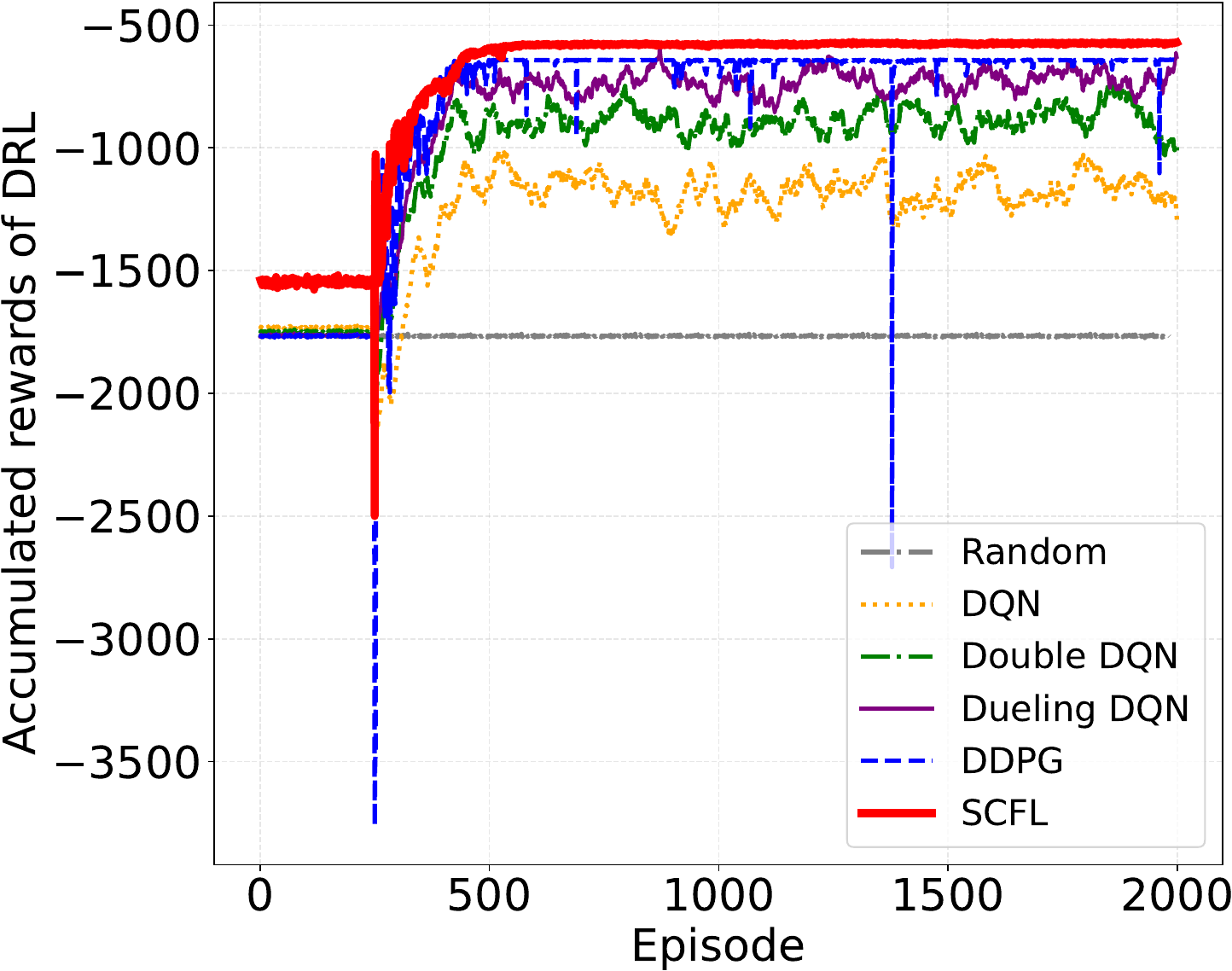}}
    \hspace{0.005\linewidth}
    \subfloat[\label{fig:subplot1}]{\includegraphics[width=0.24\linewidth]{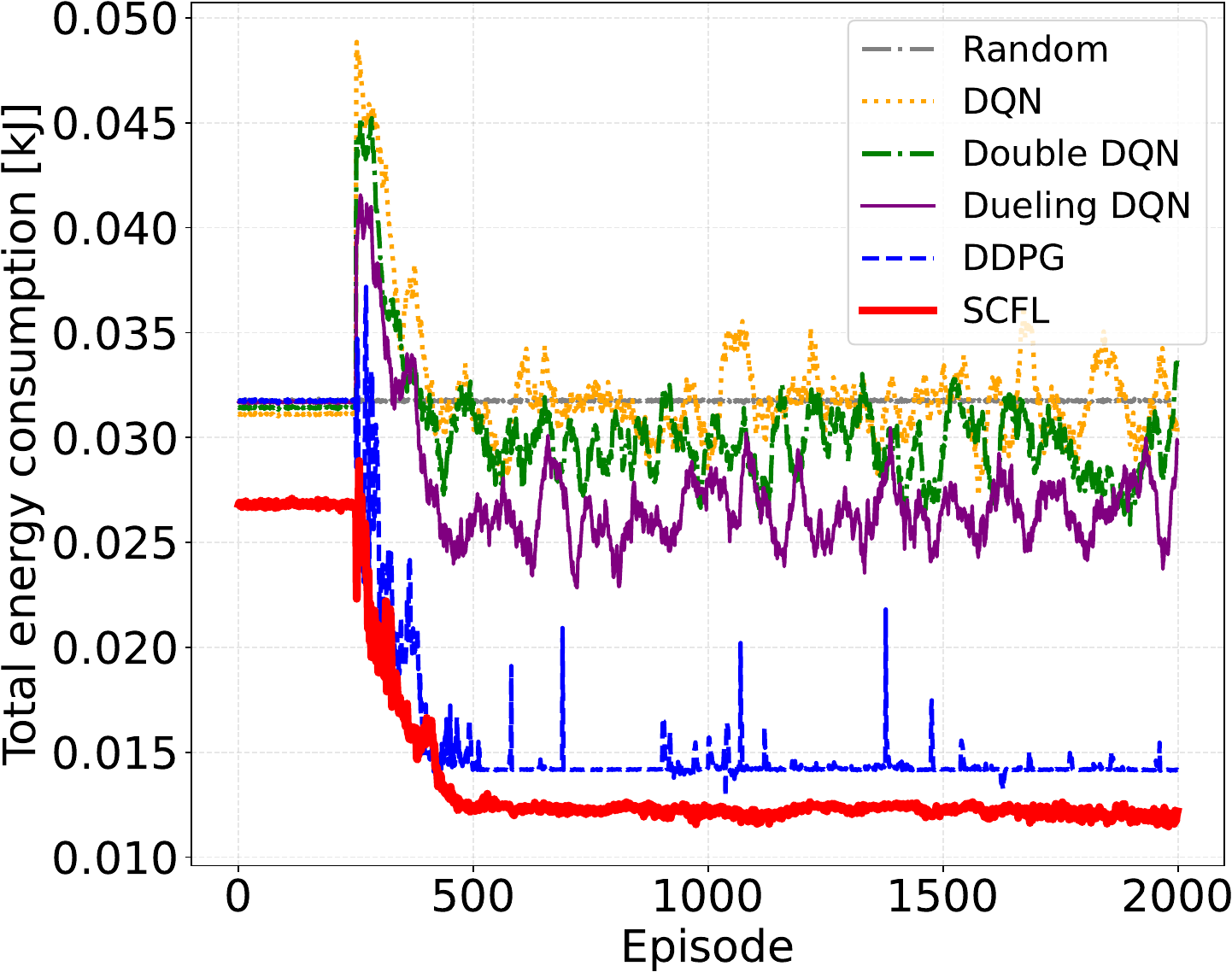}}
    \hspace{0.005\linewidth} 
    \subfloat[\label{fig:subplot6}]{\includegraphics[width=0.24\linewidth]{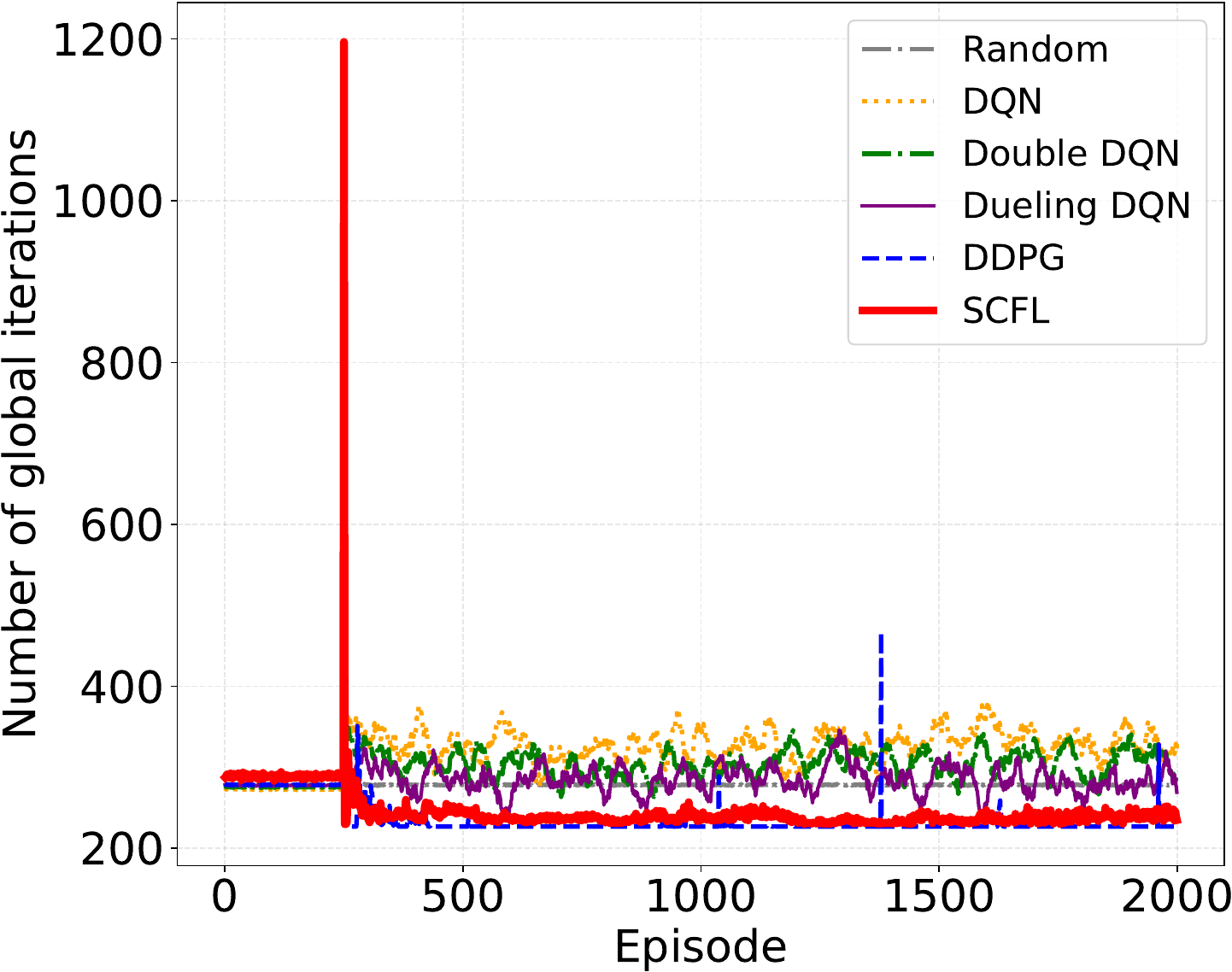}}
    \hspace{0.005\linewidth} 
    \subfloat[\label{fig:subplot8}]{\includegraphics[width=0.24\linewidth]{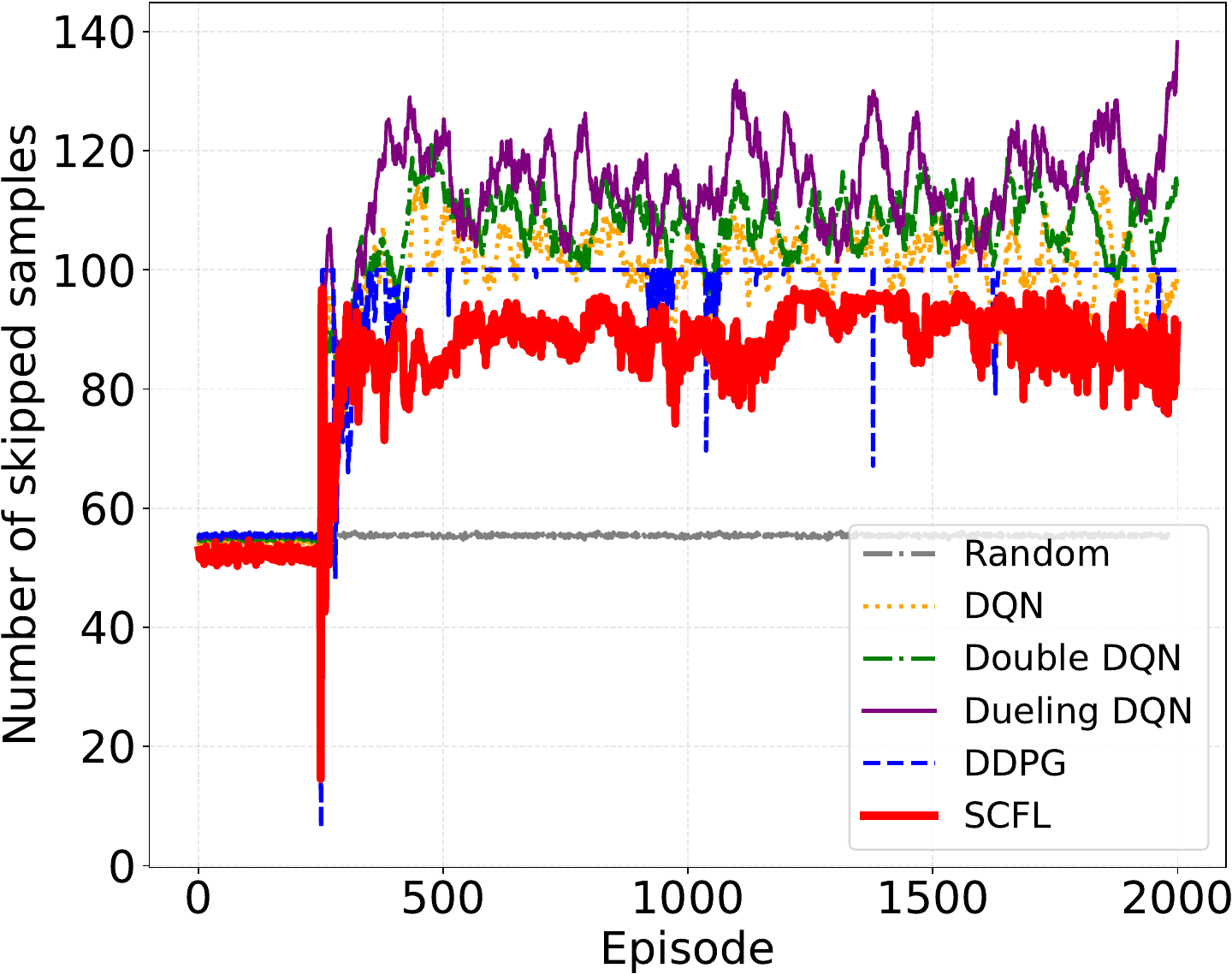}}
    \caption{The figures illustrate the performance of SCFL and other benchmarks over 2000 episodes. Figures~\ref{fig:subplot5}, \ref{fig:subplot1}, \ref{fig:subplot6}, \ref{fig:subplot8} demonstrate the convergence of accumulated DRL reward, total energy consumption $E$, $k_u$ skipped samples over training episodes, respectively.}
    \label{fig:P_SCFL}
\end{figure*}

\subsection{Convergence Analysis of SCFL}
Fig.~\ref{fig:lower-boundary-iterations} describes the influence of the sampling rate on the bound of convergence rate to verify the correctness of our bounded function in \eqref{eq:I_glob}. We evaluate the crucial parameters, including sampling delay, $L$-smooth, global accuracy, and the number of users.

Fig.~\ref{fig:eval-sampling-delay} shows the convergence rate by varying the sampling period from $0.01$s to $0.1$s. At the delay level with $\tau_u=0.01$s, the number of global iterations rapidly decreases from over $1000$ to just over $100$ rounds at the converged point, reducing nearly tenfold as the number $k$ increases from $0$ to $100$. When $\tau_u$ reaches $0.0325$s or higher, the number of rounds needed for convergence decreases to 600 and stabilizes around $400$ at $k_u=0$. Afterward, the number of iterations quickly reaches convergence at approximately $100$ rounds when $k_u=40$. Therefore, striking a balance between the choice of $\tau_u$ and the number of skip samples $k_u$ aids the SCFL algorithm in achieving stable and rapid convergence.

Fig.~\ref{fig:eval-L-smooth} compares different levels of data complexity (i.e., $L$-smoothness coefficient); containing sharper minimizers and higher $L$-smooth values. We can observe that as the $L$-smoothness level increases, the number of iterations required for the SCFL algorithm to converge also increases. This indicates that optimizing an FCL algorithm is dependent on the variations in dataset complexity. As such, for datasets with low complexity like MNIST or CIFAR-10, we only need a small number of iterations for the algorithm to reach convergence.

Fig.~\ref{fig:eval-global-accuracy} illustrates the influence of the bound of convergence rate by the variations in global accuracy. We recall that the global accuracy $\varpi$ represents the difference between the training model parameters and the optimal model parameters, and a lower global accuracy indicates a better match. When the global accuracy is lower, the number of rounds required to achieve convergence increases as mathematically shown in \eqref{eq:I_glob}. Therefore, it is essential to strike a balance between the number of rounds for convergence and $\varpi$ to ensure optimization. As observed, for $\varpi$ values within the range of $[0.06, 0.2]$, only around 200 iterations are needed for the SCFL algorithm to converge with approximately $k_u=20$.

Fig.~\ref{fig:eval-number-of-user} represents the influence of the number of users on the number of iterations required for the SCFL algorithm to converge. As $U$ increases, the complexity of the system model increases, leading to a higher number of iterations needed for convergence. When the number of users is sufficiently large (i.e., $U=100$), the number of iterations for SCFL to converge exceeds 6000 rounds. For $U$ values smaller than 100, the algorithm can quickly reach convergence with $k_u=20$, and the required number of rounds also decreases by 10 times compared to scenarios with a large number of users in the network. Notably, regardless of the network size, we can control the sampling rate to speed up the SCFL learning process.

\subsection{Learning Performance of SCFL}
\begin{figure*}[!ht]
\centering
    \subfloat[Max transmission power $p^{\mathrm{max}}_u$\label{fig:EvsmaxP}]{\includegraphics[width=0.24\linewidth]{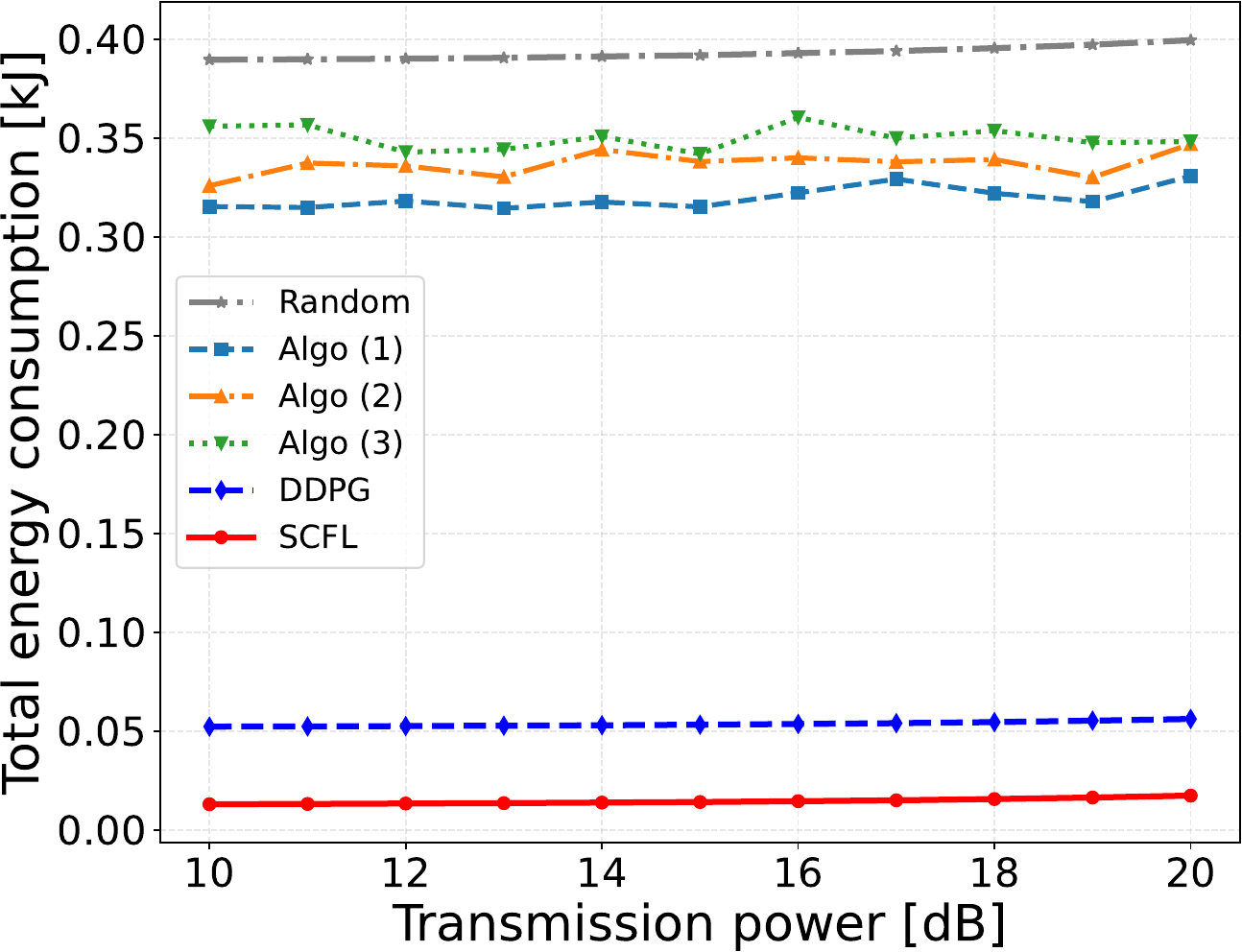}}
    \hspace{0.005\linewidth}
    \subfloat[Max CPU frequency $f^{\mathrm{max}}_u$\label{fig:EvsmaxF}]{\includegraphics[width=0.24\linewidth]{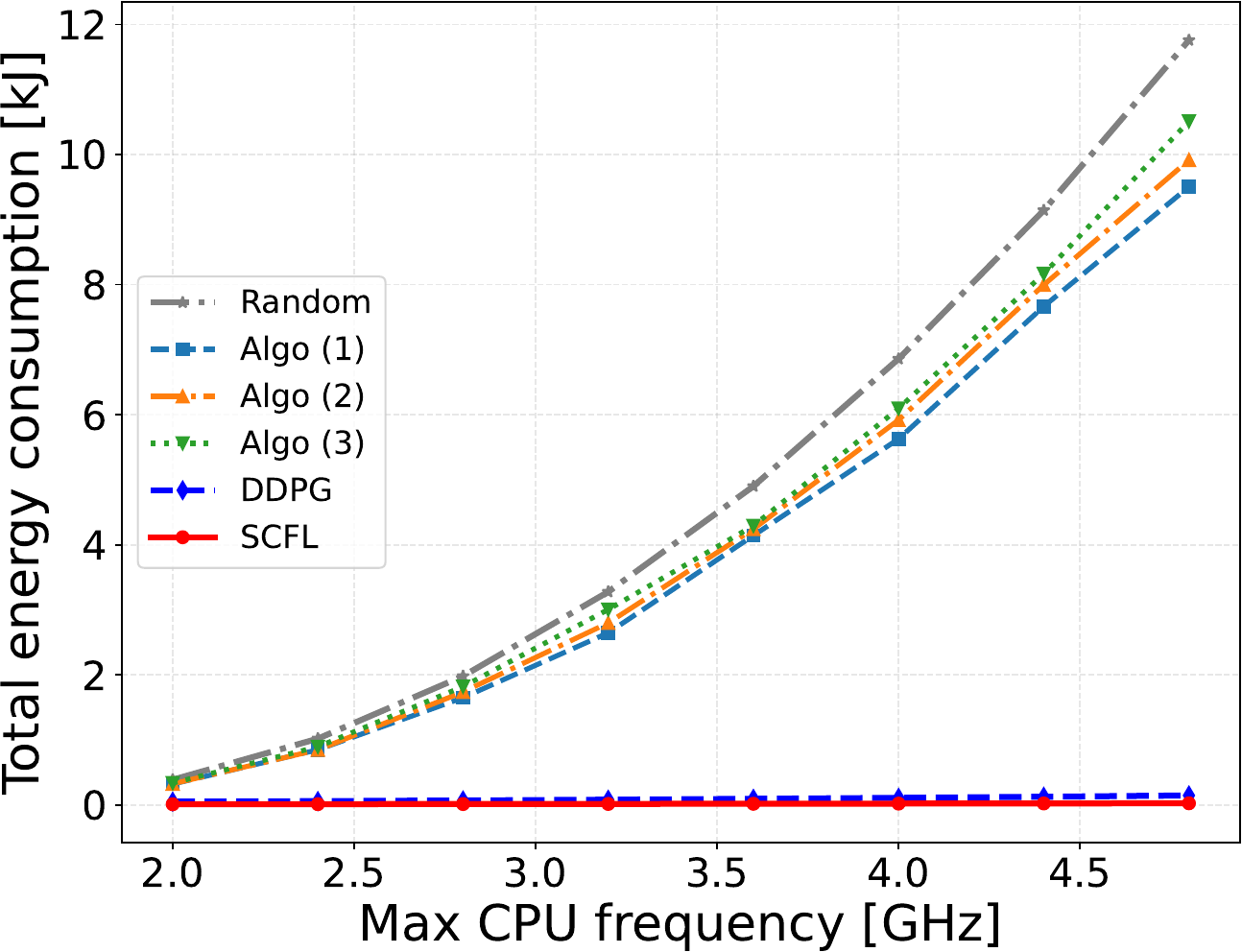}}
    \hspace{0.005\linewidth}
    \subfloat[Different bandwidth $B$\label{fig:EvsBandwidth}]{\includegraphics[width=0.24\linewidth]{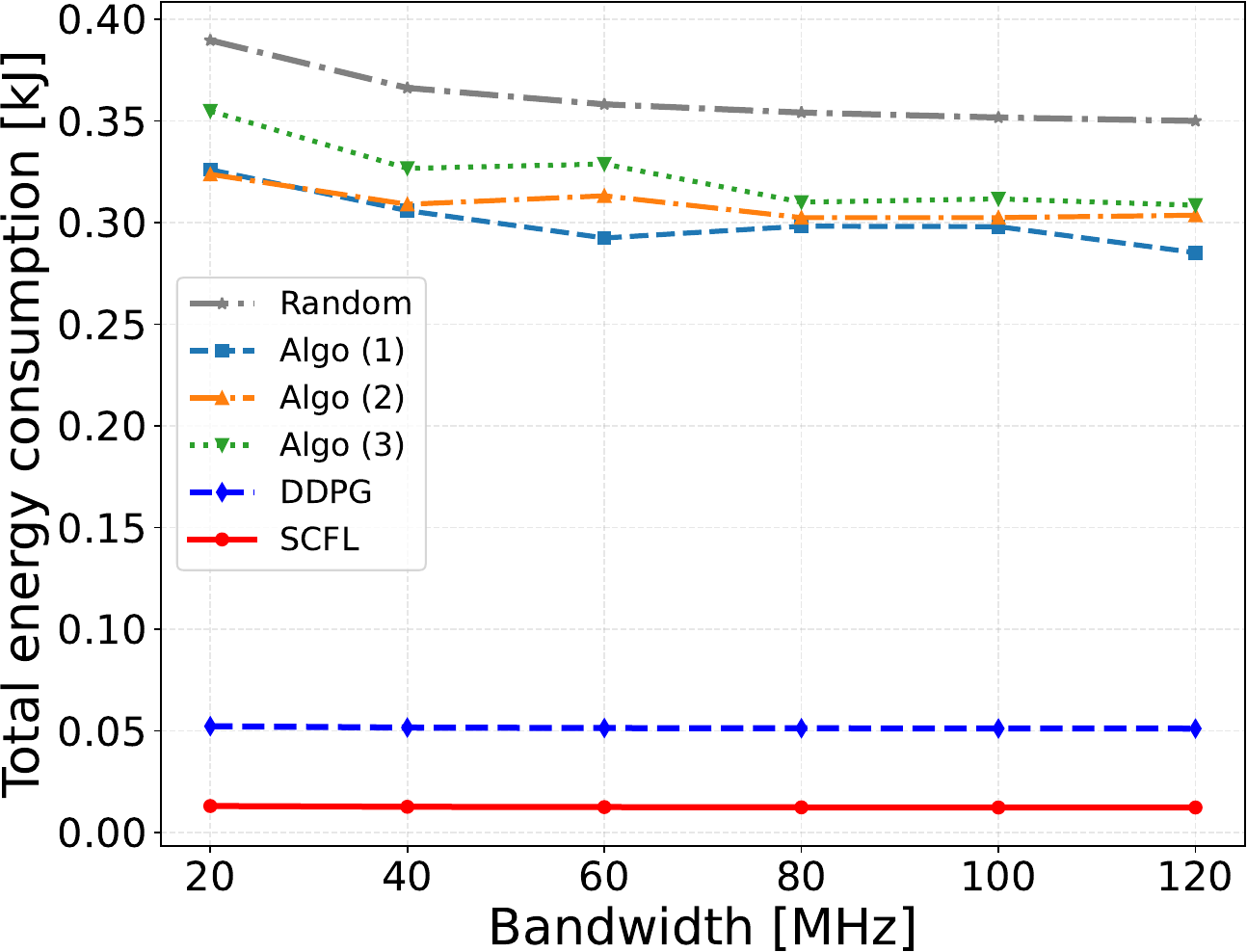}}
    \hspace{0.005\linewidth}
    \subfloat[Max skipped samples $k_{\mathrm{u\_thres}}$\label{fig:EvsMaxskip}]{\includegraphics[width=0.24\linewidth]{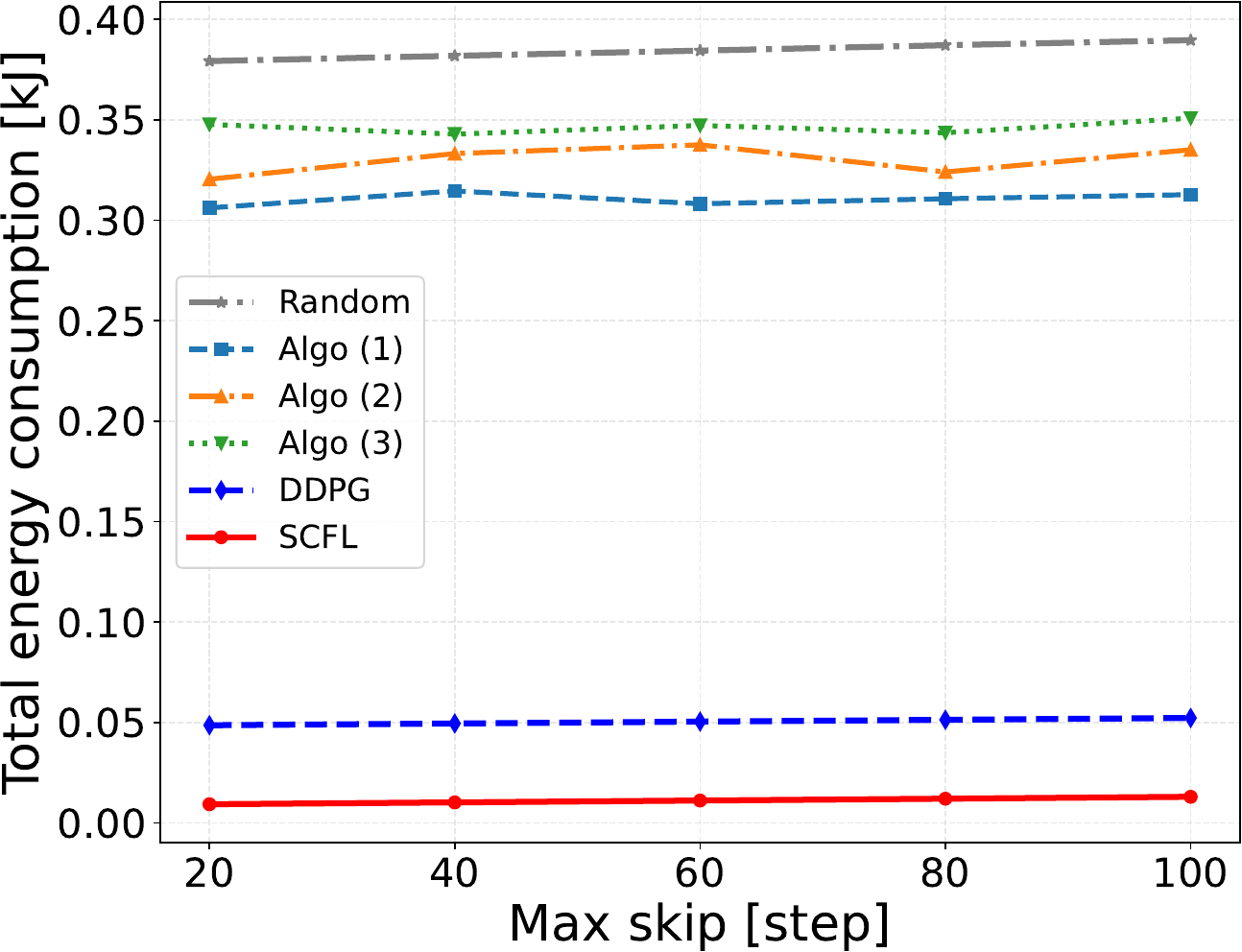}}
    \hspace{0.005\linewidth}\\
    \subfloat[Max transmission power $p^{\mathrm{max}}_u$\label{fig:TvsmaxP}]{\includegraphics[width=0.24\linewidth]{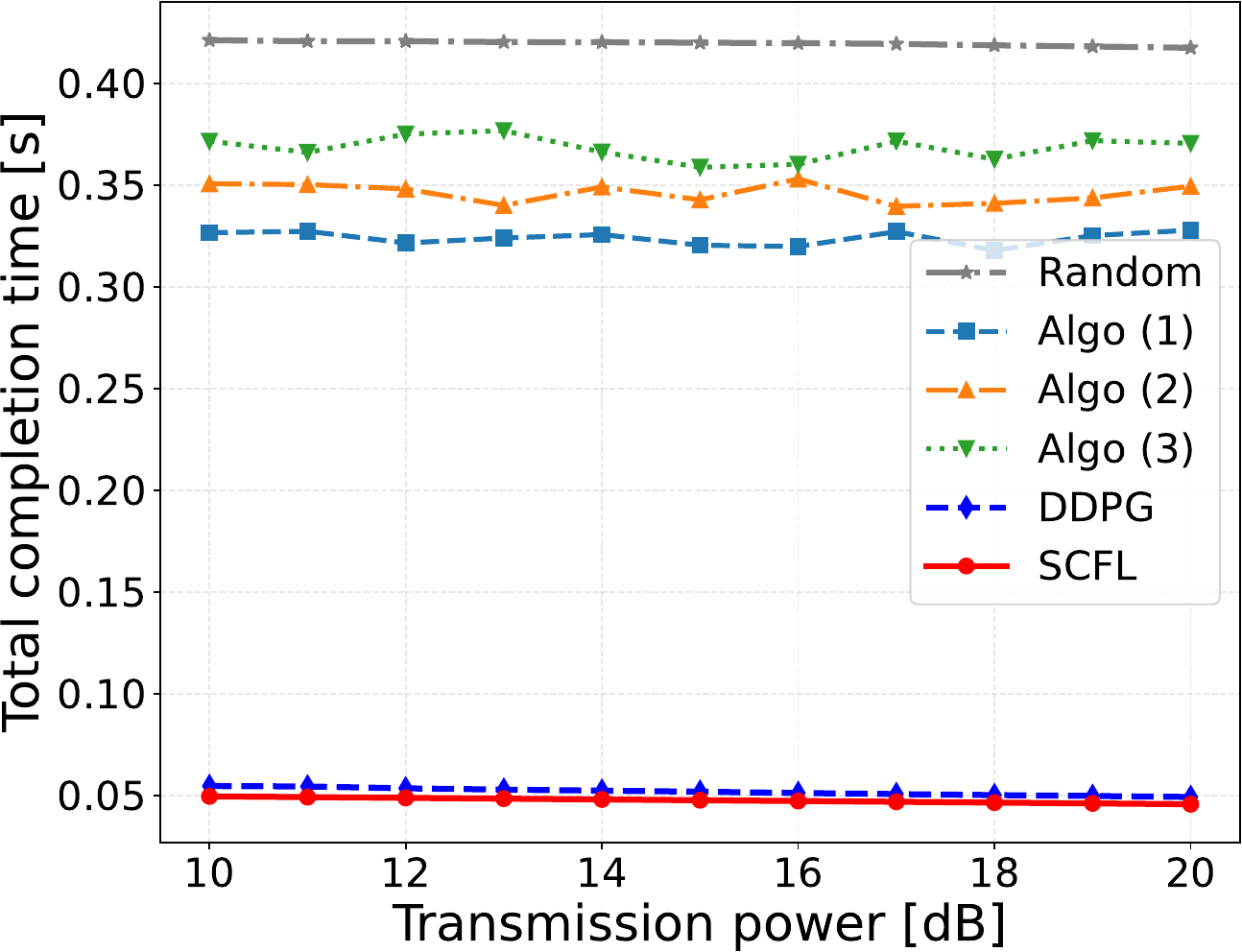}}
    \hspace{0.005\linewidth}
    \subfloat[Max CPU frequency $f^{\mathrm{max}}_u$\label{fig:TvsmaxF}]{\includegraphics[width=0.24\linewidth]{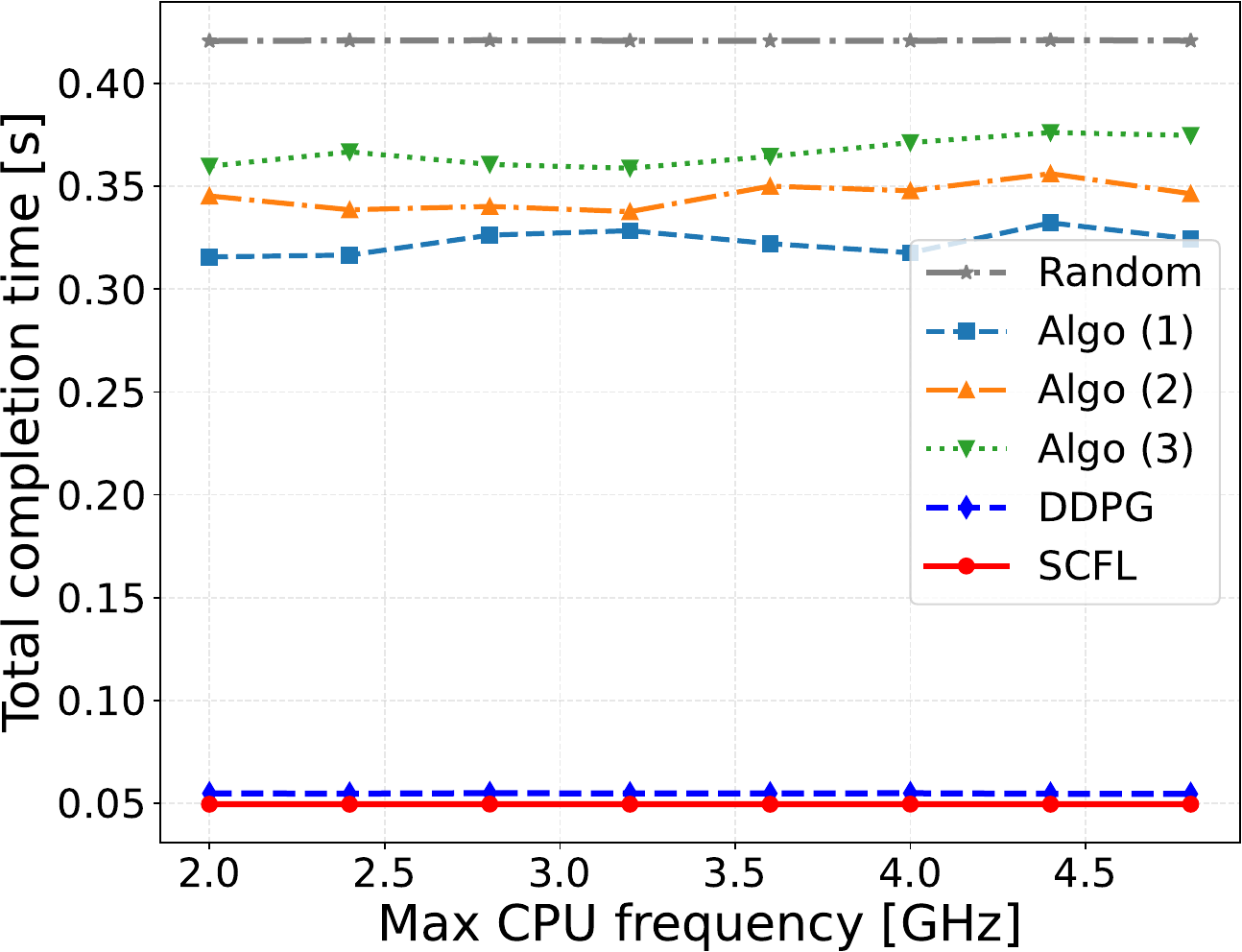}}
    \hspace{0.005\linewidth}
    \subfloat[Different bandwidth $B$\label{fig:TvsBandwidth}]{\includegraphics[width=0.24\linewidth]{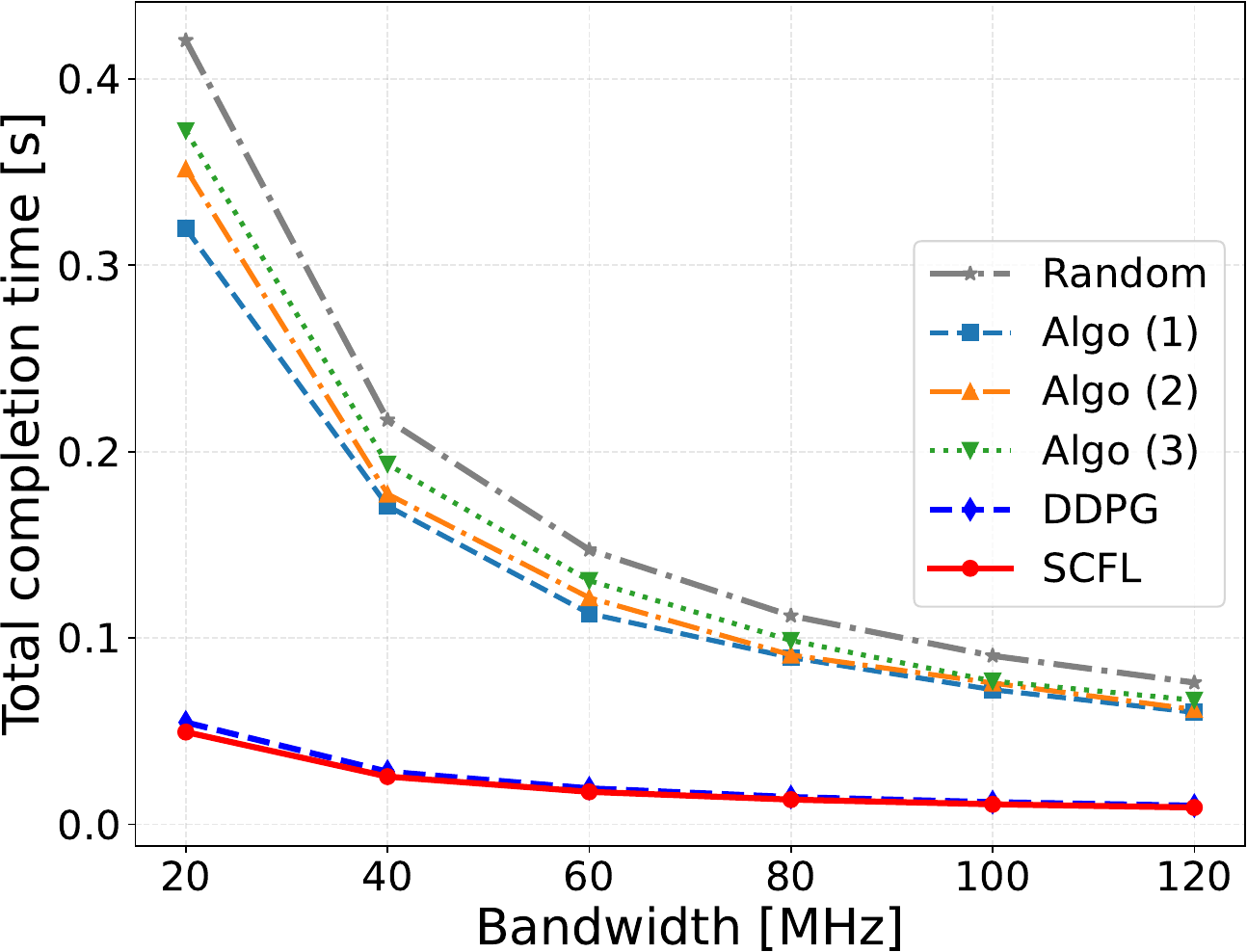}}
    \hspace{0.005\linewidth}
    \subfloat[Max skipped samples $k_{\mathrm{u\_thres}}$\label{fig:TvsMaxskip}]{\includegraphics[width=0.24\linewidth]{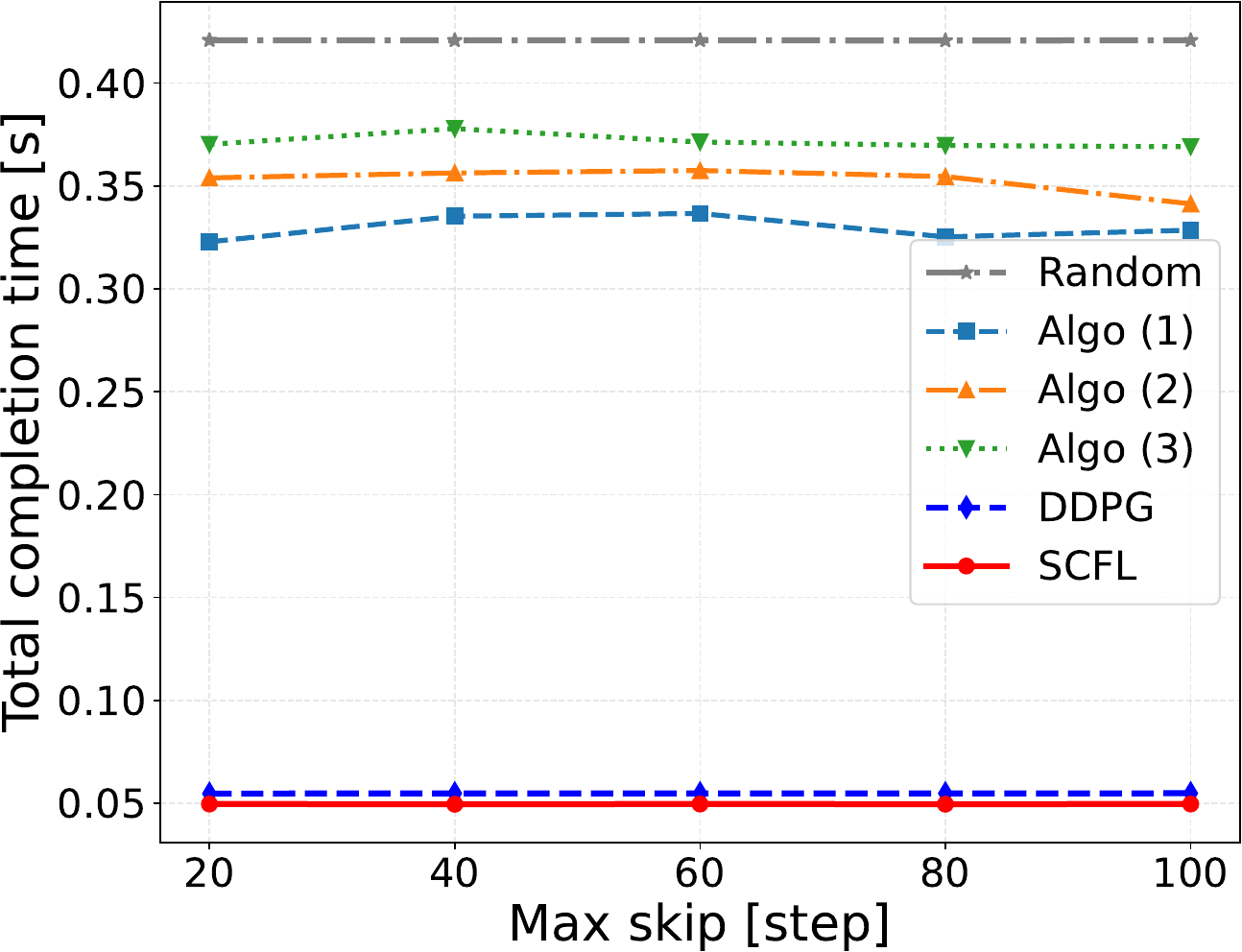}}
    \hspace{0.005\linewidth}
    \caption{{\color{black}A comparison of energy efficiency and overall completion time between SCFL and other DRL benchmarks. Specifically, Algo (1), Algo (2), and Algo (3) correspond to the approaches proposed in \cite{2020-FL-JointLearning-CommunicationsFramework}, \cite{2019-FL-OptimizationModel}, and \cite{2020-FL-EnergyEfficientFL}, respectively.}}
    \label{fig:SCFL_Total_performance}
\end{figure*}
We visualize the performance analysis of SCFL across learning episodes, illustrating that SCFL is capable of stable learning and rapid in Fig.~\ref{fig:P_SCFL}. Furthermore, the figure represents that we can achieve a fast adaptation on almost all features (from episode $250$ to $500$). The initial randomness before round $250$ can be attributed to the charging stage for the experience replay memory.

{\color{black}Fig.~\ref{fig:P_SCFL} describes SCFL bounds and affirms the stability and optimization of this algorithm in evaluation bounds across each round during the learning process. Regarding overall energy consumption, as shown in Fig.~\ref{fig:subplot1}, SCFL's performance is stable and outstanding compared to other benchmarks, including DDPG, Dueling DQN, Double DQN, DQN. Fig.~\ref{fig:subplot6} presents the optimized global iterations $I_\mathrm{glob}$ and Fig.~\ref{fig:subplot8} show the number of skipped samples after 2000 training episodes, respectively.}

On average, SCFL requires slightly fewer global iterations than other baselines to achieve convergence in learning. While SCFL exhibits a higher number of global learning rounds than the random policy, in terms of energy optimization weights, SCFL maintains a superior advantage over the random policy.

\subsection{Energy Efficiency Comparison}\label{sec:energy-efficiency}
{\color{black}For varying settings in local computation, we conducted an energy optimization analysis and total processing time comparison of SCFL against other benchmarks (i.e., \cite{2020-FL-JointLearning-CommunicationsFramework}, \cite{2019-FL-OptimizationModel}, \cite{2020-FL-EnergyEfficientFL}). As shown in Fig.~\ref{fig:SCFL_Total_performance}, our algorithm exhibits the lowest energy consumption and completion time compared to other baselines.}
As the amount of local computing increases, the total energy consumption and completion time tend to increase and stabilize slightly across all methods.
Our method performs comparatively better than the two benchmark schemes on average 86\% lower for total energy consumption ranging on various parameter settings. Regarding changes when increasing $f_u^{\max}$, the random agent shows sensitivity and a sharp increase in total energy consumption. SCFL shows an average approximate 88\% reduction compared to the DDPG method when the max CPU frequency increases from 2 GHz to 4.5 GHz, as illustrated in Fig.~\ref{fig:EvsmaxF}. The significant reduction of total energy consumption compared to the random policy in Fig.~\ref{fig:EvsmaxF} is 374.54\%, with an average reduction of all total energy consumption over than 350\%. 

In the case of total completion time illustrated in Fig.~\ref{fig:TvsmaxP}, Fig.~\ref{fig:TvsmaxF}, Fig.~\ref{fig:TvsBandwidth} and  Fig.~\ref{fig:TvsMaxskip}, the SCFL method shows an average approximate 10\% reduction compared to the DDPG method and over 88\% reduction compared to the random policy benchmark. In Fig.~\ref{fig:TvsBandwidth}, when the bandwidth increases, the random policy method downfalls rapidly from over 0.4s to approximately 0.1s, while the SCFL method and the DDPG method decrease slightly about 20ms when bandwidth increases from 40MHz to 120MHz. As bandwidth increases, the total processing and transmission time decreases; in other words, the system latency is inversely proportional to bandwidth. In the case of the FCL systems using a random policy, latency is more sensitive to bandwidth changes and decreases significantly. In contrast, with the SCFL and DDPG methods, the total processing time decreases slightly and remains more stable. Notably, the proposed SCFL method maintains lower latency than the other two methods. 

The overall results presented in Fig.~\ref{fig:SCFL_Total_performance} highlight the superiority of our proposed method across various aspects of FCL scenarios, demonstrating significant improvements in computational efficiency and time consumption when compared to other state-of-the-art methods. Our algorithm demonstrates the lowest energy consumption and completion time compared to other DRL agents, achieving approximately an 86\% reduction in energy consumption relative to the benchmarks. Additionally, SCFL shows an average reduction of over 88\% in total completion time compared to the random policy benchmark. As illustrated in various figures, the SCFL method consistently maintains lower latency and greater stability across changing conditions, highlighting its effectiveness over traditional approaches.

\begin{figure}[!ht]
    \raggedright 
    \subfloat[FCL benchmarks \label{fig:fcl-1}]{\includegraphics[width=0.49\linewidth]{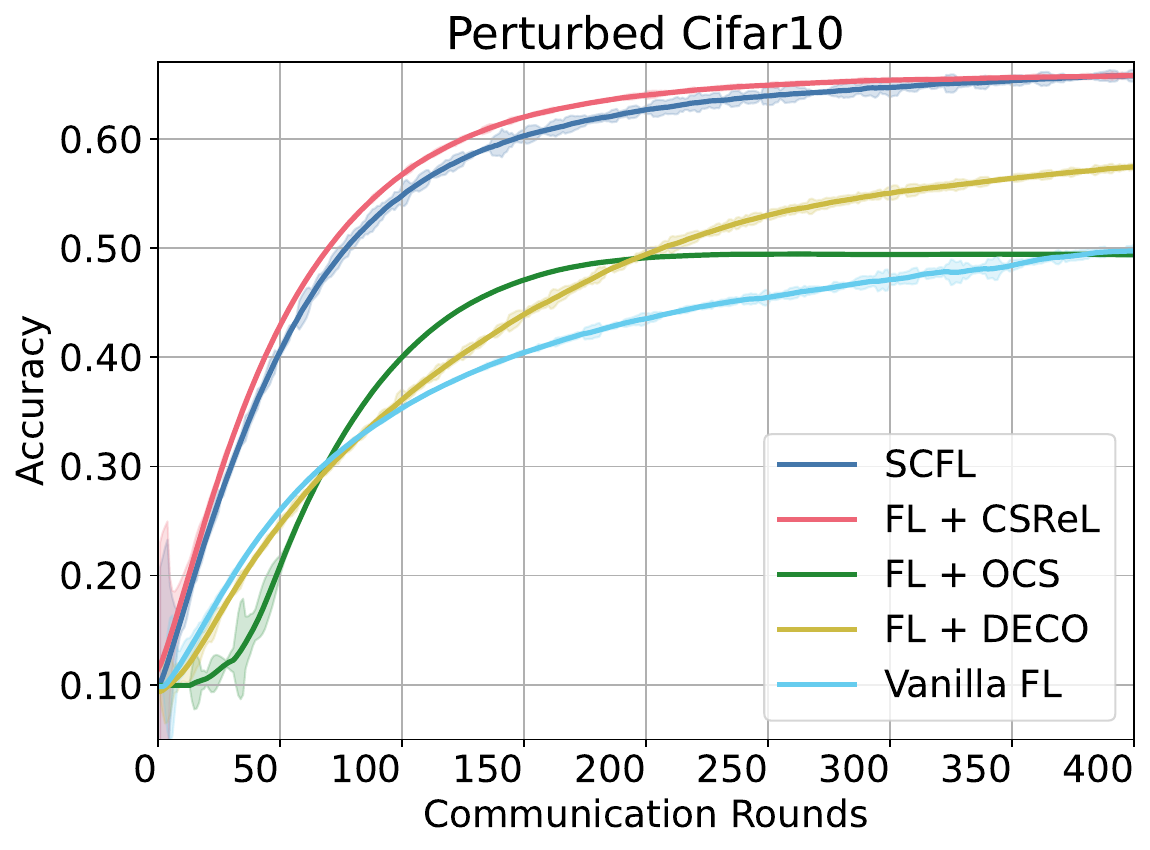}}
    \subfloat[SCFL ablation test \label{fig:fcl-2}]{\includegraphics[width=0.48\linewidth]{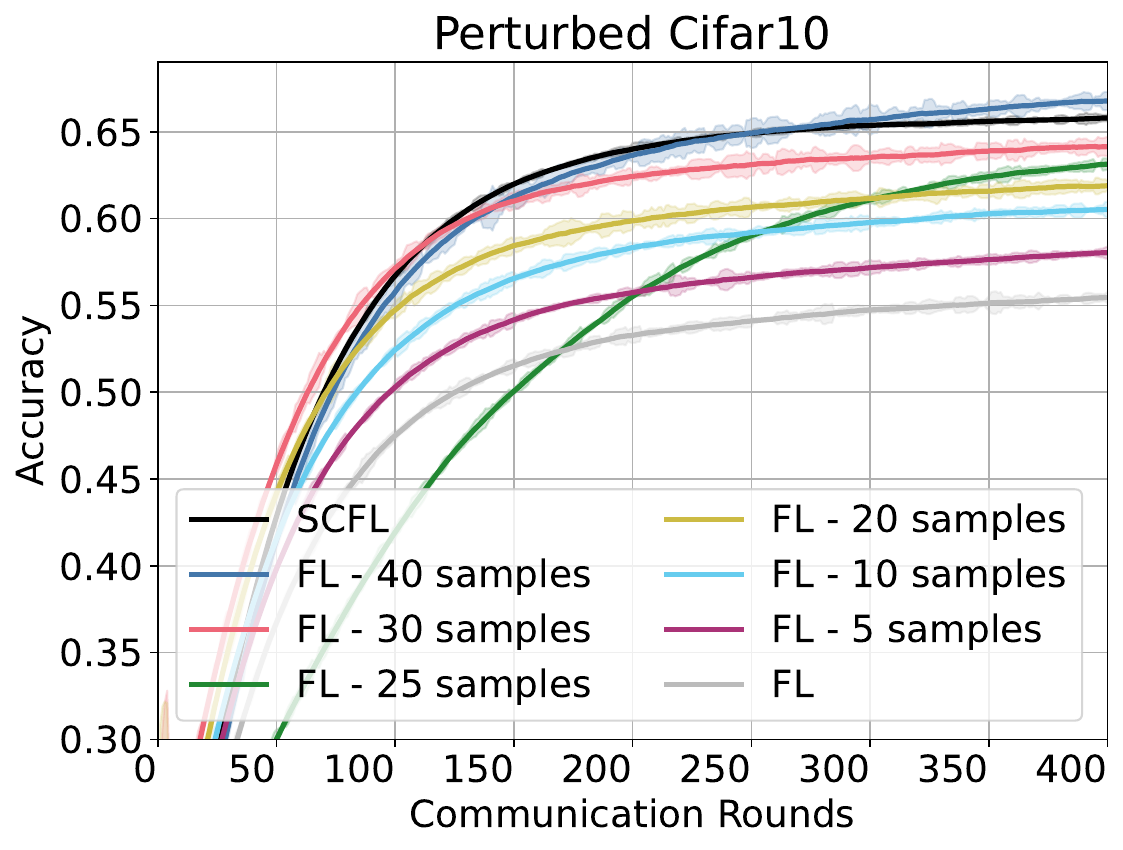}}
    \hspace{0.005\linewidth} 
    \caption{Benchmarks of FCL settings on Perturbed CIFAR10.}
    \label{fig:Acc-SCFL}
\end{figure}

{\color{black}\subsection{Comparisons of FCL Benchmarks}
Fig.~\ref{fig:fcl-1} presents a comparison between SCFL and several baseline methods, namely DECO \cite{10656539}, OCS \cite{yoon2022online}, CSReL \cite{tong2025coreset} and vanilla FL, in the context of selecting efficient samples for replay memory in continual learning. To simulate the continual learning setting, the input data is perturbed with small variance; once the perturbation exceeds a certain threshold, the data point transitions into a different sample. The results demonstrate that SCFL performs competitively with existing coreset selection methods. However, it is important to note that methods such as DECO, OCS, and CSReL operate by selecting coresets from the entire dataset, requiring all samples to be stored on the device. This approach results in energy consumption similar to that of standard FL and does not offer improvements in sampling efficiency. Unlike other baselines that focus solely on selecting samples from a static, preprocessed dataset, SCFL takes into account the temporal correlation of incoming data. In the Perturbed CIFAR-10 setting, where consecutive data points are highly correlated, existing methods tend to underperform. In contrast, SCFL improves performance by selectively sampling data such that the correlation between consecutive samples is minimized.}

{\color{black}\subsection{Ablation test on SCFL}
Fig.~\ref{fig:fcl-2} presents an ablation study of SCFL on the Perturbed CIFAR-10 dataset. In this experiment, we compare the optimized SCFL approach with variants of FL where different numbers of samples are intentionally skipped. The results indicate that when more than 25 samples are skipped, the convergence behavior remains largely unaffected. This suggests that SCFL is able to maintain performance while selectively skipping samples to improve energy efficiency. Consequently, SCFL achieves a favorable trade-off between accuracy and resource consumption, ensuring that unnecessary samples are not processed, thereby enabling energy savings through the RTS mechanism described in Section~\ref{sec:energy-efficiency}.
}

\section{Conclusion}
\label{sec:Conclusion}
In this paper, we have introduced a novel FCL framework called SCFL and provided the mathematical proof of the convergence analysis on the test set with arbitrary aspects. Our focus is on analyzing the data sampling characteristics in RTS distributed systems. By considering the impact of sampling delay between consecutive data samples on AI learning efficiency, our proposed SCFL ensures energy efficiency in the FCL system while maintaining high performance on the test dataset. Notably, our research contributes significantly by being the first to propose a convergence analysis on the test dataset. This pioneering work lays a strong foundation for future research in optimizing FCL networks. Specifically, it opens avenues to explore the relationship between data-related features, such as the number of training data and data relevance among distributed devices, and the FCL convergence via the generalization gap. {\color{black}One primary limitation of the proposed method lies in the accuracy of the generalization gap approximation, which is influenced by the sampling time. Consequently, deriving a tighter bound for this approximation could lead to more effective resource allocation algorithms. Future research can build upon the utilization of the generalization gap to better connect device-specific data characteristics with the convergence behavior of federated learning, thereby enhancing the practicality of resource allocation strategies.}

\bibliographystyle{IEEEtran}
\bibliography{0-2-SCFL-Full.bib}

\clearpage
\appendices
{\color{black}
\section{Preliminaries \& Backgrounds}
\subsection{Federated Continual Learning}
FCL \cite{2024-FCL-sur} is defined as the process of learning from data that is distributed dynamically over time. In practice, training samples originating from different distributions are presented sequentially. An FCL model, parameterized by $\theta$, must learn to perform well on the corresponding tasks with either limited or no access to previously encountered training data. Specifically, an incoming batch of training samples associated with task $t$ is denoted by $\bar{\gD}^{t,b} = \{\gX^{t,b}, \gY^{t,b}\}$, where $\gX^{t,b}$ and $\gY^{t,b}$ represent the input data and corresponding labels, respectively. Here, $t \in \gT = \{1, \ldots, T\}$ identifies the task, and $b \in \gB_t$ indexes the batch within task $t$ (with $\gT$ and $\gB_t$ representing their respective index sets).

Each task is characterized by a dataset $\gD_t$, assumed to be drawn from an underlying distribution $\mathbb{D}_t = p(\gX_t, \gY_t)$, where $\gD_t$ denotes the complete training set (with batch indices omitted), and similarly for $\gX_t$ and $\gY_t$. It is assumed that the training and testing data for each task follow the same distribution. In realistic scenarios, constraints may limit access to the task label $t$ and the corresponding output labels $\gY_t$. Furthermore, the training data for each task is typically received incrementally in batches.

In our proposed scenarios, we consider an RTS system in which data are collected in a continual manner. Due to the sequential nature of data collection, consecutive samples tend to exhibit high affinity, indicating strong similarity in their features. This high similarity increases the risk of overfitting, as the learning process may favor memorization of redundant patterns rather than the development of generalizable representations. Consequently, the model may become biased toward frequently occurring features, reducing its ability to perform well on diverse or unseen data.

\subsection{Generalization Gap}
The generalization gap \cite{neyshabur2017exploring} is a fundamental concept in machine learning that quantifies the difference between a model's performance on the training data and its performance on unseen test data. Formally, it is defined as the difference between the expected risk and the empirical risk:
\begin{align}
\mathfrak{R} = \mathbb{E}_{(x,y) \sim \mathcal{D}_{\textrm{train}}}[\ell(x, y; w)] - \mathbb{E}_{(x,y) \sim \mathcal{D}_{\textrm{test}}}[\ell(x, y; w)],
\end{align}
where $\mathcal{D}_{\textrm{train}}, \mathcal{D}_{\textrm{test}}$ denote the train and test data distribution, respectively. $\ell$ is the loss function, $w$ is the trained model, and $(x, y)$ are the training samples.

In the context of FCL, the test dataset $\mathcal{D}_{\textrm{test}}$ can be considered as a limitless data pool that contains all of the data of all tasks. Meanwhile, at every time step $t$, the train dataset $\mathcal{D}_{\textrm{train}}$ can only accessed to a limited data pool. As a consequence, there contains a statistical distance between two dataset.  
}

\section{FCL settings}\label{appendix:FCL-settings}

\section{Inductive generalization gap}
\label{appendix:inductive-generalization-gap}
Based on Lagrange theory \cite{2015-Math-LinearPartialDifferentialAnalysis}:
\begin{equation}
    f'(c) = \frac{f(a)-f(b)}{b-a},
\end{equation}
where $f(a) = \nabla \mathcal{L}_u\left( \omega^{(n)}+ \vartheta^{(n)}_u\right)$,$f(b) = \nabla \mathcal{L}_u\left( \omega^{(n)}\right)$, $f(c) = \nabla \mathcal{L}_u\left(\omega\right)$. As a result
\begin{equation}
\label{Eq:B3}
    \left(\nabla \mathcal{L}_u\left( \omega^{(n)}+ \vartheta^{(n)}_u\right)-\nabla \mathcal{L}_u\left( \omega^{(n)}\right)\right)=\nabla^2 \mathcal{L}_u(\omega)\vartheta^{(n)}_u
\end{equation}

Considering the inequality on the left of Lemma 5 in \cite{2020-FL-EnergyEfficientFL}, the formula~\eqref{Eq:B3} and assumption~\ref{eq:YangA1} we have
\begin{align}
\frac{1}{L} &\left\|\nabla \mathcal{L}_u\left( \omega^{(n)}+ \vartheta^{(n)}_u\right)-\nabla \mathcal{L}_u\left( \omega^{(n)}\right)\right\|^2
\notag\\
&=\frac{1}{L}(\nabla \mathcal{L}_u\left( \omega^{(n)}+ \vartheta^{(n)}_u\right)-\nabla \mathcal{L}_u\left( \omega^{(n)}\right))^\top
\notag\\
& \times (\nabla \mathcal{L}_u\left( \omega^{(n)}+ \vartheta^{(n)}_u\right)-\nabla \mathcal{L}_u\left( \omega^{(n)}\right))
\notag\\
&=\frac{1}{L}(\nabla \mathcal{L}_u\left( \omega^{(n)}+ \vartheta^{(n)}_u\right)-\nabla \mathcal{L}_u\left( \omega^{(n)}\right))^\top\nabla^2 \mathcal{L}_u( \omega)\vartheta^{(n)}_u
\notag\\
&\leq\frac{1}{L}(\nabla \mathcal{L}_u\left( \omega^{(n)}+ \vartheta^{(n)}_u\right)-\nabla \mathcal{L}_u\left( \omega^{(n)}\right))^\top L\boldsymbol{I}\vartheta^{(n)}_u
\notag\\
&=\left(\nabla \mathcal{L}_u\left( \omega^{(n)}+ \vartheta^{(n)}_u\right)-\nabla \mathcal{L}_u\left( \omega^{(n)}\right)\right)^\top  \vartheta^{(n)}_u.
\end{align}

Similarly for the inequality on the right of Lemma 5 in \cite{2020-FL-EnergyEfficientFL} provided that assumption~\ref{eq:YangA2} obtains the formula
\begin{align}\label{E:B1}
\frac{1}{L} &\left\|\nabla \mathcal{L}_u\left( \omega^{(n)}+ \vartheta^{(n)}_u\right)-\nabla \mathcal{L}_u\left( \omega^{(n)}\right)\right\|^2
\notag\\
& \leq\left(\nabla \mathcal{L}_u\left( \omega^{(n)}+ \vartheta^{(n)}_u\right)-\nabla \mathcal{L}_u\left( \omega^{(n)}\right)\right)^\top  \vartheta^{(n)}_u
\notag\\
& \leq \frac{1}{\mu}\left\|\nabla \mathcal{L}_u\left( \omega^{(n)}+ \vartheta^{(n)}_u\right)-\nabla \mathcal{L}_u\left( \omega^{(n)}\right)\right\|^2.
\end{align}

For the optimal solution $\omega^*$ of $\mathcal{L}\left( \omega^*\right)$, we always have $\nabla \mathcal{L}\left( \omega^*\right)=\mathbf{0}$. Combining~\eqref{eq:glob_loss} and~\eqref{E:B1}, we also have $\mu  {I} \preceq$ $\nabla^2 \mathcal{L}( \omega)$, which indicates that
\begin{equation}
\label{E:B4}
    \left\|\nabla \mathcal{L}( \omega)-\nabla \mathcal{L}\left( \omega^*\right)\right\| \geq \mu\left\| \omega- \omega^*\right\|
\end{equation}
and
\begin{equation}
\label{E:B5}
  \mathcal{L}(\omega^*)\geq \mathcal{L}( \omega)+\nabla \mathcal{L}( \omega)^\top\left( \omega^*- \omega\right)+\frac{\mu}{2}\left\| \omega^*- \omega\right\|^2.
\end{equation}

As a result, we have
\begin{align}
\label{E:B6}
\|\nabla \mathcal{L}( \omega)\|^2
&=\left\|\nabla \mathcal{L}( \omega)-\nabla \mathcal{L}\left( \omega^*\right)\right\|^2 \notag\\
&\stackrel{~\eqref{E:B4}}{\geq} \mu\left\|\nabla \mathcal{L}( \omega)-\nabla \mathcal{L}\left( \omega^*\right)\right\|\left\| \omega- \omega^*\right\|\notag\\
& \geq \mu\left(\nabla \mathcal{L}( \omega)-\nabla \mathcal{L}\left( \omega^*\right)\right)^\top\left( \omega- \omega^*\right)\notag\\
&=\mu \nabla \mathcal{L}( \omega)^\top\left( \omega- \omega^*\right)\notag\\
& \stackrel{~\eqref{E:B5}}{\geq} \mu\left(\mathcal{L}( \omega)-\mathcal{L}\left( \omega^*\right)\right)
 \end{align}   
which proves~\eqref{E:B2}:
\begin{equation}
    \label{E:B2}
    \|\nabla \mathcal{L}( \omega)\|^2 \geq \mu \nabla  \mathcal{L}( \omega)^\top\left( \omega- \omega^*\right).
\end{equation}
For the optimal solution of the problem~\eqref{eq:Yang11}, the first-order derivative condition always holds, i.e.,
\begin{equation}
\label{E:B7}
    \nabla \mathcal{L}_u\left( \omega^{(n)}+ \vartheta_u^{(n)*}\right)-\nabla \mathcal{L}_u\left( \omega^{(n)}\right)+\xi \nabla  \mathcal{L}\left( \omega^{(n)}\right)=\mathbf{0}
\end{equation}
\begin{equation}
\label{E:EB7}
\Rightarrow \nabla \mathcal{L}_u\left( \omega^{(n)}+ \vartheta_u^{(n) *}\right)=\nabla \mathcal{L}_u\left( \omega^{(n)}\right)-\xi \nabla \mathcal{L}\left( \omega^{(n)}\right).
\end{equation}

With Taylor expansion \cite{2012-LinearQuadraticApproximation-OptimalPolicy} we have the formula:
\begin{equation}
\label{Eq:Taylor}
    f(x_t+\Delta x) \approx f(x_t) +(\Delta x)^\top\nabla f(x_t) +\frac{1}{2}(\Delta x)^\top\nabla^2 f(x_t)\Delta x
\end{equation}
We are now ready to prove Theorem 1 with the above inequalities and equalities, we have
\begin{align}
\label{E:B8}
&\widetilde{\mathcal{L}}\left(\omega^{(n+1)}\right)=\mathcal{L}\left(\omega^{(n+1)}\right) + \mathfrak{R}_{\mathrm{glob}}^{(n+1)}\notag\\
&\stackrel{~\eqref{Eq:10}}{=}\mathcal{L}\left( \omega^{(n)}+\frac{1}{U}\sum^{U}_{u=1}\vartheta^{(n)}_u\right)+ \mathfrak{R}_{\mathrm{glob}}^{(n+1)}\notag\\
&\stackrel{~\eqref{Eq:Taylor},\mathrm{Assumption}~\ref{eq:YangA1}}{\leq}\mathcal{L}\left( \omega^{(n)}\right)  + \frac{1}{U}\sum_{u=1}^U\nabla \mathcal{L}\left(\omega^{(n)}\right)^\top \vartheta^{(n)}_u
+ \mathfrak{R}_{\mathrm{glob}}^{(n+1)} \notag\\
&\quad +\frac{L}{2 U^2}\left\|\sum_{u=1}^U\vartheta^{(n)}_u\right\|^2 \notag\\ 
&\stackrel{~\eqref{Eq:10}}{=}\mathcal{L}\left(\omega^{(n)}\right) + \mathfrak{R}_{\mathrm{glob}}^{(n+1)} + \frac{1}{U\xi}\sum_{u=1}^U\left[\mathcal{G}_u\left(\omega^{(n)},\vartheta^{(n)}_u\right)\right.\notag\\
&\left.\quad+\nabla \mathcal{L}_u\left(\omega^{(n)}\right)\vartheta^{(n)}_u-\mathcal{L}_u\left( \omega^{(n)}+ \vartheta^{(n)}_u\right)\right]\notag\\
&\quad+\frac{L}{2 U^2}\left\|\sum_{u=1}^U\vartheta^{(n)}_u\right\|^2\notag\\
&\stackrel{~\eqref{Eq:Taylor},\mathrm{Assumption}~\ref{eq:YangA2}}{\leq}\mathcal{L}\left( \omega^{(n)}\right) +  \mathfrak{R}_{\mathrm{glob}}^{(n+1)} \notag\\
&\quad+\frac{1}{U\xi}\sum_{u=1}^U\left[\mathcal{G}_u\left(\omega^{(n)},\vartheta^{(n)}_u\right) - \mathcal{L}_u\left(\omega^{(n)}\right) -\frac{\mu}{2}\left\| \vartheta^{(n)}_u\right\|^2\right]\notag\\
&\quad + \frac{L}{2 U^2}\left\|\sum_{u=1}^U\vartheta^{(n)}_u\right\|^2.
\end{align}

 To analyze Equation~\eqref{E:B8}, base on~\eqref{Eq:10}, we replace $\omega^{(n+1)}$ with $ \omega^{(n)}+\frac{1}{U}\sum^{U}_{u=1}\vartheta^{(n)}_u$. In the next, with the Taylor expansion in~\eqref{Eq:Taylor}, we decompose 
 \begin{align*}
     &\mathcal{L}\left(\omega^{(n)}+\frac{1}{U}\sum^{U}_{u=1}\vartheta^{(n)}_u\right) = \mathcal{L}\left( \omega^{(n)}\right)\notag\\  
     &+ \frac{1}{U}\sum_{u=1}^U\nabla \mathcal{L}\left(\omega^{(n)}\right)^\top  \vartheta^{(n)}_u+ \frac{\nabla^2\mathcal{L}(w)}{2 U^2}\left\|\sum_{u=1}^U\vartheta^{(n)}_u\right\|^2
 \end{align*}
then combine with assumption~\ref{eq:YangA1}, we have the inequality $\nabla^2F(\omega^{(n)})\leq LI$, we replace $\nabla^2F(\omega^{(n)})$ with $L$. Based on~\eqref{Eq:10} replace
 \begin{align*}
    &\nabla \mathcal{L}\left(\omega^{(n)}\right)^\top\vartheta^{(n)}_u
    =\frac{1}{\xi}\left[\mathcal{G}_u\left(\omega^{(n)},\vartheta^{(n)}_u\right)\right.\notag\\
    &\left.\quad+\nabla \mathcal{L}_u\left(\omega^{(n)}\right)\vartheta^{(n)}_u-\mathcal{L}_u\left( \omega^{(n)}+ \vartheta^{(n)}_u\right)\right]\notag\\
 \end{align*}
In the next step, using Taylor expansion to decompose $\mathcal{L}_u\left( \omega^{(n)}+ \vartheta^{(n)}_u\right)$, based on assumption~\ref{eq:YangA1}\\
$\nabla \mathcal{L}_u\left(\omega^{(n)}\right)\vartheta^{(n)}_u-\mathcal{L}_u\left( \omega^{(n)}+ \vartheta^{(n)}_u\right)$ then  is replaced by $- \mathcal{L}_u\left(\omega^{(n)}\right) -\frac{\mu}{2}\left\| \vartheta^{(n)}_u\right\|^2$.

According to Lemma~\ref{lemma:global-generalization-gap}, we have
\begin{align*}
    \mathfrak{R}_{\mathrm{glob}}^{(n)}=\frac{1}{U}\sum_{u=1}^U \mathfrak{R}_{u}^{(n)}, \;\;
    \mathfrak{R}_{\mathrm{glob}}^{(n+1)}=\frac{1}{U}\sum_{u=1}^U \mathfrak{R}_{u}^{(n+1)}.
\end{align*}
First, we decompose the change in generalization gap over time as follows:
\begin{align}
\label{Eq:Pr-Gap0}
&\mathfrak{R}_{u}^{(n+1)} - \mathfrak{R}_{u}^{(n)}\notag\\
&= \Big[\widetilde{\mathcal{L}}(\omega^{(n+1)}) - \mathcal{L}(\omega^{(n+1)})\Big] - \Big[\widetilde{\mathcal{L}}(\omega^{(n)}) - \mathcal{L}(\omega^{(n)})\Big] \notag\\
&= \Big[\widetilde{\mathcal{L}}(\omega^{(n+1)}) - \widetilde{\mathcal{L}}(\omega^{(n)})\Big] - \Big[\mathcal{L}(\omega^{(n+1)}) - \mathcal{L}(\omega^{(n)})\Big] \notag\\
&= \Big[\widetilde{\mathcal{L}}(\omega^{(n)} - \vartheta^{(n)}_u) - \widetilde{\mathcal{L}}(\omega^{(n)})\Big] \notag\\
&~~~- \Big[\mathcal{L}(\omega^{(n)} - \vartheta^{(n)}_u) - \mathcal{L}(\omega^{(n)})\Big] \notag\\
&= \Big[- \vartheta^{(n)}_u\nabla\widetilde{\mathcal{L}}(\omega^{(n)}) 
 + (\vartheta^{(n)}_u)^2\nabla^2\widetilde{\mathcal{L}}(\omega^{(n)}) + \mathcal{O}(\vartheta^{(n)}_u)
 \Big] \notag\\
&~~~- \Big[- \vartheta^{(n)}_u\nabla\mathcal{L}(\omega^{(n)}) 
 + (\vartheta^{(n)}_u)^2\nabla^2\mathcal{L}(\omega^{(n)}) + \mathcal{O}(\vartheta^{(n)}_u)
 \Big] \notag\\
&= \frac{1}{2}\Big[(\vartheta^{(n)}_u)^2\nabla^2\widetilde{\mathcal{L}}(\omega^{(n)}) - (\vartheta^{(n)}_u)^2\nabla^2\mathcal{L}(\omega^{(n)}) \Big] \notag\\
&~~~- \Big[\vartheta^{(n)}_u\nabla\widetilde{\mathcal{L}}(\omega^{(n)}) - \vartheta^{(n)}_u\nabla\mathcal{L}(\omega^{(n)})\Big].
\end{align}
Applying Lemmas~\ref{lemma:unbiased-gradient}, and \ref{lemma:unbiased-hessian}, we have the followings
\begin{align}
\label{Eq:Pr-Gap}
&\mathfrak{R}_{u}^{(n+1)} - \mathfrak{R}_{u}^{(n)}\notag\\
&\leq \frac{1}{2}\|\vartheta^{(n)}_u\|^2L 2^{H(\mathrm{Z})-H(\tilde{Z})-1} \notag\\
&\quad \times\sqrt{2\left[H\left(p\left(z|\mathrm{Z}\right)\right) - I\left(p\left(z|\Tilde{\mathrm{Z}}\right),p\left(z|\mathrm{Z}\right)\right)\right]} \notag\\
&- \|\vartheta^{(n)}_u\|^2 2^{H(\mathrm{Z})-H(\tilde{Z})} \notag \\
&\quad \times\sqrt{2\left[H\left(p\left(z|\mathrm{Z}\right)\right) - I\left(p\left(z|\Tilde{\mathrm{Z}}\right),p\left(z|\mathrm{Z}\right)\right)\right]} \notag\\
&= \left(L-2\right)2^{H(\mathrm{Z})-H(\tilde{Z})-1} \times \|\vartheta^{(n)}_u\|^2\notag\\
&\quad \times \sqrt{2\left[H\left(p\left(z|\mathrm{Z}\right)\right) - I\left(p\left(z|\Tilde{\mathrm{Z}}\right),p\left(z|\mathrm{Z}\right)\right)\right]}\notag\\
&=\frac{\left(L-2\right)\Psi}{2}\|\vartheta^{(n)}_u\|^2,
\end{align}
where $\Psi$ is the generalization gap statement and is demonstrated as: 
\begin{align}
\Psi = 2^{H(\mathrm{Z})-H(\tilde{Z})}\sqrt{2\left[H\left(p\left(z|\mathrm{Z}\right)\right) - I\left(p\left(z|\Tilde{\mathrm{Z}}\right),p\left(z|\mathrm{Z}\right)\right)\right]}.
\label{eq:psi}
\end{align}

Thus, component $\mathfrak{R}_{\mathrm{glob}}^{(n+1)}$ in~\eqref{E:B8} can be replaced by $\frac{1}{U}\sum_{u=1}^U \mathfrak{R}_{u}^{(n+1)}$. In addition, based on~\eqref{Eq:Pr-Gap}, component $\frac{1}{U}\sum_{u=1}^U \mathfrak{R}_{u}^{(n+1)}$ can be decomposed into the following: 
\begin{align}
\frac{1}{U}\sum_{u=1}^U \mathfrak{R}_{u}^{(n+1)} = \frac{1}{U}\sum_{u=1}^U \mathfrak{R}_{u}^{(n)} + \frac{1}{U}\sum_{u=1}^U \frac{\left(L+2\right)\Psi}{2}\|\vartheta^{(n)}_u\|^2. \notag
\end{align}

From there, Equation~\eqref{E:B8} becomes
\begin{align}
\label{Eq:Rglob_L34}
    &\widetilde{\mathcal{L}}\left(\omega^{(n+1)}\right)\notag\\
    &\leq \mathcal{L}\left( \omega^{(n)}\right) +  \frac{1}{U}\sum_{u=1}^U \mathfrak{R}_{u}^{(n+1)}\notag\\
    &\quad+\frac{1}{U\xi}\sum_{u=1}^U\left[\mathcal{G}_u\left(\omega^{(n)},\vartheta^{(n)}_u\right) - \mathcal{L}_u\left(\omega^{(n)}\right) -\frac{\mu}{2}\left\| \vartheta^{(n)}_u\right\|^2\right]\notag\\
    &\quad  +\frac{L}{2 U^2}\left\|\sum_{u=1}^U\vartheta^{(n)}_u\right\|^2 \notag\\
    &\stackrel{~\eqref{Eq:Pr-Gap}}{\leq} \mathcal{L}\left( \omega^{(n)}\right) + \frac{1}{U}\sum_{u=1}^U \mathfrak{R}_{u}^{(n)} + \frac{1}{U}\sum_{u=1}^U \frac{\left(L+2\right)\Psi}{2}\|\vartheta^{(n)}_u\|^2 \notag\\
    &\quad+\frac{1}{U\xi}\sum_{u=1}^U\left[\mathcal{G}_u\left(\omega^{(n)},\vartheta^{(n)}_u\right) - \mathcal{L}_u\left(\omega^{(n)}\right) -\frac{\mu}{2}\left\| \vartheta^{(n)}_u\right\|^2\right]\notag\\
    &\quad  +\frac{L}{2 U^2}\left\|\sum_{u=1}^U\vartheta^{(n)}_u\right\|^2.
\end{align}
Based on 
$$
\mathcal{L}\left( \omega^{(n)}\right) + \frac{1}{U}\sum_{u=1}^U \mathfrak{R}_{u}^{(n)} = \mathcal{L}\left( \omega^{(n)}\right) +\mathfrak{R}^{(n)}_{\mathrm{glob}} = \widetilde{\mathcal{L}}\left(\omega^{(n)}\right)
$$
and according to the triangle inequality and mean inequality that
\begin{equation}
\label{E:B9}
    \left\|\frac{1}{U} \sum_{u=1}^U  \vartheta^{(n)}_u\right\|^2 \leq\left(\frac{1}{U} \sum_{u=1}^U\left\| \vartheta^{(n)}_u\right\|\right)^2 \leq \frac{1}{U} \sum_{u=1}^U\left\| \vartheta^{(n)}_u\right\|^2,
\end{equation}
we use $\widetilde{\mathcal{L}}\left(\omega^{(n)}\right)$ in place of $\mathcal{L}\left( \omega^{(n)}\right) + \frac{1}{U}\sum_{u=1}^U \mathfrak{R}_{u}^{(n)}$. Besides, based on~\eqref{E:B9}, $\frac{L}{2 U^2}\left\|\sum_{u=1}^U\vartheta^{(n)}_u\right\|^2$ could be replaced by $\frac{L}{2 U^2}\sum_{u=1}^U\left\|\vartheta^{(n)}_u\right\|^2$. Combining the commutative, associative, and distributive properties in mathematics, Equation~\eqref{Eq:Rglob_L34} can be transformed as follows 
\begin{align}
    &\widetilde{\mathcal{L}}\left(\omega^{(n+1)}\right) \notag \\
    &\stackrel{~\eqref{E:B9}}{\leq}\widetilde{\mathcal{L}}\left(\omega^{(n)}\right) + \frac{1}{U}\sum_{u=1}^U \frac{\left(L+2\right)\Psi}{2}\|\vartheta^{(n)}_u\|^2\notag\\
    &\quad+\frac{1}{U\xi}\sum_{u=1}^U\left[\mathcal{G}_u\left(\omega^{(n)},\vartheta^{(n)}_u\right) - \mathcal{L}_u\left(\omega^{(n)}\right) -\frac{\mu}{2}\left\| \vartheta^{(n)}_u\right\|^2\right]\notag\\
    &\quad  +\frac{L}{2 U^2}\sum_{u=1}^U\left\|\vartheta^{(n)}_u\right\|^2.
\end{align}
Thus, we have
\begin{align}\label{Eq:TildeF}
    &\widetilde{\mathcal{L}}\left(\omega^{(n+1)}\right) \notag \\
    &= \widetilde{\mathcal{L}}\left(\omega^{(n)}\right)+
    \frac{1}{U\xi}\sum_{u=1}^U\left[\mathcal{G}_u\left(\omega^{(n)},\vartheta^{(n)}_u\right) - \mathcal{L}_u\left(\omega^{(n)}\right)\right.\notag\\
    &\left.\quad \frac{1}{2}\left(\xi\left(L+2\right)\Psi+\frac{\xi L}{U}-\mu\right)\left\| \vartheta^{(n)}_u\right\|^2\right].
\end{align}

Based on~\eqref{Eq:EE10}, we can obtain
\begin{align}\label{Eq:G2hkstar}
    &\mathcal{G}_u\left(\omega^{(n)},\vartheta^{(n)}_u\right) - \mathcal{L}_u\left(\omega^{(n)}\right)\notag\\
    &\stackrel{~\eqref{eq:Yang11}}{=}\mathcal{G}_u\left( \omega^{(n)},\vartheta^{(n)}_u\right)-\mathcal{G}_u\left( \omega^{(n)},0\right)\notag\\
    &=\mathcal{G}_u\left( \omega^{(n)},\vartheta^{(n)}_u\right)-\mathcal{G}_u\left( \omega^{(n)},  \vartheta_u^{(n)*}\right)\notag\\
    &\quad-\left[\mathcal{G}_u\left(\omega^{(n)}, 0\right)-\mathcal{G}_u\left( \omega^{(n)},  \vartheta_u^{(n)*}\right)\right]\notag\\
    &\stackrel{~\eqref{eq:Yang15}}{\leq}(\varpi-1)\left[\mathcal{G}_u\left(\omega^{(n)}, 0\right)-\mathcal{G}_u\left(\omega^{(n)}, \vartheta_u^{(n)*}\right)\right]\notag\\
    &\stackrel{~\eqref{eq:Yang11}}{=}(\varpi-1)\left[\mathcal{L}_u\left( \omega^{(n)}\right)-\mathcal{L}_u\left( \omega^{(n)}+ \vartheta_u^{(n)*}\right)\right.\notag\\
    &\left.\quad+\left(\nabla \mathcal{L}_u\left( \omega^{(n)}\right)-\xi \nabla \mathcal{L}\left( \omega^{(n)}\right)\right)^\top\vartheta_u^{(n)*}\right]\notag\\
    &\stackrel{~\eqref{E:EB7}}{=}(\varpi-1)\left[\mathcal{L}_u\left( \omega^{(n)}\right)-\mathcal{L}_u\left( \omega^{(n)}+ \vartheta_u^{(n) *}\right)\right.\notag\\
    &\left.\quad +\nabla \mathcal{L}_u\left( \omega^{(n)}+ \vartheta_u^{(n) *}\right)^\top  \vartheta_u^{(n) *}\right].
\end{align}

Based on Assumption~\ref{eq:YangA2}, we can obtain
\begin{align}\label{Eq:hkstar}
    &\mathcal{L}_u\left( \omega^{(n)}\right)\geq \mathcal{L}_u\left( \omega^{(n)}+ \vartheta_u^{(n) *}\right)\notag\\
    &\quad -\nabla \mathcal{L}_u\left( \omega^{(n)}+ \vartheta_u^{(n) *}\right)^\top\vartheta_u^{(n)*} + \frac{\mu}{2}\left\| \vartheta_u^{(n)*}\right\|^2\notag\\
    &\Leftrightarrow \mathcal{L}_u\left( \omega^{(n)}\right)- \mathcal{L}_u\left( \omega^{(n)}+ \vartheta_u^{(n) *}\right)\notag\\
    &\quad +\nabla \mathcal{L}_u\left( \omega^{(n)}+ \vartheta_u^{(n) *}\right)^\top\vartheta_u^{(n)*} \geq \frac{\mu}{2}\left\| \vartheta_u^{(n)*}\right\|^2
\end{align}

Applying~\eqref{Eq:hkstar} to~\eqref{Eq:G2hkstar}, we can obtain
\begin{equation}\label{Eq:2hkstar}
    \mathcal{G}_u\left(\omega^{(n)},\vartheta^{(n)}_u\right) - \mathcal{L}_u\left(\omega^{(n)}\right)\leq (1-\varpi)\frac{\mu}{2}\left\| \vartheta_u^{(n)*}\right\|^2.
\end{equation}

Applying~\eqref{Eq:2hkstar} to~\eqref{Eq:TildeF}, we can receive
\begin{align}\label{Eq:TildeFstar}
    &\widetilde{\mathcal{L}}\left(\omega^{(n+1)}\right) \leq\widetilde{\mathcal{L}}\left(\omega^{(n)}\right)+
    \frac{1}{U\xi}\sum_{u=1}^U\left[(1-\varpi)\frac{\mu}{2}\left\| \vartheta_u^{(n)*}\right\|^2\right.\notag\\
    &\left.\quad + \frac{1}{2}\left(\xi\left(L+2\right)\Psi+\frac{\xi L}{U}-\mu\right)\left\| \vartheta^{(n)}_u\right\|^2\right].
\end{align}

From~\eqref{E:B1}, the following relationship is received:
\begin{equation}
\label{E:B13}
    \left\| \vartheta_u^{(n) *}\right\|^2 \geq \frac{1}{L^2} \| \nabla \mathcal{L}_u\left( \omega^{(n)}+ \vartheta_u^{(n) *}\right)-\nabla \mathcal{L}_u\left( \omega^{(n)}\right) \|^2
\end{equation}
To streamline the training process for local clients, we employ several techniques to enhance efficiency and performance. These include early stopping for termination of training, small batch gradient descent for improved optimization, and dropout to regularize the AI model. By implementing these strategies, we can achieve a more stable convergence of the local gradient training, resulting in reduced fluctuations and enhanced generalization. Consequently, we have:
\begin{equation}\label{Eq:MDLaos}
    \left\| \vartheta^{(n)}_u\right\|^2 \leq\left\| \vartheta_u^{(n) *}\right\|^2.
\end{equation}
Because of the lack of data changes caused by the environment, $\left\| \vartheta_u^{(n) *}\right\|^2$ cannot capture the entire observation of data, then $\nabla \mathcal{L}_u$ fluctuates more than $\nabla \widetilde{\mathcal{L}}_u$, and \eqref{E:B13} becomes
\begin{equation}
\label{E:B13towF}
    \left\Vert \vartheta_u^{(n) *}\right\Vert^2 \geq \frac{1}{L^2} \left\Vert \nabla \widetilde{\mathcal{L}}_u\left( \omega^{(n)}+ \vartheta_u^{(n) *}\right)-\nabla \widetilde{\mathcal{L}}_u\left( \omega^{(n)}\right) \right\Vert^2
\end{equation}


According to~\eqref{Eq:MDLaos} and~\eqref{Eq:TildeFstar}, we can obtain
\begin{align}\label{Eq:B15}
&\mathcal{L}\left(\omega^{(n+1)}\right) \leq \mathcal{L}\left(\omega^{(n)}\right)\notag\\
&\quad -\frac{\left(\xi\left(L+2\right)\Psi+\frac{\xi L}{U}-\varpi\mu\right)}{2 U \xi} \sum_{u=1}^U\left\|\vartheta_u^{(n) *}\right\|^2\notag\\
&\leq \mathcal{L}\left(\omega^{(n)}\right)-\frac{\left(\xi\left(L+2\right)\Psi+\frac{\xi L}{U}-\varpi\mu\right)}{2 UL^2 \xi} \sum_{u=1}^U\left[\right.\notag\\
&\left.\times\left\|\nabla \mathcal{L}_u\left( \omega^{(n)}+ \vartheta_u^{(n) *}\right)-\nabla \mathcal{L}_u\left( \omega^{(n)}\right)\right\|^2\right]\notag\\
&=\mathcal{L}\left( \omega^{(n)}\right)
-\frac{\left(\xi\left(L+2\right)\Psi+\frac{\xi L}{U}-\varpi\mu\right)}{2 UL^2 \xi}\left\|\nabla \mathcal{L}\left(\omega^{(n)}\right)\right\|^2\notag\\
&\leq \mathcal{L}\left(\omega^{(n)}\right)-\frac{\xi\left(L+2\right)\Psi+\frac{\xi L}{U}-\varpi\mu}{2 UL^2 \xi}\left(\mathcal{L}\left(\omega^{(n)}\right)-\mathcal{L}\left(\omega^*\right)\right).
\end{align}
Based on \eqref{Eq:B15}, we obtain:
\begin{align}\label{Eq:B16}
&\mathcal{L}\left(\omega^{(n+1)}\right)-\mathcal{L}\left(\omega^*\right)\notag\\
&  \leq\left(1-\frac{\left[\xi\left(L+2\right)\Psi+\frac{\xi L}{U}-\varpi\mu\right]}{2 UL^2 \xi}\right)\left(\mathcal{L}\left(\omega^{(n)}\right)-\mathcal{L}\left(\omega^*\right)\right)\notag\\
&\leq\left(1-\frac{\left[\xi\left(L+2\right)\Psi+\frac{\xi L}{U}-\varpi\mu\right]}{2 UL^2 \xi}\right)^{n+1}\notag\\
&\quad\times\left(\mathcal{L}\left(\omega^{(0)}\right)-\mathcal{L}\left(\omega^*\right)\right)\notag\\
&\leq \exp \left(-(n+1) \frac{\left[\xi\left(L+2\right)\Psi+\frac{\xi L}{U}-\varpi\mu\right]}{2 UL^2 \xi}\right)\notag\\
&\quad\times\left(\mathcal{L}\left(\omega^{(0)}\right)-\mathcal{L}\left(\omega^*\right)\right).
\end{align}
where the last inequality follows from the fact that $1-x \leq$ $\exp (-x)$. To ensure that $\mathcal{L}\left( \omega^{(n+1)}\right)-\mathcal{L}\left( \omega^*\right) \leq \varrho\left(\mathcal{L}\left( \omega^{(0)}\right)-\right.$ $\mathcal{L}\left(\omega^*\right)$ ), we have~\eqref{eq:I_glob}.

\section{Proof on lemma~\ref{lemma:global-generalization-gap}: Global generalization gap}
\label{appendix:global-generalization-gap}
We consider the global generalization gap: 
\begin{align}
\mathfrak{R}_{\mathrm{glob}} 
&= \mathbb{E}_{\{x_i, y_i\} \sim \mathcal{D}} \left[ \mathbb{E}_{(x,y)\sim \mathcal{D}^{\mathrm{train}}} \left[\ell(x,y;w) \right] - \right. \notag \\    & \left. ~~~~~~~~~~~~~~~~~~
\mathbb{E}_{(x,y)\sim \mathcal{D}^{\mathrm{test}}}\left[\ell(x,y;w) \right] \right] \notag \\
&= \mathbb{E}_{\{x_i, y_i\} \sim \mathcal{D}} \left[ \mathbb{E}_{u\in \mathcal{U}} \Big\{\mathbb{E}_{(x,y)\sim \mathcal{D}^{\mathrm{train}}_{u}} \left[\ell(x,y;w) \right]\Big\} - \right. \notag \\    
& \left. ~~~~~~~~~~~~~~~~~~
\mathbb{E}_{(x,y)\sim \mathcal{D}^{\mathrm{test}}}\left[\ell(x,y;w) \right] \right] \notag \\
&= \mathbb{E}_{\{x_i, y_i\} \sim \mathcal{D}} \left[ \mathbb{E}_{u\in \mathcal{U}} \Big\{\mathbb{E}_{(x,y)\sim \mathcal{D}^{\mathrm{train}}_{u}} \left[\ell(x,y;w) \right]\Big\} - \right. \notag \\    
& \left. ~~~~~~~~~~~~~~~~~~
\mathbb{E}_{u\in \mathcal{U}} \mathbb{E}_{(x,y)\sim \mathcal{D}^{\mathrm{test}}}\left[\ell(x,y;w) \right] \right] \notag \\ 
&= \mathbb{E}_{u\in \mathcal{U}} \Big\{\mathbb{E}_{\{x_i, y_i\} \sim \mathcal{D}} \left[ \mathbb{E}_{(x,y)\sim \mathcal{D}^{\mathrm{train}}_{u}} \left[\ell(x,y;w) \right] - \right. \notag \\    
& \left. ~~~~~~~~~~~~~~~~~~
\mathbb{E}_{(x,y)\sim \mathcal{D}^{\mathrm{test}}}\left[\ell(x,y;w) \right]  \right] \Big\}.
\end{align}



\section{Proof on lemma~\ref{lemma:unbiased-gradient}: Unbiased Gradient and Bounded Variance}
\label{appendix:unbiased-gradient}
We decompose the gradients as follows: 
\begin{align}
\label{Eq:nFnF}
    &\nabla\tilde{\mathcal{L}}\left(\omega^{(n)}\right) - \nabla \mathcal{L}\left(\omega^{(n)}\right)\notag\\
    &=\Big\lVert\sum_{z\in\mathrm{Z}}\nabla\ell(z,w)\left[p\left(z|\Tilde{\mathrm{Z}}\right)-p\left(z|\mathrm{Z}\right)\right]\Big\rVert\notag\\
    &\overset{\textrm{(a1)}}{\leq} \lVert\nabla\ell(z,w)\rVert\Big\lVert p\left(z|\Tilde{\mathrm{Z}}\right)-p\left(z|\mathrm{Z}\right)\Big\rVert\notag\\
    &\overset{(\textrm{(a2)})}{\leq} \|\vartheta^{(n)}_u\|\frac{1}{p\left(z|\mathrm{Z}\right)}\sqrt{2\mathrm{D_{KL}}\left[p\left(z|\Tilde{\mathrm{Z}}\right)\Vert p\left(z|\mathrm{Z}\right)\right]},
\end{align}
where $\textrm{(a1)}$ is due to the Hölder's inequality, and $\textrm{(a2)}$ is inferred based on the Pinsker Inequality \cite{2011-MF-Pinsker}. Because $\Tilde{\mathrm{Z}}$ is the empirical global dataset, we take $p\left(z|\Tilde{\mathrm{Z}}\right) = 1$ so we have
\begin{equation}
\label{Eq:pz2Hz}
    p\left(z|\mathrm{Z}\right) = \frac{p\left(z|\Tilde{\mathrm{Z}}\right)p\left(\tilde{Z}\right)}{p(Z)} = \frac{p(\tilde{Z})}{p(Z)} = 2^{H(\mathrm{Z})-H(\tilde{Z})},
\end{equation}
with $H(\cdot)$ is the entropy of the dataset.

Based on the information theory we have the following: 
\begin{align}
\label{Eq:Dkl2I} 
    &\mathrm{D_{KL}}\left[p\left(z|\Tilde{\mathrm{Z}}\right)\Vert p\left(z|\mathrm{Z}\right)\right]\notag\\
    &= H\left(p\left(z|\Tilde{\mathrm{Z}}\right),p\left(z|\mathrm{Z}\right)\right) - H\left(p\left(z|\Tilde{\mathrm{Z}}\right)\right)\notag\\
    &= \left[H\left(p\left(z|\Tilde{\mathrm{Z}}\right),p\left(z|\mathrm{Z}\right)\right) - H\left(p\left(z|\Tilde{\mathrm{Z}}\right)\right)-H\left(p\left(z|\mathrm{Z}\right)\right)\right]\notag\\
    &\quad+ H\left(p\left(z|\mathrm{Z}\right)\right)\notag\\
    &= H\left(p\left(z|\mathrm{Z}\right)\right)\notag\\
    &\quad -\left[H\left(p\left(z|\Tilde{\mathrm{Z}}\right)\right) + H\left(p\left(z|\mathrm{Z}\right)\right) - H\left(p\left(z|\Tilde{\mathrm{Z}}\right),p\left(z|\mathrm{Z}\right)\right)\right]\notag\\
    &= H\left(p\left(z|\mathrm{Z}\right)\right) - I\left(p\left(z|\Tilde{\mathrm{Z}}\right),p\left(z|\mathrm{Z}\right)\right).
\end{align}
Arcording to the \eqref{Eq:nFnF}, \eqref{Eq:pz2Hz}, \eqref{Eq:Dkl2I} we have 
\begin{align}
    &\nabla\tilde{\mathcal{L}}\left(\omega^{(n)}\right) - \nabla \mathcal{L}\left(\omega^{(n)}\right)\notag\\
    &\leq \|\vartheta^{(n)}_u\|2^{H(\mathrm{Z})-H(\tilde{Z})}\sqrt{2\left[H\left(p\left(z|\mathrm{Z}\right)\right) - I\left(p\left(z|\Tilde{\mathrm{Z}}\right),p\left(z|\mathrm{Z}\right)\right)\right]}
\end{align}

\section{Proof on lemma~\ref{lemma:unbiased-hessian}: Unbiased Hessian and Bounded Variance}
\label{appendix:unbiased-hessian}
We decompose the Hessians as follows: 
\begin{align}
\label{Eq:nFnF-2}
    &\nabla^2\tilde{\mathcal{L}}\left(\omega^{(n)}\right) - \nabla^2 \mathcal{L}\left(\omega^{(n)}\right)\notag\\
    &=\Big\lVert\sum_{z\in\mathrm{Z}}\nabla^2\ell(z,w)\left[p\left(z|\Tilde{\mathrm{Z}}\right)-p\left(z|\mathrm{Z}\right)\right]\Big\rVert\notag\\
    &\overset{(b1)}{\leq} \lVert\nabla^2\ell(z,w)\rVert\Big\lVert p\left(z|\Tilde{\mathrm{Z}}\right)-p\left(z|\mathrm{Z}\right)\Big\rVert\notag\\
    &\overset{(b2)}{\leq} L\frac{1}{p\left(z|\mathrm{Z}\right)}\sqrt{2\mathrm{D_{KL}}\left[p\left(z|\Tilde{\mathrm{Z}}\right)\Vert p\left(z|\mathrm{Z}\right)\right]},
\end{align}
where $(b1)$ is due to the Hölder's inequality, and $(b2)$ is inferred based on the Pinsker Inequality \cite{2011-MF-Pinsker}, and the hessian follows the $L$-smooth assumption. Because $\Tilde{\mathrm{Z}}$ is the empirical global dataset, we take $p\left(z|\Tilde{\mathrm{Z}}\right) = 1$ so we have:
\begin{equation}
\label{Eq:pz2Hz-2}
    p\left(z|\mathrm{Z}\right) = \frac{p\left(z|\Tilde{\mathrm{Z}}\right)p\left(\tilde{Z}\right)}{p(Z)} = \frac{p(\tilde{Z})}{p(Z)} = 2^{H(\mathrm{Z})-H(\tilde{Z})}
\end{equation}
with $H(\cdot)$ is the entropy of the dataset. According to \eqref{Eq:Dkl2I}, \eqref{Eq:nFnF-2} and \eqref{Eq:pz2Hz-2}  we have
\begin{align}
\label{Eq:nF_nF-2}
    &\nabla^2\tilde{\mathcal{L}}\left(\omega^{(n)}\right) - \nabla^2 \mathcal{L}\left(\omega^{(n)}\right)\notag\\
    &\leq L2^{H(\mathrm{Z})-H(\tilde{Z})}\sqrt{2\left[H\left(p\left(z|\mathrm{Z}\right)\right) - I\left(p\left(z|\Tilde{\mathrm{Z}}\right),p\left(z|\mathrm{Z}\right)\right)\right]}.
\end{align}

\section{Proof on lemma~\ref{lemma:ECS-P}: Boundaries of Explicit Constraints}\label{appendix:ECS_x}
The function of the user's neural network is defined as $Y = \mathbf{W}^{\mathrm{max}}_u \sigma(\mathbf{W}(x)),$
where $\mathbf{W}(x)$ is the neural network. The sigma is bounded by $0<\sigma<1$. Therefore, $0<Y<\mathbf{W}^{\mathrm{max}}$.
For instance, $P_\mathrm{ECS}$ of local computation capacity $f_u$ is defined as follows: 
    \begin{align}
    P_\mathrm{ECS}(f_u) = \abs{\max\left\langle f_u - f^{\mathrm{max}}_u, 0 \right\rangle} + \abs{\max\left\langle -f_u, 0 \right\rangle} \geq 0,
    \end{align}

\section{Proof on Theorem~\ref{theorem:ECS_reward}: Boundaries of Explicit Constraints}\label{appendix:ECS_reward}
Using constraints in the format of definition~\ref{def:ECS-P}, we obtain the total penalty for the EI-integrated DRL system as follows: 
\begin{align}
    \mathbf{r}_\mathrm{DRL-EI} = R - \mathcal{P}_1 - \mathcal{P}_2 - \sum_i P_{\mathrm{ECS-EI},i},
\end{align}
and the total penalty for a DRL system without EI design as: 
\begin{align}
    \mathbf{r}_\mathrm{DRL} = R - \mathcal{P}_1 - \mathcal{P}_2 - \sum_i P_{\mathrm{ECS},i},
\end{align}
Thus, we have the following for the optimal reward: 
\begin{align}
    \mathbf{r}^*_\mathrm{DRL-EI}
    &= R^* - \mathcal{P}_1^* - \mathcal{P}_2^* - \sum\nolimits_i P^*_{\mathrm{ECS-EI},i} \notag \\ 
    &\overset{(a)}{=} R^* - \mathcal{P}_1^* - \mathcal{P}_2^* \notag \\ 
    &\geq R^* - \mathcal{P}^*_1 - \mathcal{P}^*_2 - \sum\nolimits_i P^*_{\mathrm{ECS},i} \notag \\
    &\geq \mathbf{r}^*_\mathrm{DRL-EI},
\end{align}
where the equality (a) holds due to Lemma~\ref{lemma:ECS-P}. In other words, the optimal penalty following the ECS is determined as $\mathbf{r}^*_\mathrm{DRL-EI} \geq \mathbf{r}^*_\mathrm{DRL}$.

\section{Proof on Lemma~\ref{lemma:worst-case-time}}
We have: 
\begin{align}
    I_\mathrm{glob}(k_u, \varpi) \geq \underset{\mathbf{\tau}, \mathbf{\varpi}}{\min}~I_\mathrm{glob}(k_u, \varpi).
\end{align}
Therefore, we have: 
\begin{align}
    T_\textrm{max} 
    &\geq I_\mathrm{glob}(k_u, \varpi)\times\left(\frac{A_u \log_2 (1/\varpi)}{f_u} + t^\mathrm{trans}_u\right)\\
    &\geq 
    \underset{\mathbf{\tau}, \mathbf{\varpi}}{\min}~I_\mathrm{glob}(k_u, \varpi)\times \left(\frac{A_u \log_2 (1/\varpi)}{f_u} + t^\mathrm{trans}_u\right).\notag
\end{align}
\end{document}